\documentclass{article}

\usepackage[preprint]{acl}

\usepackage{times}
\usepackage{latexsym}
\usepackage[utf8]{inputenc}
\usepackage[T1]{fontenc}
\usepackage{booktabs}
\usepackage{rotating}
\usepackage{amsmath,amssymb,amsfonts,amsthm}
\usepackage{comment}
\usepackage{xcolor}
\usepackage{microtype}
\usepackage{url}
\usepackage{graphicx}
\usepackage{adjustbox}
\usepackage{placeins}


\newif\ifshowrunlinks

\newcommand{\snorm}[1]{\left\lVert#1\right\rVert}

\DeclareMathOperator{\Tr}{Tr}

\newcommand{\diag}[1]{\mathrm{diag}}
\newcommand{\rank}{\mathrm{rank}}
\newcommand{\erank}{\mathrm{erank}}
\newcommand{\rerank}{\mathrm{erank}_2}

\newcommand{\R}{\mathbb{R}}

\newtheorem{problem}{Problem}

\newtheorem{theorem}{Theorem}
\newtheorem{definition}{Definition}

\newcommand{\penaltyname}{R\'enyi-2 penalty}

\title{Maximizing Capacity Utilization of LoRA Adapters for Mitigating the Learning-Forgetting Trade-off}

\author{Tomohiro Sawada \thanks{Corresponding author: \texttt{tsawada@gatech.edu}.}, Kartik Goyal \\
School of Interactive Computing\\
Georgia Institute of Technology
}

\begin{document}
\raggedbottom

\maketitle

\begin{abstract}
Low-Rank Adaptation (LoRA) often underperforms full fine-tuning, especially in heterogeneous and continual-learning settings in which substantial forgetting occurs. In general, it suffers from an undesirable learning--forgetting trade-off across diverse fine-tuning settings. We identify a potential contributor: LoRA adapters, especially high-rank ones, frequently fail to realize the available capacity afforded by the configured rank. Specifically, we find that standard LoRA updates have effective rank far below their configured rank, limiting how broadly the update energy is distributed. Therefore, we propose training mechanisms that optimize the original objective while maximizing adapter effective rank. Our main method adds a simple, stable, differentiable, and SVD-free R\'enyi-2 effective-rank regularizer to the language-modeling loss. We also connect capacity maximization to norm-bounded updates.
In mixed-domain rank sweeps on three models evaluated with
multiple random seeds, a penalty weight of $\lambda=0.01$
lowers mean Wiki perplexity at every tested rank of at least
$16$ relative to unpenalized LoRA. Mean training perplexity
improves or increases by at most $0.02$ in all but one
model--rank comparison. Single-task and scheduling experiments
further show that the balance between learning and retention
depends on the configuration.
We also observe that several LoRA variants in the literature also suffer from the identified capacity issue, and our approach provides a complementary axis of improvement to LoRA-based parameter-efficient fine-tuning.
These results suggest that effective rank is not only a post-hoc diagnostic of adapter quality, but a trainable objective for improving the learning--forgetting frontier.
Our code is available at \url{https://github.com/tomohiro-sawada/lora-capacity-utilization}.
\end{abstract}

\section{Introduction}

Low-rank adaptation (LoRA; \citealp{hu2022lora}) is a standard parameter-efficient alternative to full fine-tuning: it scales to large models, is easy to compose with existing training stacks, and is widely treated as the default way to specialize a pretrained language model on a new task or domain~\cite{schulman2025lora,biderman2024lora}.
A common piece of practitioner wisdom is that increasing LoRA rank increases adapter capacity: a larger $r$ enlarges the low-rank update subspace and can support richer task-specific changes~\cite{raschka2023lora,schulman2025lora,liu-etal-2024-alora}.
This intuition is appealing and often borne out at moderate ranks, yet it rests on an implicit assumption that is rarely checked: that the adapter \emph{actually uses} the subspace it is given.

In parallel, fine-tuning, whether full-rank or parameter-efficient, faces a persistent learning--forgetting tension.
Fitting a target distribution can improve task metrics while erasing competence on broad pretraining domains, and heterogeneous or sequential training regimes exacerbate the trade-off~\cite{kirkpatrick2017overcoming,hu2022lora}.
LoRA is not exempt: in mixed-domain and continual-learning settings, adapters often lag full fine-tuning on target-task fit while still paying a substantial retention cost on pretraining proxies~\cite{biderman2024lora,shuttleworth2024loraillusion}.
Prior work has proposed many remedies, such as
stronger parameterizations~\cite{vera},  %
norm constraints~\cite{wang2025normboundedlowrankadaptation}, %
spectral initialization~\cite{pissa}, %
and deliberately tiny adapters~\cite{tinylora}. %
However, these approaches frequently \emph{reallocate} the trade-off rather than improve it: methods that fit the fine-tuning mixture well can forget aggressively, whereas methods that limit forgetting often underfit the new data (Figure~\ref{fig:mixed-tasks-train-forgetting-tradeoff}).

\begin{figure*}[t]
    \centering
    \includegraphics[width=\textwidth]{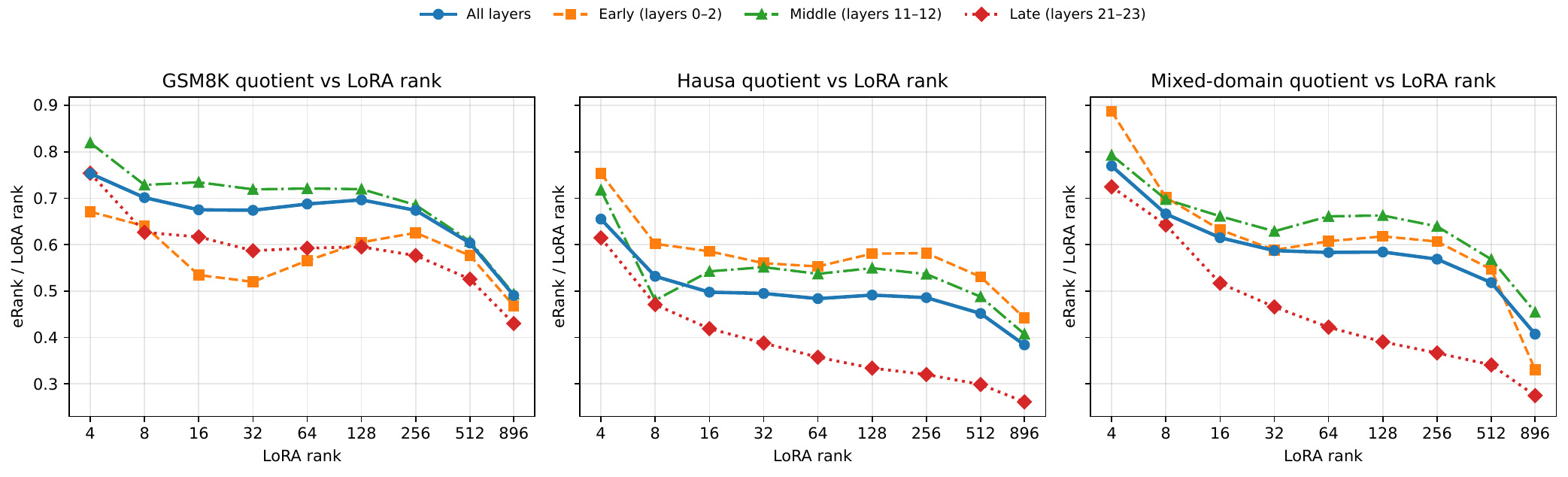}
    \caption{
        Effective-rank utilization of standard LoRA adapters across LoRA ranks, decomposed by depth.
        Each curve reports the mean utilization quotient \(\erank(\Delta W)/r\) averaged over the adapted MLP \texttt{down\_proj} layers in that group (all 24 layers; early layers 0--2; middle layers 11--12; late layers 21--23), for GSM8K (left), Hausa (center), and mixed-domain continual-learning (45\% GSM8K, 10\% WMT19 Czech--English, 45\% Aya Hausa; right) training sweeps.
        Line styles denote depth groups: all layers use solid circles, early layers dashed squares, middle layers dash-dot triangles, and late layers dotted diamonds.
        The aggregate curve falls at high rank because realized effective rank grows sublinearly in \(r\); depth groups reveal that underutilization is not uniform---late layers often fall below early and middle layers.
    }
    \label{fig:lora-rank-quotient}
\end{figure*}

We argue that a useful missing variable in this picture is not nominal configured rank $r$ but the actual realized adapter capacity that we measure in terms of \emph{effective rank}.
Let $\Delta W$ denote the accumulated LoRA update on a target layer.
Following \citet{roy2007effective}, we measure how many directions effectively carry singular-value mass using the effective rank $\erank(\Delta W)$, a continuous relaxation of matrix rank obtained from the entropy of the normalized spectrum.
Intuitively, $\erank(\Delta W)$ reports how uniformly the update distributes its singular-value mass across its singular directions: a highly skewed spectrum behaves like a low-rank update in practice even when $r$ is large.
Across three fine-tuning settings, standard LoRA training underutilizes capacity in exactly this sense: we observe in Figure~\ref{fig:lora-rank-quotient} that the utilization quotient $\erank(\Delta W)/r$ tends to be much smaller than $1.0$ and even more strikingly, it falls as configured rank grows (Section~\ref{sec:capacity}).
The key observation is therefore that \textbf{configured rank is an unreliable proxy for usable capacity}: scaling $r$ alone does not reliably deliver ``more adapter'' in the sense that matters for learning and retention.

This distinction matters because the learning--forgetting trade-off is not only about \emph{how large} an update is, but about \emph{how that update is distributed} across the orthogonal directions of the weight.
Spectral analyses of finetuned weights in prior work emphasize that adaptation can introduce directions weakly aligned with the pretrained spectrum (so-called \emph{intruder dimensions}), and that the prevalence and weighting of such directions correlate with forgetting~\cite{shuttleworth2024loraillusion}.
However, in our analysis (\S~\ref{sec:analysis}), every method that fits the mixture well, dense fine-tuning included, places most of its singular-value mass on such directions; only the methods that underfit (VeRA, TinyLoRA, SVF) avoid them.
Therefore, we argue that the relevant question is not whether intruder dimensions exist, but whether the \emph{dominant} part of the update uses the allocated rank budget in a balanced, information-bearing way.
We therefore ask: \emph{can we improve the learning--forgetting frontier by maximizing realized capacity during training?}
Since $\erank(\Delta W)\le r$~\cite{roy2007effective}, increasing utilization means driving the normalized spectrum toward uniformity subject to the task loss.

Concretely, we study regularized training objectives that augment the standard language-modeling loss $L$ with a capacity regularization term that focuses on maximizing the effective rank.
Direct spectral differentiation requires care at rank-deficient updates and incurs SVD overhead (Appendix~\ref{apdx:direct-differentiation-of-erank}.1). We therefore use an SVD-free surrogate, the R\'enyi-2 effective rank $\rerank(\Delta W)$. We develop two approaches to increasing this effective-rank surrogate.
First, \textbf{norm-bounded capacity maximization} couples a Roy--Vetterli lower bound with Norm-Bounded LoRA (NB-LoRA)~\cite{wang2025normboundedlowrankadaptation}: by enforcing an operator-norm cap and maximizing the Frobenius norm within that cap, one can raise effective rank through a principled constrained reparameterization.
This strategy works in practice but limits the adapter weight updates to a specialized parameterization which often results in unstable training empirically.
Therefore, we propose a simpler, differentiable, parameterization-agnostic and easy-to-optimize regularization strategy called the \textbf{\penaltyname} as our main method, which directly maximizes $\rerank(\Delta W)$.
This penalty is not only applicable to vanilla LoRA, but also to improved spectral variants such as PiSSA~\cite{pissa} and LoRA-XS~\cite{lora-xs}.

We evaluate whether higher realized rank shifts the learning--forgetting frontier across settings that stress different aspects of adaptation.
In mixed-domain fine-tuning, at $\lambda=0.01$, the penalty lowers
mean Wiki perplexity relative to unpenalized LoRA at every tested
rank of at least $16$ on Qwen2.5-0.5B, Qwen2.5-3B-Instruct, and
Llama-3.2-3B-Instruct. Training perplexity is generally similar
or improved, with a larger increase for Qwen2.5-3B-Instruct at
$r=16$ (Section~\ref{sec:lora-rank-sweep}).

Increasing $\lambda$ raises the reported effective rank. At the maximal rank on Qwen2.5-0.5B (Figure~\ref{fig:mixed-tasks-train-forgetting-tradeoff}), $\lambda=0.01$ and $0.1$ improve both mixture fit and retention, whereas the single run at $\lambda=1$ improves retention with Train PPL within standard LoRA's run-to-run variation. PiSSA and LoRA-XS~\cite{pissa,lora-xs} have the highest Wiki perplexities, despite lower reported mean StarCoder and GSM8K perplexities than standard LoRA; VeRA, SVF, and TinyLoRA~\cite{vera,singularvaluefinetuning,tinylora} limit forgetting primarily by underfitting the mixture. Applied to PiSSA and LoRA-XS, the regularizer can improve both task fit and retention. The same pattern holds on single-task fine-tuning setups like GSM8K, where our method, especially with late-start schedules, improves target-task fit while reducing forgetting on the reference probes.
Our singular-direction analysis~(\S~\ref{sec:analysis}) shows what the penalty leaves unchanged: penalized and standard LoRA place a similar share of singular-value mass on directions with no strong counterpart in the pretrained basis (0.97 vs. 0.96), yet the penalized adapters forget far less.

In summary, our core contributions are: \textbf{i)} we identify systematic \textbf{capacity underutilization} in standard LoRA training, especially at high configured rank, and argue that realized effective rank, not nominal rank, is the relevant unit of analysis for adapter capacity, \textbf{ii)} we formulate \textbf{effective-rank maximization} as a principled objective atop standard language-modeling loss, with both a norm-bounded reparameterization route (NB-LoRA) and a parameterization-agnostic surrogate,
 \textbf{iii)} we introduce the \textbf{\penaltyname}, a unitary-invariant, SVD-free effective-rank regularizer that is simple to implement and stable to optimize, and demonstrate improvements in the learning--forgetting trade-off across mixed-domain and single-task fine-tuning settings, and \textbf{iv)} we provide \textbf{spectral diagnostics} (intruder mass, soft global coupling) that provide a novel lens to characterize the relationship between singular directions of updated weights and the pretrained weights.

\section{Realized adapter capacity}
\label{sec:capacity}

Let $W_0 \in \R^{m\times n}$ denote the pretrained weight on a target layer and $W = W_0 + \Delta W$ the adapted weight after fine-tuning, where $\Delta W$ is the accumulated LoRA update.
Adapter \emph{capacity} is a property of $\Delta W$, not of nominal rank $r$ alone.

\begin{definition}[\citealp{roy2007effective}]
\label{def:roy-vetterli-erank}
Let $\Delta W\ne0$ and let $\sigma_i$, $1 \le i \le r$, be its singular values.
The \textbf{Roy--Vetterli effective rank} of $\Delta W$ is
\begin{equation}
    \erank(\Delta W) := \exp\bigl(H_1(\mathbf p)\bigr),
\end{equation}
where $H_1$ is the Shannon entropy and $\mathbf p$ is the normalized spectrum
\begin{equation}\label{eq:normalized-singular-values}
    p_i := \frac{\sigma_i}{\sum_j \sigma_j}.
\end{equation}
\end{definition}

This is the standard continuous relaxation of rank used in the spectral-complexity literature: $\erank(\Delta W)=1$ for a rank-one update and approaches the configured LoRA rank $r$ when singular values are nearly uniform.
Nominal rank $r$ counts how many directions are \emph{available}; $\erank(\Delta W)$ measures how evenly the update \emph{uses} them, so configured rank is an unreliable proxy for usable capacity.

Standard LoRA training underutilizes this budget: across rank sweeps on GSM8K, Hausa instruction following, and our mixed-domain continual-learning proxy, the utilization quotient $\erank(\Delta W)/r$ decreases as $r$ grows (Figure~\ref{fig:lora-rank-quotient}).
The drop is depth-dependent rather than uniform---late layers typically underutilize rank more than early or middle layers.
Because the learning--forgetting trade-off depends on \emph{how} update energy is distributed across directions, not only on nominal rank, we ask whether deliberately raising utilization during training can improve retention while fitting new data.
Section~\ref{sec:methods} turns this measurement into a trainable objective with two implementations.

\section{Maximizing realized capacity during LoRA training}
\label{sec:methods}

Our goal is to fit the language-modeling loss $L$ while maximizing \emph{realized} adapter capacity.
We want to minimize objectives of the form

\begin{equation}\label{eq:joint-optimization-problem-statement}
    L_{\mathrm{Cap}}(\Delta W) := L(W_0+\Delta W) - \lambda\,\mathrm{Cap}(\Delta W)
\end{equation}

\noindent where $\mathrm{Cap}(\cdot)$ is an effective-rank surrogate and $\lambda \ge 0$ trades off task fit against utilization.
The objective must be \textbf{sensitive to how singular-value \emph{energy} is distributed} across adapter directions (a skewed spectrum means low realized capacity), while remaining \emph{robust to adapter parameterization and stable when singular values cluster} (avoiding explicit subspace gradients through degenerate SVDs).
We develop two routes: \textbf{norm-bounded reparameterization} (NB-LoRA; §\ref{sec:nb-lora-route}) and a \textbf{parameterization-agnostic R\'enyi-2 penalty} (§\ref{sec:renyi-penalty}, our main method).
\begin{problem}[Ideal capacity-maximization objective]
\label{prob:joint-erank-objective}
Minimize \eqref{eq:joint-optimization-problem-statement} where $\mathrm{Cap}(\Delta W) = \erank(\Delta W)$ is the Roy--Vetterli effective rank (Definition~\ref{def:roy-vetterli-erank}).
\end{problem}
We pursue two alternatives based on the SVD-free surrogate $\rerank(\Delta W)=\exp(H_2(\mathbf q))$, which lower-bounds $\erank(\Delta W)$ by Theorem~\ref{thm:renyi-2-bound}. Here, $\mathbf q$ is the \emph{energy distribution}: $q_i := \frac{\sigma_i^2}{\sum_j \sigma_j^2}$.

\subsection{Why not optimize $\erank(\Delta W)$ directly?}
\label{sec:why-not-svd}

Direct spectral differentiation requires care at rank-deficient updates and incurs SVD overhead (Appendix~\ref{apdx:direct-differentiation-of-erank}.1). Losses that depend on singular vectors can involve inverse squared-singular-value gaps, but a loss depending only on singular values does not intrinsically have this instability at repeated positive values. For Roy--Vetterli effective rank, derivatives become singular as a normalized singular value tends to zero. We therefore use an SVD-free surrogate.

\subsection{Norm-bounded capacity maximization (NB-LoRA)}
\label{sec:nb-lora-route}

Since $H_1(\mathbf p) \ge H_2(\mathbf q)$ for the corresponding normalized spectra (Appendix~\ref{apdx:renyi-2-bound-proof}), it suffices to raise the R\'enyi-2 entropy $H_2(\mathbf q)$, which is maximized when the energy distribution $\mathbf q$ is uniform.
On a compact set this motivates maximizing the squared Frobenius norm $\snorm{\Delta W}_{S_2}^2$ subject to an operator-norm cap.

\begin{theorem}
\label{thm:renyi-2-bound-via-schatten}
Let $\Delta W\ne0$ and $\delta>0$. If $\snorm{\Delta W}_{S_{\infty}} \le \delta$, then $H_2(\mathbf q) \ge 2\log \frac{\snorm{\Delta W}_{S_2}}{\delta}$,
and hence
\begin{equation*}
 \erank(\Delta W) \ge \exp(H_2(\mathbf q)) \ge \frac{\snorm{\Delta W}_{S_2}^2}{\delta^2}.
\end{equation*}
\end{theorem}

See Appendix~\ref{apdx:renyi-2-bound-via-schatten} for a proof.
Problem~\ref{prob:joint-erank-objective} can therefore be approached via the auxiliary problem
\begin{equation}
    \label{eq:nb-lora-capacity-objective}
    \min_{\snorm{\Delta W}_{S_{\infty}} \le \delta} \; L(W_0 + \Delta W) - \lambda \, \frac{\snorm{\Delta W}_{S_2}^2}{\delta^2}.
\end{equation}

We enforce $\snorm{\Delta W}_{S_{\infty}} \le \delta$ with the Norm-Bounded LoRA (NB-LoRA) parameterization~\cite{wang2025normboundedlowrankadaptation}.
For a diagonal $S\in\mathbb{R}^{r\times r}$ with nonnegative entries, $\snorm{S}_{S_{\infty}} = \delta, \Delta W_{\mathrm{NB\text{-}LoRA}} = 2Y^\top S X,$
where $X,Y$ are obtained via a Stiefel--Cayley map~\cite{absil2009optimization,trockman2021orthogonalizing}.
Then $\snorm{\Delta W}_{S_{\infty}} \le \snorm{S}_{S_{\infty}} = \delta$, satisfying Theorem~\ref{thm:renyi-2-bound-via-schatten}.

NB-LoRA is feasible and raises effective rank on Hausa multilingual instruction following (Table~\ref{tab:aya-hausa-nb-lora}).
However, this ties the updates to a Stiefel--Cayley parameterization~\cite{absil2009optimization,trockman2021orthogonalizing} which can make optimization difficult in our experiments.
Our main method below avoids this reparameterization entirely (§\ref{sec:renyi-penalty}).

\subsection{R\'enyi-2 penalty: parameterization-agnostic capacity maximization}
\label{sec:renyi-penalty}

We optimize a stable surrogate of $\erank$ that depends only on the energy spectrum of $\Delta W$.
The \textbf{\penaltyname} uses R\'enyi-2 (collision) entropy~\cite{renyi1961measures} and can be added to the standard LM loss without changing the adapter parameterization; its gradients do not involve singular-value-gap denominators (see §\ref{sec:why-not-svd}).

\begin{definition}[\penaltyname]
\label{def:renyi-2-penalty}
Let $\Delta W\ne0$ and let $\sigma_i$, $1 \le i \le r$, be its singular values.
The \textbf{\penaltyname} is
\begin{equation*}
    \rerank(\Delta W) := \exp(H_2(\mathbf q)) = \frac{\snorm{\Delta W}_F^4}{\Tr{(\Delta W^\top \Delta W)^2}},
\end{equation*}
\end{definition}
\noindent Intuitively, $\rerank(\Delta W)$ measures the effective number of LoRA directions the update actually uses, based only on how energy is distributed across singular directions.
\begin{theorem}\label{thm:renyi-2-bound}
\begin{equation}
	\rank(\Delta W) \ge \erank(\Delta W) \ge \rerank(\Delta W) \ge 1
\end{equation}
\end{theorem}
\noindent The first two inequalities are equalities when the spectrum is uniform; the third is an equality if and only if exactly one singular value is nonzero. See Appendix~\ref{apdx:proof-of-basic-properties} for a proof and other properties.
Maximizing $\rerank$ therefore raises a lower bound on Roy--Vetterli $\erank$.

\begin{problem}[Effective-rank penalized LoRA optimization]
\label{prob:penalized-lora-objective}
Let $L$ be a training objective.
We consider
\begin{equation}\label{eq:penalized-loss}
    \min_{\Delta W} \; L(W_0 + \Delta W) - \lambda\,\rerank(\Delta W),
\end{equation}
where $\lambda \ge 0$ may be fixed or scheduled (Section~\ref{sec:weight-scheduler}).
\end{problem}

In experiments, we use the logarithmic variant $(-\lambda\log\rerank(\Delta W))$, which rescales the per-module gradient of the original penalty by $(1/\rerank(\Delta W))$; see Appendix~\ref{apdx:implementation-tricks} for implementation details.

As mentioned above, this penalty is not exclusive to LoRA and can be applied to advanced variants of LoRA like PiSSA and LoRA-XS (§\ref{sec:pissa-lora-xs-penalty}).
We evaluate the R\'enyi-2 surrogate without an SVD using the trace-ratio formula in Definition~\ref{def:renyi-2-penalty}. The surrogate is also unitarily invariant (Appendix~\ref{apdx:proof-of-basic-properties}).
Hence, this penalty is smooth and differentiable while avoiding challenges like differentiating through SVD and managing reparameterizations for weight updates. Moreover, we also need not materialize $\Delta W$ itself: the penalty is computed from low-rank factors $A$ and $B$ (Appendix~\ref{apdx:implementation-tricks}).
Thus, our method provides a simple, SVD-free way to encourage higher realized capacity while optimizing the language-modeling objective (Appendix~\ref{apdx:gradient-of-renyi-2-penalty}).

\section{Experiments}
\label{sec:experiments}
We primarily experimented with Qwen2.5~\cite{qwen}: 0.5B base, 3B-instruct, and 7B-instruct.
For \emph{single-task} fine-tuning settings, we used math reasoning~(GSM8K~\cite{cobbe2021training}), low-resource language instruction fine-tuning~(Aya Hausa~\cite{singh2024aya}), and machine translation (WMT19 cs-en~\cite{wmt19translate}) datasets.
We also experimented in a supervised \emph{mixed-domain fine-tuning setting that proxies continual learning} -- we fine-tune with assistant-only SFT on a fixed $10^6$-token budget over a shuffled mixture of three domains: GSM8K ($45\%$), WMT19 Czech--English ($10\%$), and Aya Hausa instruction-following ($45\%$).
To \emph{measure forgetting} in this continual learning setting, we report held-out perplexities on fixed reference slices from the Wiki and StarCoder components of OLMo-Mix-1124~\cite{olmo20242olmo2furious} (pretraining-distribution proxies, denoted Wiki PPL and StarCoder PPL) and on GSM8K and WMT19 Czech--English (in-mixture math and translation retention, respectively).
Baselines include standard LoRA, dense down-projection fine-tuning, and the following adapter variants: PiSSA~\cite{pissa}, LoRA-XS~\cite{lora-xs}, TinyLoRA~\cite{tinylora}, SVF~\cite{singularvaluefinetuning}, VeRA~\cite{vera}, and unpenalized NB-LoRA~\cite{wang2025normboundedlowrankadaptation}. Unless noted, adapters target all MLP \texttt{down\_proj} layers at LoRA rank $r{=}896$; dense down-projection fine-tuning updates the same module subset. We typically experiment with $\lambda\in\{0,0.01,0.1,1.0\}$.
For full details about our setup, baselines, and datasets, refer to Appendix~\ref{sec:appendix-baselines}.
Below, we report representative slices of our empirical comparison to summarize our major findings and takeaways. Additional comparisons appear in Appendix~\ref{sec:additional-results}.

\subsection{Mixed-domain fine-tuning results}
\label{sec:mixed-task-results}

\begin{figure*}[t]
    \centering
    \includegraphics[width=\textwidth]{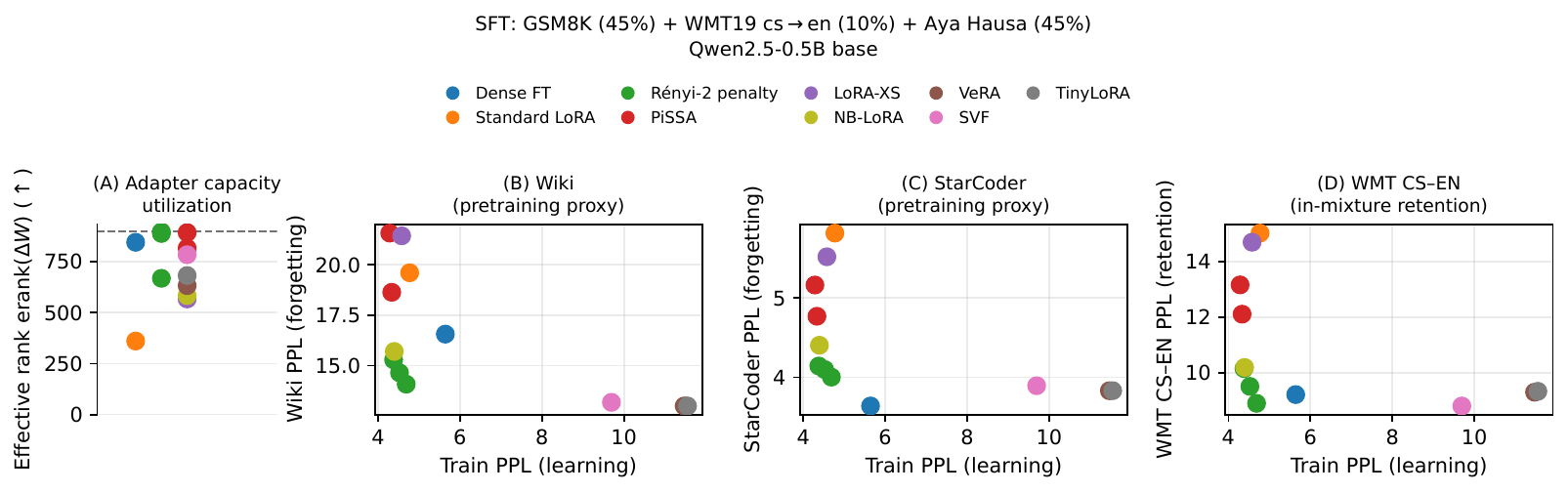}
    \caption{
        Dense FT denotes dense down-projection fine-tuning. Mixed-domain continual-learning baselines on Qwen2.5-0.5B (base, not Instruct; rank $r{=}896$ unless noted).
        Each point represents a method--penalty configuration; the legend maps dot colors to adapter type. Points show multi-run means or single-run endpoints; sample SDs are reported in Appendix Table~\ref{tab:mixed-task-baseline-comparison}.
        \textbf{(A)} Effective rank $\mathrm{erank}(\Delta W)$ shown as a vertical strip from $0$ to nominal rank $896$ (horizontal dashed line), with horizontal offsets only for visibility; higher indicates fuller adapter capacity utilization.
        \textbf{(B--D)} Learning--forgetting trade-offs: Train PPL (mixture learning) versus Wiki PPL, StarCoder PPL (pretraining proxies), and WMT CS--EN PPL (in-mixture translation retention); lower-left is better.
        The three \penaltyname{} points use different regularization weights $\lambda$.
        Unpenalized PiSSA and LoRA-XS have the highest Wiki perplexities; VeRA, SVF, and TinyLoRA limit forgetting with weaker mixture fit. NB-LoRA approaches the $\lambda=0.01$ learning--retention trade-off.
        Unpenalized PiSSA uses two runs after excluding one resumed execution with apparent data replay; the other multi-run points use three executions. Dense FT, LoRA at $\lambda=1$, and penalized PiSSA at $\lambda=0.1$ are single runs.
        Full numeric values and the PiSSA exclusion are documented in Appendix Table~\ref{tab:mixed-task-baseline-comparison} and its accompanying text.
    }
    \label{fig:mixed-tasks-train-forgetting-tradeoff}
\end{figure*}

We first evaluate whether increasing adapter effective rank mitigates the learning--forgetting trade-off in a \emph{heterogeneous} supervised fine-tuning setting that proxies continual learning.
We report the Roy--Vetterli effective rank $\erank(\Delta W)$ of the accumulated update (higher indicates fuller use of available adapter capacity), training-mixture perplexity (Train PPL), and perplexity on the retention probes described above. This reported diagnostic differs from the R\'enyi-2 surrogate optimized by the penalty.
Lower perplexity is better throughout. Figure~\ref{fig:mixed-tasks-train-forgetting-tradeoff} summarizes all methods across adapter capacity (panel~A) and the learning--forgetting trade-off (panels~B--D).
The favorable region in the trade-off panels is the lower-left corner (strong learning with limited forgetting).
Appendix Table~\ref{tab:mixed-task-baseline-comparison} reports the full numbers; entries with $\pm$ are means and sample standard deviations over three training executions, except PiSSA ($n=2$), and the remaining rows are single runs. PiSSA excludes one resumed execution with apparent data replay (Appendix~\ref{subsec:mixed-domain-baselines-numeric}); its small two-run SD does not establish low population variability.

\paragraph{Comparison to standard LoRA and dense down-projection fine-tuning.}
Standard LoRA attains Train PPL $4.77$ but underutilizes capacity ($\erank=361$ in panel~A). The dense down-projection fine-tuning baseline (a single run) has weaker mixture fit (Train PPL $5.64$) but lower StarCoder and GSM8K perplexities ($3.64$ and $1.817$, versus $5.82$ and $1.929$ for standard LoRA).
Penalized LoRA at $\lambda=0.1$ improves Train and Wiki PPL relative to this dense baseline ($4.53$ vs.\ $5.64$ and $14.65$ vs.\ $16.56$), while the dense baseline retains lower StarCoder and GSM8K PPL. Thus neither method dominates on all metrics.

\paragraph{Methods that learn the mixture but forget pretraining.}
PiSSA~\cite{pissa} and LoRA-XS~\cite{lora-xs} fit the mixture at least as well as standard LoRA (Train PPL $4.29$ and $4.58$ vs.\ $4.77$), but do not uniformly improve retention.
Both have higher Wiki perplexity than standard LoRA ($21.6$ and $21.4$ vs.\ $19.6$), despite lower StarCoder and GSM8K perplexities ($5.16$ and $5.52$; $1.878$ and $1.901$, versus $5.82$ and $1.929$ for standard LoRA).
A single penalized PiSSA run at $\lambda=0.1$ has lower Wiki and StarCoder perplexity than the unpenalized mean ($18.6$ vs.\ $21.6$ and $4.77$ vs.\ $5.16$) but slightly higher Train PPL ($4.34$ vs.\ $4.29$); Section~\ref{sec:pissa-lora-xs-penalty} examines the penalty on these variants in a controlled sweep.
\paragraph{Methods that limit forgetting by underfitting the mixture.}
At the other extreme, VeRA~\cite{vera}, SVF~\cite{singularvaluefinetuning}, and TinyLoRA keep Wiki and StarCoder perplexity low ($13.0$--$13.2$ and $3.8$--$3.9$), but at the cost of weak mixture learning: Train PPL remains $9.7$--$11.5$. They reflect heavily constrained adapters, limited in their ability to fit heterogeneous training mixtures, thereby preserving pretraining fit.

\paragraph{\penaltyname{} improves the learning--forgetting trade-off.}
Adding the \penaltyname{} to standard LoRA increases the effective rank with $\lambda$ ($668$, $888$, and $896$ at $\lambda\in\{0.01,0.1,1.0\}$, vs.\ $361$ without it) and shifts solutions toward the favorable region of panels~B--D.
Relative to standard LoRA, every penalty weight improves Wiki, StarCoder, and GSM8K perplexity; $\lambda{=}0.01$ and $\lambda{=}0.1$ also \emph{improve} Train PPL ($4.38$ and $4.53$ vs.\ $4.77$).
At $\lambda{=}1.0$ (a single run), Wiki perplexity is $5.5$ points and GSM8K perplexity $0.13$ points below the standard-LoRA mean, with Train PPL inside the baseline's run-to-run variation ($4.69$ vs.\ $4.77\pm0.09$). The corresponding WMT cs--en reduction is $6.1$ points ($15.02\rightarrow8.91$); WMT and GSM8K are separate probes.
For $\lambda=0.01$ and $0.1$, the improvement over unpenalized LoRA is at least twice the baseline's run-to-run standard deviation on every metric, so both learning and retention improve.
The same within-rank pattern holds on Llama-3.2-3B-Instruct (Table~\ref{tab:arxiv-curated-llama-3p2-3b-it-baselines}: at $r{=}3072$, $\lambda{=}0.1$ lowers Train PPL from $3.98$ to $3.68$ and Wiki PPL from $18.5$ to $13.3$) and on Pythia-2.8B (Table~\ref{tab:arxiv-curated-pythia-2p8b-rank-sweep}: at $r{=}2560$, Wiki PPL falls from $24.9$ to $15.3$), with small Wiki gains on OLMo-2 (Tables~\ref{tab:arxiv-curated-olmo-2-13b-it-rank-sweep}--\ref{tab:arxiv-curated-olmo-2-32b-it-rank-sweep}).

\paragraph{Norm-bounded LoRA recovers much of the gain.}
Unpenalized NB-LoRA, the norm-bounded route of Section~\ref{sec:nb-lora-route}, also raises effective rank ($583$ vs.\ $361$) and improves on standard LoRA on every metric (Train PPL $4.40$, Wiki $15.7$, StarCoder $4.40$, GSM8K $1.809$).
It lands close to the \penaltyname{} at $\lambda{=}0.01$ (Train $4.38$, Wiki $15.3$, StarCoder $4.14$), while $\lambda{=}0.1$ retains more (Wiki $14.6$, StarCoder $4.10$) at slightly higher Train PPL ($4.53$).
This is consistent with Theorem~\ref{thm:renyi-2-bound-via-schatten}, but it also means that these experiments do not separate effective-rank maximization from norm control.

\subsection{Effectiveness of Norm-bounded Instantiation}
While most of the paper is focused on our \penaltyname{} approach, in this section we observe that our other instantiation based on norm-bounded LoRA~(§\ref{sec:nb-lora-route}), although a little unstable, also performs reasonably well. Table~\ref{tab:aya-hausa-nb-lora} compares NB-LoRA, with and without the Frobenius-norm capacity term in Equation~\eqref{eq:nb-lora-capacity-objective}, against standard LoRA and dense down-projection fine-tuning. We observe that simply bounding the norm raises $\erank$ ($504$ vs.\ $341$) and lowers Wiki perplexity ($20.1$ vs.\ $21.1$). Increasing $\lambda$ raises effective rank further and lowers Wiki perplexity to $19.1$. Train PPL ($23.7$--$24.3$) and StarCoder PPL ($4.02$--$4.11$) are slightly worse than standard LoRA's ($22.6$ and $4.00$), and the training fit is worse than dense down-projection fine-tuning's ($21.0$). In the mixed-domain setting, unpenalized NB-LoRA improves on standard LoRA on every metric (Section~\ref{sec:mixed-task-results}).
\begin{table}[t]
    \centering
    \small
\setlength{\tabcolsep}{2pt}
    \begin{adjustbox}{max width=\columnwidth}
    \begin{tabular}{@{}lrrrr@{}}
    \toprule
    Method &  \shortstack{eRank\\($\uparrow$)} & \shortstack{Train PPL\\($\downarrow$)} & \shortstack{StarCoder PPL\\($\downarrow$)} & \shortstack{Wiki PPL\\($\downarrow$)} \\
    \midrule
{Dense FT}                     & 805.510        & 20.977    & 4.243 & 23.393 \\
{Standard LoRA}                       & 340.517       & 22.580     & 3.998 & 21.098 \\
    \midrule
{Norm-bounded LoRA} & 503.963 & 23.919 & 4.071 & 20.089 \\
{$\lambda$ = 0.01} & 527.513 & 23.729 & 4.024 & 19.815 \\
{$\lambda$ = 0.1} & 663.745 & 23.923 & 4.108 & 19.331 \\
{$\lambda$ = 1.0} & 814.076 & 24.307 & 4.048 & 19.066 \\
    \bottomrule
    \end{tabular}
    \end{adjustbox}
    \caption{Dense FT denotes dense down-projection fine-tuning. NB-LoRA on Aya Hausa using Qwen2.5-0.5B. Here, $\lambda$ weights the Frobenius-norm capacity term in Equation~\eqref{eq:nb-lora-capacity-objective}; the unpenalized NB-LoRA row uses $\lambda=0$.
}
\label{tab:aya-hausa-nb-lora}
\end{table}

\subsection{\penaltyname{} on LoRA variants}
\label{sec:pissa-lora-xs-penalty}
Since our approach is readily applicable to LoRA variants~(\S~\ref{sec:methods}), we apply the \penaltyname{} to PiSSA and LoRA-XS -- the two variants with the highest Wiki perplexity in Figure~\ref{fig:mixed-tasks-train-forgetting-tradeoff}. Table~\ref{tab:gsm8k-pissaloraxs-rank-penalty} shows that the penalty improves their results on single-task GSM8K adaptation.
\paragraph{PiSSA: high rank, poor retention.}
At $\lambda{=}0$, PiSSA already attains large $\erank$ ($829$, near dense down-projection fine-tuning's $839$) but severe Wiki forgetting.
Increasing $\lambda$ monotonically raises $\erank$ to $896$ and reduces Wiki and StarCoder perplexity by $3.5$ and $0.4$ points, respectively, while slightly improving GSM8K Train PPL. %
The penalty therefore \emph{reshapes} how an already high-rank PiSSA update uses its degrees of freedom, rather than merely increasing nominal rank.

\paragraph{LoRA-XS: under-used rank, severe forgetting.}
LoRA-XS without penalty also demonstrates severe forgetting with a moderate $\erank=565$.
Our regularizer increases $\erank$ by approximately $330$ to $895$ and cuts Wiki perplexity by $5.4$ points at $\lambda{=}1$, with the best Train PPL in the sweep.
This is direct evidence that subspace-constrained adapters can leave capacity on the table, and that maximizing the R\'enyi-2 effective-rank surrogate can reduce this underutilization.

While our regularizer improves PiSSA and LoRA-XS, neither matches standard LoRA with the R\'enyi-2 penalty as shown in Table~\ref{tab:gsm8k-baseline-comparison}.
Thus rank regularization is a broadly useful add-on, but the choice of adapter still matters for the learning--forgetting trade-off.

\begin{table}[tbp]
\centering
\small
\setlength{\tabcolsep}{3pt}
\begin{adjustbox}{max width=\columnwidth}
\begin{tabular}{@{}lrrrr@{}}
\toprule
Method                           & \shortstack{eRank\\($\uparrow$)}   & \shortstack{Train\\PPL ($\downarrow$)} & \shortstack{StarCoder\\PPL ($\downarrow$)} & \shortstack{Wiki\\PPL ($\downarrow$)} \\ \midrule
PiSSA ($\lambda=0$)                  & 829.002      & 1.871   & 4.323        & 23.173 \\
$\lambda=0.01$      & 877.896      & 1.869   & 4.233        & 22.255 \\
$\lambda=0.1$       & 893.976      & 1.851   & 4.104        & 20.951 \\
$\lambda=1.0$       & 895.722      & 1.825   & 3.934        & 19.706 \\
\midrule
LoRA-XS ($\lambda=0$)                  & 564.981      & 1.892   & 5.066        & 25.479 \\
$\lambda=0.01$      & 721.998      & 1.904   & 4.625        & 23.195 \\
$\lambda=0.1$       & 880.442      & 1.896   & 4.371        & 21.398 \\
$\lambda=1.0$       & 895.027      & 1.847   & 4.134        & 20.121 \\
\bottomrule
\end{tabular}
\end{adjustbox}
\caption{
    PiSSA and LoRA-XS penalty-weight comparisons on single-task GSM8K adaptation for Qwen2.5-0.5B, targeting MLP down-projection matrices ($r=896$, $\alpha=1792$). Train PPL measures GSM8K fit; StarCoder and Wiki PPL are held-out retention probes. For both methods, the penalty ramps from step $0$, with width $W=200$ for all nonzero penalty weights shown.
}
\label{tab:gsm8k-pissaloraxs-rank-penalty}
\end{table}

\subsection{Effect of \penaltyname{} across LoRA rank configurations}
\label{sec:lora-rank-sweep}
\begin{figure*}[t]
    \centering
    \includegraphics[width=\linewidth]{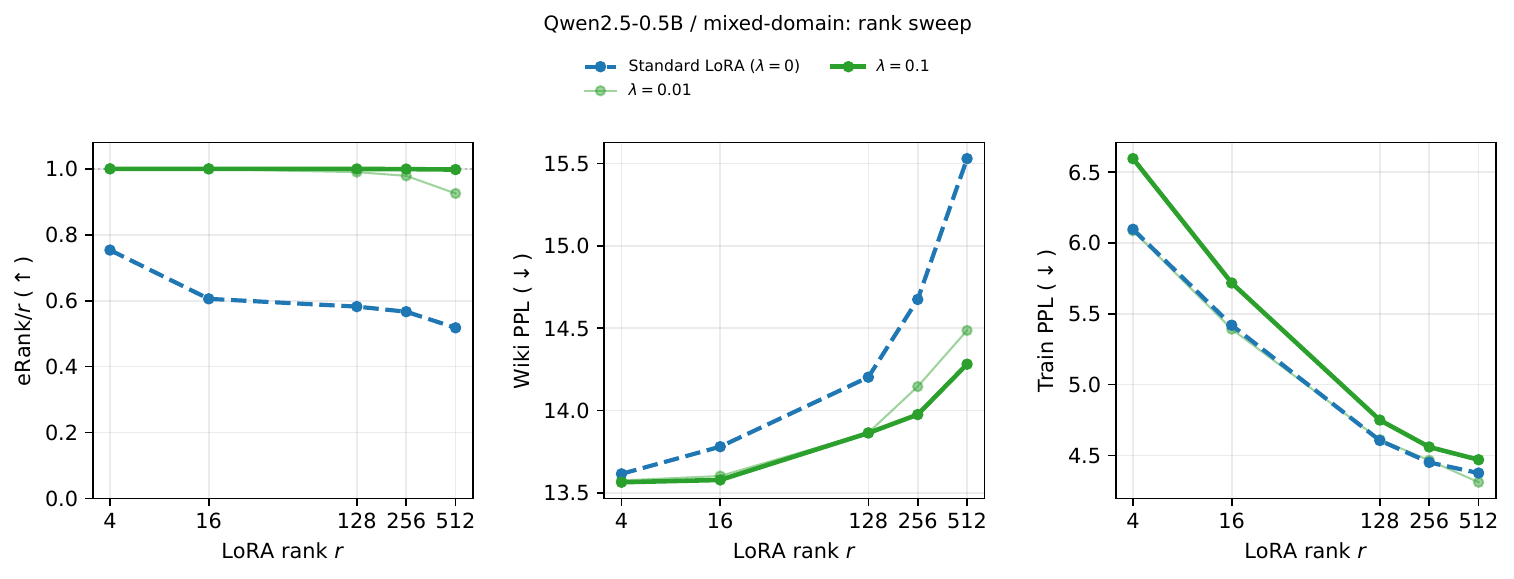}
    \caption{%
        LoRA rank sweep on Qwen2.5-0.5B on the mixed-domain continual-learning proxy setup~(aggregated metrics in Appendix Table~\ref{tab:arxiv-curated-qwen-2p5-0p5b-rank-sweep}).
        The panels show effective-rank utilization, Wiki perplexity, and training perplexity. Points use the table's means across three seeds per setting, with repeated executions averaged within each seed first; effective-rank utilization divides the eRank mean by $r$. Colors indicate penalty weight $\lambda$. At $r=256,\lambda=0.1$, the reported means of both Train PPL and Wiki PPL are lower than for standard LoRA at $r=128$.
    }
    \label{fig:mixed-combined-05b-rank-sweep}
\end{figure*}

Our main mixed-domain comparison uses the maximal matrix rank $r{=}896$ for the adapted projections. We now examine smaller rank budgets~($r\in\{4,16,128,256,512\}$) to distinguish the effect of nominal rank from spectral utilization.
Figure~\ref{fig:mixed-combined-05b-rank-sweep} summarizes this on Qwen2.5-0.5B with assistant-only SFT on the mixed-domain continual-learning proxy setting~(run means; full metrics in Appendix Table~\ref{tab:arxiv-curated-qwen-2p5-0p5b-rank-sweep}).

\paragraph{Cross-model retention gains at a small penalty weight.}
Across the reported mixed-domain rank sweeps on Qwen2.5-0.5B,
Qwen2.5-3B-Instruct, and Llama-3.2-3B-Instruct,
$\lambda=0.01$ lowers mean Wiki perplexity relative to
unpenalized LoRA at every tested rank of at least $16$
(Tables~\ref{tab:arxiv-curated-qwen-2p5-0p5b-rank-sweep},
\ref{tab:arxiv-curated-qwen-2p5-3b-it-rank-sweep},
and~\ref{tab:arxiv-curated-llama-3p2-3b-it-rank-sweep}).
Mean Train PPL improves or increases by at most $0.020$ in
all but one of these comparisons; the exception is
Qwen2.5-3B-Instruct at $r=16$, where it increases from
$4.211$ to $4.524$.
These results support a consistent Wiki retention benefit
in this setting, but not universal improvement on both objectives.

\paragraph{Scaling rank on standard LoRA trades learning for forgetting.}
For standard LoRA ($\lambda{=}0$), Train PPL improves steadily with $r$ ($6.10$ at $r{=}4$ to $4.38$ at $r{=}512$), but Wiki and StarCoder perplexity \emph{degrade} over the same range ($13.62\rightarrow 15.53$ and $3.93\rightarrow 4.18$).
Echoing Figure~\ref{fig:lora-rank-quotient}, the effective rank remains well below the nominal rank budget ($\erank=145.0$ at $r{=}256$ and $265.0$ at $r{=}512$, i.e.\ only $\approx 57\%$ and $\approx 52\%$ utilization).
\paragraph{Our penalty increases capacity utilization and flattens the forgetting cost of rank.}
Across ranks, the penalty raises $\erank$ toward the ceiling $r$: $\lambda{=}0.01$ reaches $92$--$100\%$ of $r$, and at $\lambda{=}0.1$ we observe $\erank\approx r$ at every rank in the sweep.
The retention benefit of the penalty is rank-dependent: at $r{=}4$, Wiki and StarCoder perplexities are nearly unchanged, whereas at $r{=}128$, $256$, and $512$, $\lambda{=}0.01$ reduces Wiki by $0.34$, $0.53$, and $1.04$ PPL relative to unpenalized LoRA at the same $r$, with Train PPL unchanged (within $0.07$). A larger weight, $\lambda{=}0.1$, lowers Wiki by up to $0.2$ further but costs Train PPL at every rank, from $0.50$ at $r{=}4$ to $0.09$ at $r{=}512$.
At $r{=}256$, our approach with $\lambda{=}0.1$ achieves Train PPL $4.56$ and Wiki $13.98$, improving both target-task fit and retention relative to standard LoRA at $r{=}128$ (Train $4.61$, Wiki $14.20$).

\paragraph{Limits at extreme ranks.}
At $r{=}1$, both effective-rank definitions equal $1$ for every nonzero update, so the exact capacity term is constant. At the smallest measured rank, $r{=}4$, increasing utilization produces only small changes in retention, whereas the separate $r{=}896$ comparison shows substantial within-rank improvements. Reaching the rank ceiling also does not guarantee a benefit: at $r{=}4$, $\lambda{=}0.1$ saturates $\erank$ but raises Train PPL from $6.10$ to $6.60$ with no clear retention gain. Appendix~\ref{subsec:extreme-ranks} discusses these limits and the distinction between maximal-rank LoRA and dense down-projection fine-tuning.

\subsection{Schedule for scaling \penaltyname{}}
\label{sec:weight-scheduler}

The mixed-domain experiments above apply the penalty from the first step. Here we ask what the start step changes, using a short single-task run in which the early phase is a large share of training.
Table~\ref{tab:weight-scheduler-ablations} reports a scheduling ablation on SFT of Qwen2.5-0.5B on the Aya Hausa instruction-following dataset for ${\sim}500$ optimizer steps (${\sim}3\times 10^5$ tokens).
We fix the terminal penalty weight to $\lambda_1{=}0.1$ and a linear ramp of length $W{=}200$ steps, and vary the start step $S$ at which the \penaltyname{} is first applied to the language-modeling loss.%

\paragraph{Retention improves whenever the penalty is used.}
Across all choices of $S$, adding the penalty reduces forgetting on the pretraining proxies: StarCoder and Wiki perplexity are lower than for standard LoRA ($4.05$ and $21.8$) in every penalized row.
The strongest retention is at early starts (e.g., $S{=}50$: StarCoder $3.75$, Wiki $18.7$), while $S{=}400$ still improves over standard LoRA ($3.86$ and $19.8$) despite engaging the penalty only in the final segment of training.

\paragraph{Earlier penalty activation harms fine-tuning.}
In contrast, Train PPL worsens as the penalty is applied earlier: $22.6$ (no penalty) $\rightarrow$ $24.7$ at $S{=}0$, with intermediate values at $S{=}50$ and $S{=}100$ ($23.9$ and $23.7$).
Only a late start ($S{=}400$) recovers baseline-level Train PPL ($22.4$).
The same trend appears in $\erank(\Delta W)$: early starts yield effective ranks of approximately $785$--$838$, whereas $S{=}400$ yields $\erank=430$, only modestly above unpenalized LoRA ($341$).
On this short budget, $S{=}400$ also does not complete the ramp, so the run ends with effective weight ${\approx}0.05$---yet retention still improves.

\paragraph{Why timing matters.}
Two mechanisms may help explain the training--retention trade-off.
First, \emph{linear mode connectivity} \citep{frankle2020linearmodeconnectivitylottery}: after a sufficient number of steps, the model may enter a basin of the language-modeling loss in which the solution is comparatively robust to perturbations.
Applying the penalty in that late phase reshapes the adapter spectrum without derailing Hausa fit.
Activating it from the outset interferes with the initial trajectory into that basin and inflates Train PPL even when forgetting is minimized.

Second, there is a more direct, penalty-specific effect.
Let $R(\Delta W)=\rerank(\Delta W)$ denote the quantity we encourage in the regularizer.
The Frobenius norm $\snorm{\nabla R(\Delta W)}_F$ can be large when $\snorm{\Delta W}_F$ is small: in the closed-form gradient (Appendix~\ref{apdx:gradient-of-renyi-2-penalty}, Equation~\eqref{eq:renyi-2-gradient}), the dominant denominators involve $\Tr((\Delta W^\top \Delta W)^2)$, which vanishes as $\Delta W\to 0$.
For the implemented logarithmic penalty $-\log R(\Delta W)$, the gradient is $-\nabla R/R$; since $1\le R\le r$, the same small-norm behavior carries over.
Early in LoRA training the update is near the origin, so the rank penalty can produce disproportionately large gradient components and compete aggressively with the SFT objective.
Once $\Delta W$ has grown, the same formula yields more modest penalty gradients, so a late start both respects optimization geometry and avoids the small-norm regime where $R$ is stiff.

\paragraph{Takeaway.}
Start time is a second knob alongside $\lambda$: an early start maximizes retention and a late start protects fit. The mixed-domain results use the early setting, so their Train PPL cost at $\lambda=0.1$ (Table \ref{tab:arxiv-curated-qwen-2p5-0p5b-rank-sweep}) may be reducible with a later start; we did not rerun those sweeps.
For task-first training, we recommend delaying the penalty until after an initial period of task training ($S$ scaled with total training length), rather than applying full weight from initialization.

\begin{table}[t]
\centering
\small
\setlength{\tabcolsep}{2pt}
\setlength{\tabcolsep}{4pt}
\begin{adjustbox}{max width=\columnwidth}
\begin{tabular}{@{}lrrrr@{}}
\toprule
\shortstack{Start step\\(max: 500)} & \shortstack{eRank\\($\uparrow$)} & \shortstack{Train PPL\\($\downarrow$)} & \shortstack{StarCoder PPL\\($\downarrow$)} & \shortstack{Wiki PPL\\($\downarrow$)} \\ \midrule
\textit{Standard LoRA} & 341.397 & 22.604 & 4.054 & 21.822 \\
\midrule
\multicolumn{5}{l}{\textit{\penaltyname} $\lambda=0.1$} \\
0 & 838.185 & 24.737 & 3.792 & 18.917 \\
50 & 809.978 & 23.860 & 3.753 & 18.700 \\
100 & 785.069 & 23.651 & 3.756 & 18.824 \\
400 & 429.841 & 22.410 & 3.861 & 19.768 \\
\bottomrule
\end{tabular}
\end{adjustbox}
\caption{R\'enyi-2 penalty scheduling on Aya Hausa instruction following. The terminal weight is $\lambda_1=0.1$, the linear ramp width is $W=200$, and training lasts approximately $500$ optimizer steps.
}
\label{tab:weight-scheduler-ablations}
\end{table}

\section{Singular Direction Analysis}
\label{sec:analysis}

\noindent
We analyze \emph{how} different fine-tuning methods couple to the pretrained solution at the level of singular directions and subspaces. For each of three MLP \texttt{down\_proj} layers (early, middle, late), we compute full right singular bases
\footnote{
    Full-basis diagnostics depend on the arbitrary choice of nullspace bases, including weighted unaligned mass through its maximum-overlap criterion. We interpret them descriptively for the computed SVD bases.
}
for the pretrained and finetuned weights and form the energy overlap
    $A_{ij} \;=\; \cos^2\!\big(v_{\mathrm{ft},i}, v_{p,j}\big)$.
Fixing $\epsilon=0.9$, we call a finetuned direction $i$ \emph{aligned} if it has at least one strong pretrained partner, $\max_j A_{ij}\ge \epsilon^2$, and \emph{unaligned} otherwise. We report both a \emph{count} view (how many indices lack a strong partner) and a \emph{mass} view, weighting each direction by an L1-normalized spectral weight $p_i\propto\sigma_i$ on finetuned singular values:
    $\text{WeightedUnaligned} \;=\; \sum_i p_i\,\mathbf{1}\!\left[\max_j A_{ij}<\epsilon^2\right]$.
Unlike ``intruder dimension'' analysis of prior work~\cite{shuttleworth2024loraillusion}, this diagnostic is deliberately ``full ambient''. These unaligned-direction metrics are local: each finetuned direction is assessed by its strongest match to the pretrained basis. To probe global structure, we reuse the same overlap matrix $A$ to form the positive kernel $K_{ij}=\exp(\beta A_{ij})$, then obtain the doubly stochastic coupling $B_\beta=\mathrm{Sinkhorn}(K)\in\mathcal{B}_d$,
where $\mathcal B_d$ is the set of nonnegative $d\times d$ matrices whose rows and columns each sum to one. This coupling can be interpreted as a soft permutation between finetuned and pretrained singular indices~\citep{sinkhorn1964relationship,mena2018learning} (we use $\beta=10$). We summarize $B_\beta$ by the mean row entropy normalized by $\log d$:
\begin{equation}
    H_{\mathrm{soft}}=-\frac{1}{d\log d}\sum_{i,j}B_{\beta,ij}\log B_{\beta,ij}.
\end{equation}
Lower values indicate a sharper, more permutation-like global matching. We use natural logarithms; the implementation clamps their arguments below at $10^{-12}$ for numerical stability. These quantities for the mixed-domain continual fine-tuning are reported in Table~\ref{tab:mixed-intruder-complement} for various fine-tuning strategies.
\begin{table}[t]
\centering
\small
\setlength{\tabcolsep}{2pt}
\begin{adjustbox}{max width=\columnwidth}
\begin{tabular}{@{}lrrr@{}}
\toprule
Method & \shortstack{Weighted\\unaligned ($\downarrow$)} & \shortstack{Unaligned\\count ($\downarrow$)} & \shortstack{$H_{\mathrm{soft}}$\\($\downarrow$)} \\
\midrule
\multicolumn{4}{l}{\textit{Standard}} \\
Dense FT & 0.940 & 4696.300 & 0.900 \\
Standard LoRA & 0.960 & 4725.700 & 0.910 \\
\midrule
\multicolumn{4}{l}{\textit{\penaltyname{} (Ours)}} \\
R\'enyi-2 penalty weight = 0.01 & 0.972 & 4728.700 & 0.910 \\
R\'enyi-2 penalty weight = 0.1 & 0.973 & 4734 & 0.910 \\
R\'enyi-2 penalty weight = 1.0 & 0.973 & 4729 & 0.910 \\
\midrule
\multicolumn{4}{l}{\textit{LoRA variants}} \\
PiSSA & 0.968 & 4721 & 0.910 \\
VeRA & 0 & 1 & 0.340 \\
TinyLoRA & 0.018 & 334.300 & 0.520 \\
LoRA-XS & 0.991 & 4731.700 & 0.910 \\
SVF & 0.001 & 3822 & 0.780 \\
\bottomrule
\end{tabular}
\end{adjustbox}
\caption{Dense FT denotes dense down-projection fine-tuning. Spectral mismatch between finetuned and pretrained weights in the mixed-domain setting for Qwen2.5-0.5B. Entries are averaged over the three analyzed MLP down-projection layers, using alignment threshold $\epsilon=0.9$.
    }
\label{tab:mixed-intruder-complement}
\end{table}

Three qualitative regimes emerge. First, VeRA and TinyLoRA exhibit near-saturated coupling to the pretrained spectrum: weighted unaligned mass is near zero, and the unaligned count is also small (especially for VeRA), with small $H_{\mathrm{soft}}$. Second, a broad LoRA-like band—including Standard LoRA, \penaltyname{}, PiSSA, LoRA-XS, and (to a slightly lesser extent) dense down-projection fine-tuning—leaves $\gtrsim 96\%$ of directions unaligned at this strict threshold. In this regime, \emph{count-based} comparisons are nearly flat, but \emph{where the spectral mass lands} still differs: LoRA-XS places the largest fraction of spectral weight on directions without a strong pretrained partner, consistent with its adaptation being carried by dominant singular directions that are more decoupled from pretraining. Third, SVF with an intermediate $H_{\mathrm{soft}}$ illustrates that count and mass can decouple sharply: it has near-zero weighted unaligned mass but a much lower unaligned rate than the LoRA-like band, indicating that SVF concentrates most of the finetuned matrix's singular-value mass on a small set of strongly pretrained-like directions while leaving a long tail of low-weight directions without strong partners.

In these experiments, fine-tuning produces many directions without a strong match to the pretrained singular basis; methods differ in how much singular-value mass these directions carry.

\section{Related Work}
\citet{roy2007effective}'s effective rank summarizes how uniformly singular-value mass is distributed; we build on their definition and on R\'enyi entropies~\cite{renyi1961measures,cover2006elements} to obtain a unitary-invariant surrogate ($\rerank$) suitable for gradient-based training.
To our knowledge, prior PEFT work has not targeted \emph{maximizing realized adapter rank during language-model fine-tuning} as a means to improve the learning--forgetting frontier; rank-stabilized scaling~\cite{kalajdzievski2023rslora} and spectral initializations~\cite{pissa,milora2024} address adjacent but distinct failure modes.

A large body of prior work studies \emph{how} to configure LoRA---rank and scaling~\cite{raschka2023lora,kalajdzievski2023rslora}, which modules to target~\cite{schulman2025lora}, and whether LoRA can match full fine-tuning in post-training regimes~\cite{schulman2025lora,biderman2024lora}.
These works largely treat \emph{configured rank} $r$ or trainable parameter count as mere hyperparameters.
Orthogonal to utilization, PEFT methods also differ in their trainable \emph{budget} and in how the update is tied to the pretrained spectrum.
VeRA~\cite{vera}, Singular Value fine-tuning~\cite{singularvaluefinetuning}, and TinyLoRA~\cite{tinylora} restrict trainable degrees of freedom by several orders of magnitude.
LoRA-XS~\cite{lora-xs} also reduces the budget, training only an $r\times r$ core inside a frozen truncated-SVD subspace ($1.9\times10^{7}$ versus $1.2\times10^{8}$ parameters at $r{=}896$ in our setup), yet still admits a rank-$r$ update.
PiSSA~\cite{pissa} keeps LoRA's budget and instead changes the initialization to the principal singular directions of the pretrained weight.
We benchmark against all three regimes in Section~\ref{sec:mixed-task-results}; our penalty is complementary in that it improves utilization \emph{within} a fixed parameterization rather than reallocating the budget itself.

Fine-tuning improves target-task metrics but can erase competence on broad pretraining domains, and heterogeneous or sequential training amplifies the tension~\cite{kirkpatrick2017overcoming}.
\citet{biderman2024lora} systematically compare LoRA and full fine-tuning, showing that low-rank adapters typically \emph{learn less} on the target domain but also \emph{forget less} off it, and that full fine-tuning's effective update rank can exceed typical LoRA configurations by an order of magnitude.
Norm- and magnitude-aware variants such as DoRA~\cite{liu2024dora}, DeLoRA~\cite{bini2025delora}, and Norm-Bounded LoRA~\cite{wang2025normboundedlowrankadaptation} pursue a related goal---making directional updates behave more like full fine-tuning~\cite{aghajanyan2020intrinsicdimensionalityexplainseffectiveness}---but through reparameterization rather than explicit control of spectral utilization.
Continued-pretraining methods such as ReLoRA~\cite{lialin2023relora} accumulate high-rank structure by merging and restarting low-rank updates over long horizons; we instead ask how to use a \emph{single} adapter's rank budget more fully during standard SFT.

A growing line of work engineers fine-tuning through the singular structure of weight matrices.
PiSSA~\cite{pissa} initializes adapters from dominant singular components; MiLoRA~\cite{milora2024} instead adapts \emph{minor} components to reduce interference with pretrained knowledge; AdaLoRA~\cite{adalora} adaptively reallocates rank across modules; and SVF~\cite{singularvaluefinetuning} trains singular values directly.
These methods change \emph{which} subspace or spectrum is updated, often at the cost of an explicit or implicit SVD.
Our \penaltyname{} is orthogonal and complementary to these approaches.

\section{Conclusion}

Standard LoRA training often underutilizes its allocated subspace
so nominal rank is an unreliable proxy for usable adapter capacity in the learning--forgetting trade-off.
We proposed effective-rank maximization via the \penaltyname{}, a parameterization-agnostic, SVD-free term added to the language-modeling loss, and showed across mixed-domain SFT, task-specific SFT, rank sweeps, and scheduling ablations that it can reduce forgetting on pretraining proxies while maintaining competitive target-task fit.
In mixed-domain rank sweeps on three models, $\lambda=0.01$
lowers mean Wiki perplexity at every tested rank of at least
$16$ relative to unpenalized LoRA, with mean Train PPL improving
or increasing by at most $0.02$ in all but one comparison.
Our single-task and scheduling experiments also demonstrate
configuration-dependent learning and retention outcomes:
the penalty can improve retention without a large training-fit
cost, but does not eliminate the learning--forgetting trade-off.
Spectral diagnostics provide descriptive comparisons of the resulting updates.
Effective rank is therefore not merely a post-hoc diagnostic but a trainable objective that regularizes toward fuller utilization.

\section*{Limitations}
While capacity maximization improves the learning--forgetting frontier in several configurations, benefits depend on the metric and execution, and the method still requires task-specific tuning of $\lambda$ and may not be beneficial when the goal is deliberate underfitting or strong retention of a very narrow domain. The present study focuses on LoRA-style adapters and closely related spectral variants; extending the same objective to other PEFT families or to full fine-tuning may require different regularization schedules or parameterizations.

\section*{Ethical Considerations}
The paper studies general-purpose language-model adaptation on public benchmark data and does not introduce new user data collection. The main ethical risk is indirect: methods that improve adaptation efficiency can be repurposed to more rapidly specialize models for harmful or deceptive uses. We therefore frame the contribution as an optimization and analysis tool, not as a deployment recommendation, and recommend the usual safeguards for downstream release, including dataset provenance checks, benchmark leakage review, and task-appropriate human oversight.

\bibliography{sections/backmatter/bibliography/include}

\begin{thebibliography}{33}
\providecommand{\natexlab}[1]{#1}

\bibitem[{Absil et~al.(2009)Absil, Mahony, and
  Sepulchre}]{absil2009optimization}
P.-A. Absil, Robert Mahony, and Rodolphe Sepulchre. 2009.
\newblock \href {https://doi.org/10.1515/9781400830244} {\emph{Optimization
  Algorithms on Matrix Manifolds}}.
\newblock Princeton University Press, Princeton.

\bibitem[{Aghajanyan et~al.(2020)Aghajanyan, Zettlemoyer, and
  Gupta}]{aghajanyan2020intrinsicdimensionalityexplainseffectiveness}
Armen Aghajanyan, Luke Zettlemoyer, and Sonal Gupta. 2020.
\newblock \href {https://arxiv.org/abs/2012.13255} {Intrinsic dimensionality
  explains the effectiveness of language model fine-tuning}.
\newblock \emph{Preprint}, arXiv:2012.13255.

\bibitem[{Barrault et~al.(2019)Barrault, Bojar, Costa-juss{\`a}, Federmann,
  Fishel, Graham, Haddow, Huck, Koehn, Malmasi, Monz, M{\"u}ller, Pal, Post,
  and Zampieri}]{wmt19translate}
Lo{\"i}c Barrault, Ond{\v{r}}ej Bojar, Marta~R. Costa-juss{\`a}, Christian
  Federmann, Mark Fishel, Yvette Graham, Barry Haddow, Matthias Huck, Philipp
  Koehn, Shervin Malmasi, Christof Monz, Mathias M{\"u}ller, Santanu Pal, Matt
  Post, and Marcos Zampieri. 2019.
\newblock \href {https://doi.org/10.18653/v1/W19-5301} {Findings of the 2019
  conference on machine translation ({WMT}19)}.
\newblock In \emph{Proceedings of the Fourth Conference on Machine Translation
  (Volume 2: Shared Task Papers, Day 1)}, pages 1--61, Florence, Italy.
  Association for Computational Linguistics.

\bibitem[{Bałazy et~al.(2025)Bałazy, Banaei, Aberer, and Tabor}]{lora-xs}
Klaudia Bałazy, Mohammadreza Banaei, Karl Aberer, and Jacek Tabor. 2025.
\newblock \href {https://arxiv.org/abs/2405.17604} {{LoRA-XS}: Low-rank
  adaptation with extremely small number of parameters}.
\newblock \emph{Preprint}, arXiv:2405.17604.

\bibitem[{Biderman et~al.(2024)Biderman, Portes, Ortiz, Paul, Greengard,
  Jennings, King, Havens, Chiley, Frankle, Blakeney, and
  Cunningham}]{biderman2024lora}
Dan Biderman, Jacob Portes, Jose Javier~Gonzalez Ortiz, Mansheej Paul, Philip
  Greengard, Connor Jennings, Daniel King, Sam Havens, Vitaliy Chiley, Jonathan
  Frankle, Cody Blakeney, and John~Patrick Cunningham. 2024.
\newblock \href {https://openreview.net/forum?id=aloEru2qCG} {{LoRA} learns
  less and forgets less}.
\newblock \emph{Transactions on Machine Learning Research}.

\bibitem[{Bini et~al.(2025)Bini, Girrbach, and Akata}]{bini2025delora}
Massimo Bini, Leander Girrbach, and Zeynep Akata. 2025.
\newblock \href {https://arxiv.org/abs/2503.18225} {{DeLoRA}: Decoupling angles
  and strength in low-rank adaptation}.
\newblock \emph{Preprint}, arXiv:2503.18225.

\bibitem[{Cobbe et~al.(2021)Cobbe, Kosaraju, Bavarian, Chen, Jun, Kaiser,
  Plappert, Tworek, Hilton, Nakano et~al.}]{cobbe2021training}
Karl Cobbe, Vineet Kosaraju, Mohammad Bavarian, Mark Chen, Heewoo Jun, Lukasz
  Kaiser, Matthias Plappert, Jerry Tworek, Jacob Hilton, Reiichiro Nakano,
  et~al. 2021.
\newblock Training verifiers to solve math word problems.
\newblock \emph{arXiv preprint arXiv:2110.14168}.

\bibitem[{Cover and Thomas(2006)}]{cover2006elements}
Thomas~M. Cover and Joy~A. Thomas. 2006.
\newblock \emph{Elements of information theory}, 2nd edition.
\newblock Wiley-Interscience.

\bibitem[{Frankle et~al.(2020)Frankle, Dziugaite, Roy, and
  Carbin}]{frankle2020linearmodeconnectivitylottery}
Jonathan Frankle, Gintare~Karolina Dziugaite, Daniel~M. Roy, and Michael
  Carbin. 2020.
\newblock \href {https://arxiv.org/abs/1912.05671} {Linear mode connectivity
  and the lottery ticket hypothesis}.
\newblock \emph{Preprint}, arXiv:1912.05671.

\bibitem[{Hu et~al.(2022)Hu, Shen, Wallis, Allen-Zhu, Li, Wang, Wang, and
  Chen}]{hu2022lora}
Edward~J. Hu, Yelong Shen, Phillip Wallis, Zeyuan Allen-Zhu, Yuanzhi Li, Shean
  Wang, Lu~Wang, and Weizhu Chen. 2022.
\newblock \href {https://openreview.net/forum?id=nZeVKeeFYf9} {{LoRA}: Low-rank
  adaptation of large language models}.
\newblock In \emph{International Conference on Learning Representations}.

\bibitem[{Kalajdzievski(2023)}]{kalajdzievski2023rslora}
Damjan Kalajdzievski. 2023.
\newblock \href {https://arxiv.org/abs/2312.03732} {A rank stabilization
  scaling factor for fine-tuning with {LoRA}}.
\newblock \emph{Preprint}, arXiv:2312.03732.

\bibitem[{Kirkpatrick et~al.(2017)Kirkpatrick, Pascanu, Rabinowitz, Veness,
  Desjardins, Rusu, Milan, Quan, Ramalho, Grabska-Barwinska, Hassabis, Clopath,
  Kumaran, and Hadsell}]{kirkpatrick2017overcoming}
James Kirkpatrick, Razvan Pascanu, Neil Rabinowitz, Joel Veness, Guillaume
  Desjardins, Andrei~A. Rusu, Kieran Milan, John Quan, Tiago Ramalho, Agnieszka
  Grabska-Barwinska, Demis Hassabis, Claudia Clopath, Dharshan Kumaran, and
  Raia Hadsell. 2017.
\newblock \href {https://doi.org/10.1073/pnas.1611835114} {Overcoming
  catastrophic forgetting in neural networks}.
\newblock \emph{Proceedings of the National Academy of Sciences},
  114(13):3521--3526.

\bibitem[{Kopiczko et~al.(2024)Kopiczko, Blankevoort, and Asano}]{vera}
Dawid~J. Kopiczko, Tijmen Blankevoort, and Yuki~M. Asano. 2024.
\newblock \href {https://arxiv.org/abs/2310.11454} {{VeRA}: Vector-based random
  matrix adaptation}.
\newblock \emph{Preprint}, arXiv:2310.11454.

\bibitem[{Lialin et~al.(2023)Lialin, Muckatira, Shivagunde, and
  Rumshisky}]{lialin2023relora}
Vladislav Lialin, Sherin Muckatira, Namrata Shivagunde, and Anna Rumshisky.
  2023.
\newblock \href {https://arxiv.org/abs/2307.05695} {{ReLoRA}: High-rank
  training through low-rank updates}.
\newblock \emph{Preprint}, arXiv:2307.05695.

\bibitem[{Liu et~al.(2024{\natexlab{a}})Liu, Wang, Yin, Molchanov, Wang, Cheng,
  and Chen}]{liu2024dora}
Shih-Yang Liu, Chien-Yi Wang, Hongxu Yin, Pavlo Molchanov, Yu-Chiang~Frank
  Wang, Kwang-Ting Cheng, and Min-Hung Chen. 2024{\natexlab{a}}.
\newblock \href {https://proceedings.mlr.press/v235/liu24bn.html} {{DoRA}:
  Weight-decomposed low-rank adaptation}.
\newblock In \emph{Proceedings of the 41st International Conference on Machine
  Learning}, pages 32100--32121.

\bibitem[{Liu et~al.(2024{\natexlab{b}})Liu, Lyn, Zhu, Tian, and
  Graham}]{liu-etal-2024-alora}
Zequan Liu, Jiawen Lyn, Wei Zhu, Xing Tian, and Yvette Graham.
  2024{\natexlab{b}}.
\newblock \href {https://doi.org/10.18653/v1/2024.naacl-long.35} {{AL}o{RA}:
  Allocating low-rank adaptation for fine-tuning large language models}.
\newblock In \emph{Proceedings of the 2024 Conference of the North American
  Chapter of the Association for Computational Linguistics: Human Language
  Technologies (Volume 1: Long Papers)}, pages 622--641, Mexico City, Mexico.
  Association for Computational Linguistics.

\bibitem[{Mena et~al.(2018)Mena, Belanger, Linderman, and
  Snoek}]{mena2018learning}
Gonzalo Mena, David Belanger, Scott Linderman, and Jasper Snoek. 2018.
\newblock \href {https://openreview.net/forum?id=Byt3oJ-0W} {Learning latent
  permutations with {Gumbel--Sinkhorn} networks}.
\newblock In \emph{International Conference on Learning Representations}.

\bibitem[{Meng et~al.(2025)Meng, Wang, and Zhang}]{pissa}
Fanxu Meng, Zhaohui Wang, and Muhan Zhang. 2025.
\newblock \href {https://arxiv.org/abs/2404.02948} {{PiSSA}: Principal singular
  values and singular vectors adaptation of large language models}.
\newblock \emph{Preprint}, arXiv:2404.02948.

\bibitem[{Morris et~al.(2026)Morris, Mireshghallah, Ibrahim, and
  Mahloujifar}]{tinylora}
John~X. Morris, Niloofar Mireshghallah, Mark Ibrahim, and Saeed Mahloujifar.
  2026.
\newblock \href {https://arxiv.org/abs/2602.04118} {Learning to reason in 13
  parameters}.
\newblock \emph{Preprint}, arXiv:2602.04118.

\bibitem[{OLMo et~al.(2024)OLMo, Walsh, Soldaini, Groeneveld, Lo, Arora,
  Bhagia, Gu, Huang, Jordan, Lambert, Schwenk, Tafjord, Anderson, Atkinson,
  Brahman, Clark, Dasigi, Dziri, Guerquin, Ivison, Koh, Liu, Malik, Merrill,
  Miranda, Morrison, Murray, Nam, Pyatkin, Rangapur, Schmitz, Skjonsberg,
  Wadden, Wilhelm, Wilson, Zettlemoyer, Farhadi, Smith, and
  Hajishirzi}]{olmo20242olmo2furious}
Team OLMo, Pete Walsh, Luca Soldaini, Dirk Groeneveld, Kyle Lo, Shane Arora,
  Akshita Bhagia, Yuling Gu, Shengyi Huang, Matt Jordan, Nathan Lambert, Dustin
  Schwenk, Oyvind Tafjord, Taira Anderson, David Atkinson, Faeze Brahman,
  Christopher Clark, Pradeep Dasigi, Nouha Dziri, and 21 others. 2024.
\newblock \href {https://arxiv.org/abs/2501.00656} {2 {OLMo} 2 furious}.
\newblock \emph{Preprint}, arXiv:2501.00656.

\bibitem[{Raschka(2023)}]{raschka2023lora}
Sebastian Raschka. 2023.
\newblock \href
  {https://magazine.sebastianraschka.com/p/practical-tips-for-finetuning-llms}
  {Practical tips for finetuning {LLMs} using {LoRA} (low-rank adaptation)}.
\newblock Ahead of AI / Raschka AI Research Lab.

\bibitem[{R{\'e}nyi(1961)}]{renyi1961measures}
Alfr{\'e}d R{\'e}nyi. 1961.
\newblock On measures of entropy and information.
\newblock In \emph{Proceedings of the Fourth Berkeley Symposium on Mathematical
  Statistics and Probability, Volume 1: Contributions to the Theory of
  Statistics}, pages 547--561. University of California Press, Berkeley, Calif.

\bibitem[{Roy and Vetterli(2007)}]{roy2007effective}
Olivier Roy and Martin Vetterli. 2007.
\newblock The effective rank: A measure of effective dimensionality.
\newblock In \emph{2007 15th European signal processing conference}, pages
  606--610. IEEE.

\bibitem[{Schulman and {Thinking Machines Lab}(2025)}]{schulman2025lora}
John Schulman and {Thinking Machines Lab}. 2025.
\newblock \href {https://doi.org/10.64434/tml.20250929} {{LoRA} without
  regret}.
\newblock \emph{Thinking Machines Lab: Connectionism}.
\newblock September 29, 2025.

\bibitem[{Shuttleworth et~al.(2024)Shuttleworth, Andreas, Torralba, and
  Sharma}]{shuttleworth2024loraillusion}
Reece Shuttleworth, Jacob Andreas, Antonio Torralba, and Pratyusha Sharma.
  2024.
\newblock \href {https://arxiv.org/abs/2410.21228} {{LoRA} vs full fine-tuning:
  An illusion of equivalence}.
\newblock \emph{Preprint}, arXiv:2410.21228.

\bibitem[{Singh et~al.(2024)Singh, Vargus, D’souza, Karlsson, Mahendiran, Ko,
  Shandilya, Patel, Mataciunas, O’Mahony et~al.}]{singh2024aya}
Shivalika Singh, Freddie Vargus, Daniel D’souza, B{\"o}rje~F Karlsson,
  Abinaya Mahendiran, Wei-Yin Ko, Herumb Shandilya, Jay Patel, Deividas
  Mataciunas, Laura O’Mahony, et~al. 2024.
\newblock Aya dataset: An open-access collection for multilingual instruction
  tuning.
\newblock In \emph{Proceedings of the 62nd Annual Meeting of the Association
  for Computational Linguistics (Volume 1: Long Papers)}, pages 11521--11567.

\bibitem[{Sinkhorn(1964)}]{sinkhorn1964relationship}
Richard Sinkhorn. 1964.
\newblock A relationship between arbitrary positive matrices and doubly
  stochastic matrices.
\newblock \emph{The Annals of Mathematical Statistics}, 35(2):876--879.

\bibitem[{Sun et~al.(2022)Sun, Chen, He, Wang, Feng, Han, Ding, Cheng, Li, and
  Wang}]{singularvaluefinetuning}
Yanpeng Sun, Qiang Chen, Xiangyu He, Jian Wang, Haocheng Feng, Junyu Han, Errui
  Ding, Jian Cheng, Zechao Li, and Jingdong Wang. 2022.
\newblock \href {https://arxiv.org/abs/2206.06122} {Singular value fine-tuning:
  Few-shot segmentation requires few-parameters fine-tuning}.
\newblock \emph{Preprint}, arXiv:2206.06122.

\bibitem[{Trockman and Kolter(2021)}]{trockman2021orthogonalizing}
Asher Trockman and J.~Zico Kolter. 2021.
\newblock \href {https://arxiv.org/abs/2104.07167} {Orthogonalizing
  convolutional layers with the {Cayley} transform}.
\newblock \emph{arXiv preprint arXiv:2104.07167}.

\bibitem[{Wang et~al.(2024)Wang, Li, Wang, Chen, and Chen}]{milora2024}
Hanqing Wang, Yixia Li, Shuo Wang, Guanhua Chen, and Yun Chen. 2024.
\newblock \href {https://arxiv.org/abs/2406.09044} {{MiLoRA}: Harnessing minor
  singular components for parameter-efficient {LLM} finetuning}.
\newblock \emph{Preprint}, arXiv:2406.09044.

\bibitem[{Wang et~al.(2025)Wang, Dvijotham, and
  Manchester}]{wang2025normboundedlowrankadaptation}
Ruigang Wang, Krishnamurthy Dvijotham, and Ian~R. Manchester. 2025.
\newblock \href {https://arxiv.org/abs/2501.19050} {Norm-bounded low-rank
  adaptation}.
\newblock \emph{Preprint}, arXiv:2501.19050.

\bibitem[{Yang et~al.(2025)Yang, Yang, Zhang, Hui, Zheng, Yu, Li, Liu, Huang,
  Wei, Lin, Yang, Tu, Zhang, Yang, Yang, Zhou, Lin, Dang, Lu, Bao, Yang, Yu,
  Li, Xue, Zhang, Zhu, Men, Lin, Li, Tang, Xia, Ren, Ren, Fan, Su, Zhang, Wan,
  Liu, Cui, Zhang, and Qiu}]{qwen}
An~Yang, Baosong Yang, Beichen Zhang, Binyuan Hui, Bo~Zheng, Bowen Yu,
  Chengyuan Li, Dayiheng Liu, Fei Huang, Haoran Wei, Huan Lin, Jian Yang,
  Jianhong Tu, Jianwei Zhang, Jianxin Yang, Jiaxi Yang, Jingren Zhou, Junyang
  Lin, Kai Dang, and 23 others. 2025.
\newblock \href {https://arxiv.org/abs/2412.15115} {Qwen2.5 technical report}.
\newblock \emph{Preprint}, arXiv:2412.15115.

\bibitem[{Zhang et~al.(2023)Zhang, Chen, Bukharin, Karampatziakis, He, Cheng,
  Chen, and Zhao}]{adalora}
Qingru Zhang, Minshuo Chen, Alexander Bukharin, Nikos Karampatziakis, Pengcheng
  He, Yu~Cheng, Weizhu Chen, and Tuo Zhao. 2023.
\newblock \href {https://arxiv.org/abs/2303.10512} {{AdaLoRA}: Adaptive budget
  allocation for parameter-efficient fine-tuning}.
\newblock \emph{Preprint}, arXiv:2303.10512.

\end{thebibliography}

\appendix

\section{Proof of Theorem~\ref{thm:renyi-2-bound-via-schatten}}
\label{apdx:renyi-2-bound-via-schatten}

We bound the R\'enyi-2 entropy $H_2(\mathbf q)$ of the energy distribution $\mathbf q$ when the adapter satisfies an operator-norm cap.
NB-LoRA enforces this hypothesis via $\snorm{\Delta W}_{S_{\infty}} \le \delta$ (Section~\ref{sec:methods}); the argument below is parameterization-agnostic.

\begin{proof}[Proof of Theorem~\ref{thm:renyi-2-bound-via-schatten}]
    Assume $\Delta W\ne 0$ and $\delta>0$. Let $\sigma_1,\ldots,\sigma_r$ be the singular values of $\Delta W$.
    If $\snorm{\Delta W}_{S_{\infty}} \le \delta$, then $\sigma_i \le \delta$ for all $i$, hence $\sigma_i^4 \le \delta^2 \sigma_i^2$ and
    \begin{equation}
        \begin{aligned}
        \snorm{\Delta W}_{S_4}^4
        &= \sum_i \sigma_i^4 \\
        &\le \delta^2 \sum_i \sigma_i^2
        = \delta^2 \snorm{\Delta W}_{S_2}^2.
        \end{aligned}
    \end{equation}
    Since
    \begin{equation}
        \begin{aligned}
        H_2(\mathbf q)
        &=
        -\log \sum_i q_i^2
        \\
        &=
        4\log \snorm{\Delta W}_{S_2}
        - \log \snorm{\Delta W}_{S_4}^4,
        \end{aligned}
    \end{equation}
    we obtain
    \[
        H_2(\mathbf q)
        \ge 2\log\!\left(\frac{\snorm{\Delta W}_{S_2}}{\delta}\right).
    \]
    Exponentiating and applying the entropy comparison proved in
    Appendix~\ref{apdx:renyi-2-bound-proof} gives
    \[
        \erank(\Delta W)
        \ge \exp(H_2(\mathbf q))
        \ge \frac{\snorm{\Delta W}_{S_2}^2}{\delta^2}.
    \]
\end{proof}

\section{Comparison of Effective Ranks}
\label{apdx:renyi-2-bound-proof}

Recall the amplitude distribution \(\mathbf p\) in Equation~\eqref{eq:normalized-singular-values} and the energy distribution \(\mathbf q\) introduced in Section~\ref{sec:methods},
with \(\erank(\Delta W)=\exp(H_1(\mathbf p))\) and
\(\rerank(\Delta W)=\exp(H_2(\mathbf q))\)
(Theorem~\ref{thm:renyi-2-bound}).
For a probability vector \(\mathbf r\), write the Rényi-2 (collision) entropy as
\(H_2(\mathbf r)=-\log\sum_i r_i^2\)~\cite{cover2006elements,renyi1961measures}.

\begin{proof}[Proof of $\erank(\Delta W)\ge\rerank(\Delta W)$ in Theorem~\ref{thm:renyi-2-bound}]
Assume $\Delta W\ne 0$, and let $\sigma$ be its vector of singular values.
R\'enyi entropies are nonincreasing in their order parameter~\cite{renyi1961measures}, so
$H_1(\mathbf p)\ge H_2(\mathbf p)$.
The collision probabilities are
\[
    \sum_i p_i^2=\frac{\|\sigma\|_2^2}{\|\sigma\|_1^2},
    \qquad
    \sum_i q_i^2=\frac{\|\sigma\|_4^4}{\|\sigma\|_2^4}.
\]
H\"older's inequality (with conjugate exponents $\frac{3}{2}$ and 3) gives
\begin{align*}
    \sum_i\sigma_i^2
    &=\sum_i\sigma_i^{2/3}\sigma_i^{4/3} \\
    &\le \left(\sum_i\sigma_i\right)^{2/3}
          \left(\sum_i\sigma_i^4\right)^{1/3}.
\end{align*}
Cubing and rearranging yields
\[
    \frac{\|\sigma\|_2^2}{\|\sigma\|_1^2}
    \le \frac{\|\sigma\|_4^4}{\|\sigma\|_2^4}.
\]
Taking negative logarithms therefore gives
\[
    H_1(\mathbf p)\ge H_2(\mathbf p)\ge H_2(\mathbf q).
\]
Exponentiating proves the claimed effective-rank inequality.
\end{proof}

\section{Basic Properties of \penaltyname}
\label{apdx:proof-of-basic-properties}

This appendix collects the elementary properties of the R\'enyi-2 effective rank
$\rerank(\Delta W)$ used in Section~\ref{sec:methods}.
The middle inequality $\erank(\Delta W)\ge \rerank(\Delta W)$ in
Theorem~\ref{thm:renyi-2-bound} is proved separately in
Appendix~\ref{apdx:renyi-2-bound-proof}; gradient and LoRA implementation details
appear in Appendices~\ref{apdx:gradient-of-renyi-2-penalty}
and~\ref{apdx:implementation-tricks}.

\paragraph{Setup.}
Let $\Delta W\ne 0$ with $\Delta W\in\R^{m\times n}$ have singular value decomposition
$\Delta W=U\Sigma V^\top$ with singular values
$\sigma_1\ge\cdots\ge\sigma_k>0$ and $\sigma_{k+1}=\cdots=0$, where
$k=\rank(\Delta W)$.
Write $\|\sigma\|_p=(\sum_{i=1}^k\sigma_i^p)^{1/p}$ for $p\ge 1$.
The amplitude and energy distributions are
\begin{align*}
    p_i&=\frac{\sigma_i}{\|\sigma\|_1},
    &q_i&=\frac{\sigma_i^2}{E}, \\
    E&=\sum_{i=1}^k\sigma_i^2
      =\snorm{\Delta W}_F^2.
\end{align*}
For a probability vector $\mathbf r$, the R\'enyi entropy is
\[
    H_\alpha(\mathbf r)
    =\frac{1}{1-\alpha}\log\sum_i r_i^\alpha,
    \qquad \alpha>0,\quad\alpha\ne 1.
\]
At order one we use Shannon entropy. In particular,
\begin{align*}
    H_2(\mathbf r)&=-\log\sum_i r_i^2, \\
    \erank(\Delta W)&=\exp(H_1(\mathbf p)), \\
    \rerank(\Delta W)&=\exp(H_2(\mathbf q)).
\end{align*}

\paragraph{Lemma 1 (Definition equivalence).}
The two expressions in the definition of $\rerank(\Delta W)$ agree:
\begin{equation}
    \begin{aligned}
    \rerank(\Delta W)
    &=
    \exp(H_2(\mathbf q))
    =
    \frac{\snorm{\Delta W}_F^4}{\Tr\bigl((\Delta W^\top \Delta W)^2\bigr)}
    \\
    &=
    \frac{\bigl(\sum_{i=1}^k \sigma_i^2\bigr)^2}
         {\sum_{i=1}^k \sigma_i^4}.
    \end{aligned}
\end{equation}

\begin{proof}
By definition of $\mathbf q$,
\begin{align*}
    \exp(H_2(\mathbf q))
    &=\left(\sum_i q_i^2\right)^{-1} \\
    &=\frac{E^2}{\sum_i\sigma_i^4}.
\end{align*}
The SVD and cyclicity of the trace give
\begin{align*}
    \Tr(\Delta W^\top\Delta W)&=\sum_i\sigma_i^2=E, \\
    \Tr\bigl((\Delta W^\top\Delta W)^2\bigr)&=\sum_i\sigma_i^4.
\end{align*}
Substitution yields both claimed formulas. Zero singular values do not
contribute to either sum.
\end{proof}

\paragraph{Lemma 2 (Unitary invariance).}
For orthogonal $U_1\in\R^{m\times m}$ and $V_1\in\R^{n\times n}$,
\[
\rerank(U_1\Delta W V_1)=\rerank(\Delta W).
\]
In particular,
\[
\rerank(\Delta W)=\rerank(\Sigma)=\bigl(\sum_i\sigma_i^2\bigr)^2/\sum_i\sigma_i^4.
\]

\begin{proof}
Unitary invariance of the Frobenius norm gives
$\snorm{U_1\Delta W V_1}_F=\snorm{\Delta W}_F$.
Let $X=\Delta W^\top\Delta W$ and $\widetilde X=V_1^\top X V_1$.
Then
\[
\Tr\bigl(\widetilde X^2\bigr)=\Tr(V_1^\top X V_1 V_1^\top X V_1)=\Tr(X^2),
\]
so both numerator and denominator in Lemma~1 are unchanged.
The singular-value form follows from Lemma~1 applied to $\Sigma$.
\end{proof}

Thus $\rerank(\Delta W)$ depends only on the multiset of singular values, not on the
left or right singular subspaces. The trace-ratio formula in Lemma~1 allows
evaluation without an SVD (Section~\ref{sec:methods}).

\paragraph{Lemma 3 (Bounds $1\le \rerank\le k$).}
With $k=\rank(\Delta W)$,
\[
1\le \rerank(\Delta W)\le k.
\]
Equality at the upper bound holds iff $\sigma_1^2=\cdots=\sigma_k^2$
(equivalently, $\sigma_1=\cdots=\sigma_k$ on the support).

\begin{proof}
Since $\mathbf q$ is a probability vector,
\[
\sum_i q_i^2 \le \bigl(\max_i q_i\bigr)\sum_i q_i \le 1,
\]
with equality iff one $q_i=1$.
Hence $H_2(\mathbf q)\ge 0$ and $\rerank(\Delta W)=\exp(H_2(\mathbf q))\ge 1$.

For the upper bound, by convexity of $x\mapsto x^2$ (or Cauchy--Schwarz),
\[
\sum_{i=1}^k q_i^2 \ge k\bigl(\tfrac{1}{k}\sum_i q_i\bigr)^2 = 1/k,
\]
with equality iff $q_1=\cdots=q_k=1/k$.
Therefore $H_2(\mathbf q)\le \log k$ and $\rerank(\Delta W)\le k$.
The equality condition is $q_i\equiv 1/k$, i.e.\ uniform energy
$\sigma_i^2\equiv E/k$.
\end{proof}

\paragraph{Lemma 4 ($\rank\ge \erank$).}
\[
\rank(\Delta W)\ge \erank(\Delta W),
\]
with equality iff $\sigma_1=\cdots=\sigma_k$ on the support.

\begin{proof}
Shannon entropy of a $k$-point distribution satisfies $0\le H_1(\mathbf p)\le \log k$,
so $1\le \erank(\Delta W)=\exp(H_1(\mathbf p))\le k=\rank(\Delta W)$.
Equality $H_1(\mathbf p)=\log k$ holds iff $\mathbf p$ is uniform, i.e.\
$\sigma_i$ are equal on the support~\cite{roy2007effective}.
\end{proof}

\paragraph{Lemma 5 (Equality cases in Theorem~\ref{thm:renyi-2-bound}).}
Let $k=\rank(\Delta W)$.

\textbf{(a) Uniform spectrum.}
The chain $\rank(\Delta W)=\erank(\Delta W)=\rerank(\Delta W)$ holds iff
$\sigma_1=\cdots=\sigma_k>0$.
In that case all three quantities equal $k$.

\textbf{(b) Rank-one energy.}
$\rerank(\Delta W)=1$ iff exactly one singular value is positive
(equivalently, $\mathbf q$ is a point mass).

\begin{proof}
For~(a), Lemma~4 gives $\rank=\erank=k$ iff $\mathbf p$ is uniform.
Lemma~3 gives $\rerank=k$ iff $\mathbf q$ is uniform.
For positive $\sigma_i$, $\mathbf p$ uniform iff $\mathbf q$ uniform iff
$\sigma_1=\cdots=\sigma_k$.

For~(b), $\rerank=1$ iff $H_2(\mathbf q)=0$ iff $\sum_i q_i^2=1$.
For a probability vector, $\sum_i q_i^2=1$ iff one $q_i=1$ and the rest are zero,
which is equivalent to a single positive $\sigma_i$.
\end{proof}

\paragraph{Lemma 6 ($H_1(\mathbf p)\ge H_2(\mathbf q)$).}
For the paired distributions $\mathbf p,\mathbf q$ from the same singular values,
\[
H_1(\mathbf p)\ge H_2(\mathbf q).
\]
Moreover, $H_2(\mathbf q)$ is maximized at the uniform distribution
$\mathbf q=(1/k,\ldots,1/k)$, with maximum value $\log k$.

\begin{proof}
The entropy comparison is proved in Appendix~\ref{apdx:renyi-2-bound-proof}.
The maximum and its equality condition follow from Lemma~3.
\end{proof}

Thus maximizing $H_2(\mathbf q)$ raises a lower bound on $H_1(\mathbf p)$.
It does not, by itself, guarantee that $H_1(\mathbf p)$ increases at every step.

\paragraph{Remark (Evaluation from trace statistics).}
Lemma~1 expresses $\rerank$ using only
\[
    \Tr(\Delta W^\top\Delta W)
    \quad\text{and}\quad
    \Tr\bigl((\Delta W^\top\Delta W)^2\bigr).
\]
In contrast, evaluating $\erank$ requires the normalized singular values.
The Gram matrix $\Delta W^\top\Delta W$ determines these values, but the two
trace statistics above generally do not suffice to recover them.
This is the computational distinction behind the choice of $\rerank$ in
\eqref{eq:penalized-loss}.

\paragraph{Corollary (LoRA rank ceiling).}
If $\Delta W=BA$ with $B\in\R^{m\times r}$ and $A\in\R^{r\times n}$, then
$\rank(\Delta W)\le r$ and hence $\rerank(\Delta W)\le r$.

\begin{proof}
$\rank(BA)\le\min\{\rank(B),\rank(A)\}\le r$.
Lemma~3 gives $\rerank(\Delta W)\le \rank(\Delta W)\le r$.
\end{proof}

When training drives $\rerank(\Delta W)\approx r$, the adapter uses nearly all of its
nominal rank budget with approximately uniform singular-value energy.

\begin{proof}[Proof of Theorem~\ref{thm:renyi-2-bound}]
Lemma~4 gives $\rank(\Delta W)\ge \erank(\Delta W)$.
Appendix~\ref{apdx:renyi-2-bound-proof} gives $\erank(\Delta W)\ge \rerank(\Delta W)$.
Lemma~3 gives $\rerank(\Delta W)\ge 1$.
The equality characterizations follow from Lemma~5.
\end{proof}

\paragraph{Related results deferred to other appendices.}
The gradient of $\rerank$ and its numerical stability relative to SVD-based
differentiation of $\erank$ are treated in
Appendix~\ref{apdx:gradient-of-renyi-2-penalty}.
Efficient evaluation from LoRA factors $A,B$ without forming $\Delta W$ is given in
Appendix~\ref{apdx:implementation-tricks}.
In the reported sweeps, increasing the penalty weight often increases the measured Roy--Vetterli effective rank. This is an empirical observation, not a consequence of the definition.

\section{Differentiating Effective Rank}
\label{apdx:direct-differentiation-of-erank}

\subsection{Differentiating the Roy--Vetterli effective rank}

This section distinguishes the sensitivity of singular vectors from the
sensitivity of a loss that depends only on singular values.
Let $X=\Delta W\in\mathbb R^{m\times n}$, with $m\ge n$, have thin SVD
$X=U\Sigma V^\top$. Initially assume that its singular values are positive
and distinct. Differentiating gives
\[
    dX=dU\Sigma V^\top+U\,d\Sigma V^\top+U\Sigma\,dV^\top.
\]
Define the reduced differentials
\[
    M=U^\top dX\,V,\qquad C=U^\top dU,\qquad D=V^\top dV.
\]
Differentiating $U^\top U=V^\top V=I$ shows that $C$ and $D$ are
skew-symmetric. Hence
\[
    M=C\Sigma+d\Sigma-\Sigma D.
\]
On the diagonal, $d\sigma_i=M_{ii}$. Off the diagonal,
\begin{align*}
    M_{ij}&=C_{ij}\sigma_j-\sigma_i D_{ij}, \\
    M_{ji}&=-C_{ij}\sigma_i+\sigma_j D_{ij}.
\end{align*}
Solving this system yields
\begin{align*}
    C_{ij}&=\frac{\sigma_j M_{ij}+\sigma_i M_{ji}}
                     {\sigma_j^2-\sigma_i^2}, \\
    D_{ij}&=\frac{\sigma_i M_{ij}+\sigma_j M_{ji}}
                     {\sigma_j^2-\sigma_i^2}.
\end{align*}
Thus differentiation of singular vectors involves inverse squared-singular-value
gaps. At repeated singular values the individual singular vectors are not unique.

\paragraph{Losses depending only on singular values.}
These gap denominators do not establish an intrinsic instability of every
spectral loss. If $L$ depends only on $\sigma$, then
\[
    \nabla_X L
    =U\,\operatorname{diag}\!\left(
        \frac{\partial L}{\partial\sigma_i}\right)V^\top.
\]
For the Roy--Vetterli effective rank, write
\begin{align*}
    S&=\sum_i\sigma_i, & p_i&=\frac{\sigma_i}{S}, \\
    H&=-\sum_i p_i\log p_i, & R&=\exp(H).
\end{align*}
Direct differentiation gives
\[
    \frac{\partial R}{\partial\sigma_i}
    =-\frac{R}{S}\bigl(\log p_i+H\bigr).
\]
At equal positive singular values, these derivatives are equal, so the matrix
gradient is independent of the choice of basis in the repeated singular subspace.
However, the logarithm becomes singular as a normalized singular value tends
to zero. Rank-deficient updates therefore require care. A generic SVD backward
implementation may also encounter difficulties at repeated singular values,
depending on how unused singular-vector derivatives are handled.

\subsection{Gradient of the R\'enyi-2 penalty}
\label{apdx:gradient-of-renyi-2-penalty}

For $X\ne 0$, the R\'enyi-2 effective rank can be evaluated directly:
\begin{equation}
    \rerank(X)=\frac{N(X)}{D(X)},\qquad
    \begin{aligned}
        N(X)&=\snorm{X}_F^4, \\
        D(X)&=\Tr\bigl((X^\top X)^2\bigr).
    \end{aligned}
\end{equation}
No singular-value decomposition is needed for this expression or its gradient.

\paragraph{Numerator.}
Using $\snorm{X}_F^2=\Tr(X^\top X)$ gives
\begin{align*}
    dN
    &=2\snorm{X}_F^2\,\Tr(dX^\top X+X^\top dX) \\
    &=4\snorm{X}_F^2\,\Tr(X^\top dX).
\end{align*}
Consequently, $\nabla_X N=4\snorm{X}_F^2X$.

\paragraph{Denominator.}
Expanding the matrix product gives
\begin{align*}
    dD
    &=\Tr(dX^\top XX^\top X)
      +\Tr(X^\top dX X^\top X) \\
    &\quad+\Tr(X^\top X dX^\top X)
      +\Tr(X^\top XX^\top dX).
\end{align*}
By cyclicity and invariance of the trace under transposition, all four terms
are equal. Thus
\[
    dD=4\Tr\bigl((XX^\top X)^\top dX\bigr),
    \qquad \nabla_X D=4XX^\top X.
\]

\paragraph{Quotient rule.}
Combining the two gradients yields
\begin{equation}\label{eq:renyi-2-gradient}
    \begin{aligned}
        \nabla_X\rerank(X)
        &=\frac{4\snorm{X}_F^2}{D(X)}X \\
        &\quad-\frac{4\snorm{X}_F^4}{D(X)^2}XX^\top X.
    \end{aligned}
\end{equation}
This is the gradient of the effective rank itself. For the negative-entropy
penalty $-H_2(\mathbf q)=-\log\rerank(X)$, the chain rule gives
\[
    \nabla_X[-H_2(\mathbf q)]
    =-\frac{\nabla_X\rerank(X)}{\rerank(X)}.
\]

\paragraph{Numerical implications.}
The gradient uses matrix products and traces, with no inverse singular-value
gaps or logarithms of individual singular values. Moreover,
\[
    D(X)=\sum_i\sigma_i^4>0\qquad\text{for }X\ne 0.
\]
Therefore $\rerank$ is smooth at every nonzero matrix, including matrices
with repeated or zero singular values. This does not imply uniform numerical
stability near $X=0$: the ratio is undefined there, and its gradient can grow
as the matrix scale decreases. Any numerical stabilization at zero must be
specified separately in the implementation.
Efficient evaluation from LoRA factors is described in
Appendix~\ref{apdx:implementation-tricks}.

\section{Implementation tricks}
\label{apdx:implementation-tricks}

\paragraph{Goal.}
Section~\ref{sec:methods} adds the \penaltyname{} to the language-modeling loss as a function of the adapter update $\Delta W$.
Because $\rerank(\Delta W)$ depends only on unitarily invariant quantities---the Frobenius norm and traces of powers of $\Delta W^\top \Delta W$---we never need to form the dense update matrix during training.
This appendix records the low-rank factor identities used in our implementation and explains how they extend to non-standard LoRA parameterizations.

\paragraph{LoRA factor notation.}
Fix a single targeted layer and write the adapter update as
\begin{equation}
    \Delta W \;=\; \gamma\, B A,
\end{equation}
where $A \in \R^{r \times n}$, $B \in \R^{m \times r}$, and $\gamma=\alpha/r>0$. Here, $\alpha$ is the LoRA scaling hyperparameter.
Thus $\Delta W \in \R^{m \times n}$ has rank at most $r$.
In standard LoRA, $A$ and $B$ are the trainable low-rank factors; other parameterizations below expose \emph{effective} factors $(A,B,\gamma)$ with the same product structure.

\paragraph{Gram identities.}
Define the reduced Gram product
\begin{equation}
    G \;:=\; (B^\top B)(A A^\top) \;\in\; \R^{r \times r}.
\end{equation}
Then

\begin{align}
    \snorm{\Delta W}_F^2
    &= \gamma^2 \,\Tr\!\big((B^\top B)(A A^\top)\big)
    \nonumber\\
    &= \gamma^2 \,\Tr(G),
    \label{eq:gram-frob} \\
    \Tr\!\big((\Delta W^\top \Delta W)^2\big)
    &= \gamma^4 \,\Tr\!\Big(\big((B^\top B)(A A^\top)\big)^2\Big)\\
    &= \gamma^4 \,\Tr(G^2).
    \label{eq:gram-collision}
\end{align}

Both identities follow from cyclicity of the trace.
For \eqref{eq:gram-frob}, use

\begin{align*}
	\snorm{\Delta W}_F^2
	&= \Tr(\Delta W^\top \Delta W)  \\
	&= \gamma^2\Tr(A^\top B^\top B A) \\
	&= \gamma^2\Tr(B^\top B A A^\top)
\end{align*}

For \eqref{eq:gram-collision}, expand $(\Delta W^\top \Delta W)^2 = \gamma^4 A^\top B^\top B A A^\top B^\top B A$ and apply the same cyclic reorderings.
Only the $r \times r$ Gram matrices $A A^\top$ and $B^\top B$ (and their product $G$) are required; the full $m \times n$ matrix $\Delta W$ never appears.

\paragraph{Evaluating the \penaltyname{} without materializing $\Delta W$.}
Recall from Section~\ref{sec:methods} that
\begin{equation}
    \rerank(\Delta W)
    \;=\;
    \frac{\snorm{\Delta W}_F^4}{\Tr\!\big((\Delta W^\top \Delta W)^2\big)}.
\end{equation}
Substituting \eqref{eq:gram-frob}--\eqref{eq:gram-collision} yields
\begin{equation}
    \rerank(\Delta W)
    \;=\;
    \frac{\big(\gamma^2 \,\Tr(G)\big)^2}{\gamma^4 \,\Tr(G^2)}
    \;=\;
    \frac{\Tr(G)^2}{\Tr(G^2)}.
\end{equation}
Equivalently, the R\'enyi-2 (collision) entropy of the energy distribution $\mathbf q$ satisfies
\begin{equation}
    \begin{aligned}
    H_2(\mathbf q)
    &=
    -\log \frac{\Tr(G^2)}{\Tr(G)^2}
    \\
    &=
    2\log \Tr(G) - \log \Tr(G^2).
    \end{aligned}
\end{equation}

\textbf{Logarithmic training penalty.} In our implementation, we add \(-\lambda H_2(\mathbf q)=-\lambda\log\rerank(\Delta W)\), rather than \(-\lambda\rerank(\Delta W)\), to the language-modeling loss. For nonzero updates, \(\nabla H_2=\rerank(\Delta W)^{-1}\nabla\rerank(\Delta W)\), so the two penalties encourage the same per-module ascent direction, but with different gradient magnitudes. The logarithmic form therefore encourages higher effective rank while introducing a state-dependent rescaling of the penalty gradient relative to the language-modeling gradient.

Automatic differentiation through \eqref{eq:gram-frob}--\eqref{eq:gram-collision} is sufficient; it is not necessary to instantiate the closed-form dense gradient in Appendix~\ref{apdx:gradient-of-renyi-2-penalty}, which is written in terms of $\Delta W$ for exposition.
The same Gram recipe also implements the Jensen / Frobenius surrogate $-\snorm{\Delta W}_F^2 = -\gamma^2 \Tr(G)$ used in our norm-bound motivation (Theorem~\ref{thm:renyi-2-bound-via-schatten}).

\paragraph{Complexity and memory.}
Explicitly storing the dense update requires $\Theta(mn)$ additional memory.
Computing its Gram matrix using the smaller dimension costs
\[
    \Theta\bigl(mn\min\{m,n\}\bigr).
\]
The factor-based route instead uses
\[
    \begin{aligned}
        \text{additional memory}
        &:\quad \Theta(r^2), \\
        \text{arithmetic}
        &:\quad \Theta\bigl(r^2(m+n)+r^3\bigr).
    \end{aligned}
\]
The quadratic storage holds the reduced Gram matrices and their product $G$.
The cubic term forms $G$; once $G$ is available, evaluating $\Tr(G^2)$ requires only quadratic work:
\[
    \Tr(G^2)=\sum_{i,j}G_{ij}G_{ji}.
\]
Thus the factor-based route reduces additional storage when $r\ll\min\{m,n\}$.
We aggregate the per-module penalties over the targeted MLP \texttt{down\_proj} layers.

\paragraph{Parameterization-agnostic effective factors.}
The \penaltyname{} attaches to the language-modeling loss, not to a specific adapter parameterization.
Any module whose update can be written as $\Delta W = \gamma B A$ can be penalized with \eqref{eq:gram-frob}--\eqref{eq:gram-collision} once effective factors are exposed:
\begin{itemize}
    \item \textbf{Standard LoRA:} $A$ and $B$ are the PEFT low-rank factors.
    \item \textbf{LoRA-XS:} $\Delta W = \tilde U R \tilde V^\top$ with fixed truncated-SVD subspace $(\tilde U, \tilde V)$; set $B=\tilde U$ and $A = R \tilde V^\top$.
    \item \textbf{PiSSA:} For the accumulated update $\Delta W=\gamma(BA-B_0A_0)$, use effective factors
    \[
        B_{\mathrm{eff}}=[B\;\;B_0],\qquad
        A_{\mathrm{eff}}=\begin{bmatrix}A\\-A_0\end{bmatrix}.
    \]
    Apply the same Gram identities to these factors; this representation has effective factor width $2r$.
    \item \textbf{NB-LoRA:} apply the Cayley map to obtain semi-orthogonal blocks and read off effective $(A,B,\gamma)$ from the resulting $2 Y^\top S X$ construction (Section~\ref{sec:nb-lora-route}).
\end{itemize}
In each case, gradients flow into the trainable parameters through the effective factors; the frozen subspace buffers in LoRA-XS do not need to be recomputed each step.

\paragraph{Numerical stability.}
We evaluate \eqref{eq:gram-frob}--\eqref{eq:gram-collision} in \texttt{float32} even when the model trains in \texttt{bfloat16}, and clamp $\Tr(G)$ and $\Tr(G^2)$ below by a small $\varepsilon>0$ before taking logarithms.
Unlike SVD-based objectives, the penalty does not divide by singular-value gaps (Appendix~\ref{apdx:direct-differentiation-of-erank}); the only delicate regime is $\Delta W \approx 0$, where $\Tr(G^2)\to 0$ and the penalty gradient grows.
This motivates the late-start schedules in Section~\ref{sec:weight-scheduler} and, optionally, geometry-aware gating of $\lambda$ based on the relative adapter norm $\snorm{\Delta W}_F / \snorm{W_0}_F$, which is also computable from \eqref{eq:gram-frob} without materializing $\Delta W$.

\paragraph{Logging singular values.}
Training uses only \eqref{eq:gram-frob}--\eqref{eq:gram-collision}.
For logging Roy--Vetterli $\erank(\Delta W)$ and related diagnostics in tables, we optionally recover singular values without forming $\Delta W$ by QR factorization of $\gamma B$ followed by an SVD of the $r \times n$ matrix $R_{\mathrm{qr}} A$, where $R_{\mathrm{qr}}$ is the reduced QR factor.
This is $\mathcal O(r^2(m+n))$ plus $\mathcal O(r^3)$ and is used for metrics only, not in the training objective.

\section{Training details}
\label{sec:appendix-baselines}

This appendix records the hyperparameters and training protocols used to produce the main results.
Numeric outcomes appear in the main text (e.g.\ Figure~\ref{fig:mixed-tasks-train-forgetting-tradeoff}) and in the extended per-domain baseline tables that follow in the appendix.

\subsection{Default hyperparameters}
\label{subsec:default-hp}

\paragraph{Scope.}
Our primary supervised fine-tuning (SFT) experiments fine-tune \texttt{Qwen/Qwen2.5-0.5B} with assistant-only loss; adaptation is restricted to the MLP down-projection matrices.
LoRA-based methods use the maximum feasible rank for this policy on the base model ($r{=}896$), with scaling $\alpha{=}2r$.
Dense down-projection fine-tuning updates the same dense \texttt{down\_proj} subset rather than the entire checkpoint.
Comparisons across adapter parameterizations are based on each method's published or default hyperparameters and are not parameter-count matched; trainable parameter totals are reported in Table~\ref{tab:mixed-task-baseline-comparison}.

\paragraph{Optimization (SFT, 0.5B).}
We use AdamW with learning rate $10^{-4}$, weight decay $0.01$, and a linear learning-rate warmup of $20$ optimizer steps followed by a constant peak rate.
Each optimizer step uses micro-batch size $1$ and gradient accumulation $2$ (effective batch size $2$ packed sequences).
Training sequences are capped at $512$ tokens; documents are packed into blocks with dynamic batch padding.
LoRA dropout is $0$.
Evaluations run every $100$ optimizer steps.

\paragraph{Training budgets.}
Three budget regimes appear in the main text:
\begin{itemize}
    \item \textbf{Mixed-domain continual-learning proxy} (Section~\ref{sec:mixed-task-results}): a fixed budget of $10^6$ \emph{supervised target tokens} on a shuffled mixture of GSM8K ($45\%$), WMT19 Czech--English ($10\%$), and Aya Hausa instruction-following ($45\%$), with shuffle buffer $50{,}000$.
    \item \textbf{Single-task SFT} (per-domain tables in the appendix): one pass over the training split.
    \item \textbf{LoRA rank sweep} (Section~\ref{sec:lora-rank-sweep}): $10^6$ supervised target tokens on the mixed-domain continual-learning proxy (45\% GSM8K, 10\% WMT19 Czech--English, 45\% Aya Hausa), sweeping $r\in\{4,16,128,256,512\}$.
\end{itemize}
The weight-scheduler ablation (Section~\ref{sec:weight-scheduler}, Table~\ref{tab:weight-scheduler-ablations}) uses a \emph{separate}, shorter budget of ${\sim}500$ optimizer steps (${\sim}3\times 10^5$ tokens) on Hausa only.

\paragraph{Dense down-projection fine-tuning selection (0.5B).}
Dense down-projection fine-tuning baselines sweep learning rate $\in\{10^{-5},\,3\times 10^{-5},\,10^{-4}\}$, micro-batch size $\in\{1,2,4\}$, weight decay $\in\{0,\,0.01,\,0.1\}$, and scheduler shape $\in\{\text{linear warmup},\,\text{warmup-cosine}\}$ (54 grid points per SFT dataset).
Among \emph{completed} grid points on the same training budget, we select hyperparameters using the lowest \textbf{Train PPL} (the smoothed training proxy, not held-out validation perplexity). The dense mixed-domain baseline described below is reported under the configuration below, without a claim that it is the best point in the original grid.
Unless specified below, selected dense runs follow the shared SFT protocol of Section~\ref{subsec:default-hp}: AdamW, micro-batch size $1$, gradient accumulation $2$ (effective batch size $2$ packed sequences), $20$-step linear warmup unless noted otherwise, and trainability restricted to MLP down-projection matrices.
Warmup-cosine schedules use minimum LR ratio $0.1$; total optimizer steps are $2000$ for single-task dense down-projection fine-tuning sweeps.

\smallskip
\noindent\textbf{Configurations for reported dense baselines.}
\begin{itemize}
    \item \textbf{Mixed continual-learning} (Table~\ref{tab:mixed-task-baseline-comparison}): the dense down-projection fine-tuning baseline uses learning rate $3\times10^{-5}$, weight decay $0$, micro-batch $4$, gradient accumulation $2$, and $20$-step linear LR warmup followed by a constant rate, under the $10^6$ supervised-token cap. Spectral analyses use a separate dense baseline.
    \item \textbf{GSM8K}: learning rate $10^{-4}$, weight decay $0.01$, micro-batch $1$, warmup-cosine ($20$ warmup steps, $2000$ total optimizer steps, min LR ratio $0.1$).
    \item \textbf{WMT19 cs--en}: learning rate $10^{-4}$, weight decay $0.01$, micro-batch $1$, linear warmup ($20$ steps), one training epoch.
    \item \textbf{Aya Hausa}: learning rate $10^{-4}$, weight decay $0.01$, micro-batch $1$, linear warmup ($20$ steps), one training epoch.
\end{itemize}
Table values use each metric's last logged value unless a caption states otherwise; this need not coincide with the final saved checkpoint or the last logging step of another metric.

\paragraph{Precision and hardware.}
We load \texttt{Qwen/Qwen2.5} checkpoints with Hugging Face Transformers and do \emph{not} enable automatic mixed precision, loss scaling, or fp32 master weights.
For Qwen2.5, the released checkpoints are stored in \texttt{bfloat16}; our loader does not upcast base weights to \texttt{float32} before training.
Subspace-based adapter initialization (LoRA-XS, PiSSA) computes truncated SVDs in \texttt{float32} once at setup and stores the resulting fixed bases in the base weight dtype.
The \penaltyname{} terms \eqref{eq:gram-frob}--\eqref{eq:gram-collision} are likewise evaluated in \texttt{float32} for numerical stability (Appendix~\ref{apdx:implementation-tricks}).
SFT runs do not apply gradient clipping.
All SLURM jobs target one NVIDIA A40 GPU with $48$\,GiB host memory unless noted otherwise.

\subsection{\penaltyname{} configuration}
\label{subsec:rank-penalty-hp}

\paragraph{Penalty form.}
In all reported penalty experiments, we use the logarithmic variant of
Equation~\eqref{eq:penalized-loss}, adding
$-\lambda\sum_{\ell}H_2(\mathbf q_\ell)$ to the language-modeling loss,
where $\ell$ indexes the targeted adapter modules.
We evaluate this penalty with stability constant $\varepsilon{=}10^{-8}$,
as detailed in Appendix~\ref{apdx:implementation-tricks}.
All reported penalty coefficients refer to this logarithmic form;
main SFT sweeps vary the terminal coefficient
$\lambda\in\{0,\,0.01,\,0.1,\,1.0\}$.

\paragraph{Penalty schedule.}
The penalty weight is 0 for optimizer steps $t<S$, then increases linearly from 0 to $\lambda$ over W steps. Mixed-domain experiments (Tables~\ref{tab:mixed-task-baseline-comparison}--\ref{tab:arxiv-curated-pythia-2p8b-rank-sweep}) use $S=0, W=50$. Single-task experiments use $S=400, W=200$, except as follows:
For Aya Hausa with Qwen2.5-0.5B (Table~\ref{tab:aya-hausa-lora-rank-sweep-qwen-25-05b}), ranks $r\leq128$ use $S=400$, $W=100$; rank $256$ uses $S=400$ with $W=50$ for $\lambda=0.1$ and $W=200$ otherwise; rank $512$ uses $W=200$ with $S=0$.
Tables~\ref{tab:gsm8k-pissaloraxs-rank-penalty} and~\ref{tab:wmt19-cs-en-baseline-comparison-qwen-25-3b} note their own settings.
The scheduling ablation (Table~\ref{tab:weight-scheduler-ablations}) fixes $\lambda_1{=}0.1$ and $W{=}200$ while varying $S\in\{0,50,100,400\}$.

\subsection{SFT datasets and preprocessing}
\label{subsec:sft-data}

\paragraph{Datasets.}
Single-task runs use the Hugging Face identifiers \texttt{openai/gsm8k}, \texttt{wmt/wmt19:cs-en}, and \texttt{CohereLabs/aya\_dataset:Hausa} with standard \texttt{train}/\texttt{validation} splits.
The mixed-domain proxy concatenates the three sources with mixture weights $0.45/0.10/0.45$ and deterministic shuffling (mixture seed~$0$).
Hausa has no public test split; we therefore report validation perplexity for in-domain Aya metrics.

\paragraph{Supervision.}
For structured prompt--response SFT training examples, cross-entropy loss is computed only on assistant/target tokens (prompt tokens are masked).
This matches chat-style instruction tuning.

\paragraph{Forgetting probes.}
We evaluate fixed reference slices from the \texttt{wiki} and \texttt{starcoder} configurations of \texttt{allenai/olmo-mix-1124}, reporting \textbf{Wiki PPL} and \textbf{StarCoder PPL}. Each exported slice contains 2,000 documents selected from the source \texttt{train} split by streaming shuffle with seed 0 and buffer size 50,000. Evaluation uses at most 20 batches for mixed-domain runs and 50 batches for single-task runs unless otherwise stated.

Unlike the prompt--response SFT examples, these reference documents are plain text without assistant-role boundaries. Each document is treated as a fully supervised span, including its appended EOS token. Padding is excluded from the loss, and targets follow the standard causal shift within each block. Reported perplexity is the exponential of the token-weighted mean negative log-likelihood over the evaluated batches.

\paragraph{Reported metrics.}
Unless a table caption states otherwise, all numbers are taken from the \emph{terminal} training state of a completed run; we do not reload an in-run best-validation checkpoint. Early stopping is disabled.
Periodic evals run every $100$ optimizer steps; perplexity and forgetting probes below are the values logged at the last eval before the run hits its token or epoch budget.
\emph{Train PPL} is $\exp(H)$ where $H$ is an exponential moving average ($\beta{=}0.99$) of the assistant-only training loss; it is a smoothed training-loss proxy, not held-out validation perplexity.
\emph{Wiki PPL} and \emph{StarCoder PPL} are reference-probe perplexities on fixed OLMo-Mix-1124 slices, using the plain-text supervision protocol and batch limits specified above.
In the mixed-domain setting we additionally log in-mixture retention probes on held-out GSM8K and WMT19 Czech--English validation text; Appendix Table~\ref{tab:mixed-task-baseline-comparison} reports both in-mixture probes.
\emph{eRank} is the Roy--Vetterli effective rank of the accumulated adapter update $\Delta W$, averaged over all targeted \texttt{down\_proj} modules, taken from the last logged rank diagnostic. This is distinct from the R\'enyi-2 surrogate used in the objective.
Dense down-projection fine-tuning baselines select hyperparameters via a separate sweep (Section~\ref{subsec:default-hp}), but reported metrics still refer to the terminal state of the selected run.
Post-hoc spectral analyses (intruder dimensions, subspace alignment, singular-direction mixing) use measurements from the final saved checkpoint.

\subsection{Baseline adapter hyperparameters}
\label{subsec:baseline-adapters}

Table~\ref{tab:mixed-task-baseline-comparison} compares the following adapter families under the shared SFT protocol above.
Method-specific knobs are held fixed across datasets unless a caption states otherwise.

\paragraph{Standard LoRA.}
Trainable low-rank factors $(A,B)$ with rank $r{=}896$ and $\alpha{=}2r$.

\paragraph{PiSSA~\cite{pissa}.}
Each targeted weight is decomposed as $W_0 = W_{\mathrm{res}} + \gamma B_0 A_0$ using the top-$r$ SVD of the pretrained matrix.
Here, $\gamma=\alpha/r$ follows Appendix~\ref{apdx:implementation-tricks}; the accumulated update relative to the original pretrained weight is
\[
    \Delta W=\gamma(BA-B_0A_0).
\]
The residual $W_{\mathrm{res}}$ is stored frozen; trainable factors $(A,B)$ are initialized to $(A_0,B_0)$ and have the same shapes as standard LoRA at rank~$r$.
PiSSA therefore has the \emph{same trainable parameter count} as standard LoRA at matched rank and targets, not a singular-value-only parameterization.
At $r{=}896$ on Qwen2.5-0.5B this yields ${\sim}1.24\times 10^8$ trainable adapter parameters across all \texttt{down\_proj} layers.

\paragraph{LoRA-XS~\cite{lora-xs}.}
The pretrained matrix supplies fixed subspace bases $\tilde U = U_r \Sigma_r$ and $\tilde V = V_r$ from a truncated SVD; only a trainable core $R\in\R^{r\times r}$ is optimized.
This substantially reduces trainable parameter count relative to standard LoRA at the same nominal rank (Table~\ref{tab:mixed-task-baseline-comparison}, \# Params.\ column).

\paragraph{VeRA~\cite{vera}.}
Shape-shared frozen random matrices plus a trainable scaling vector $d$, with initial scaling value $0.1$ and random seed $0$ for reproducible adapter construction across layers of the same shape bucket.

\paragraph{SVF~\cite{singularvaluefinetuning}.}
Trainable gains on top singular directions starting from the leading singular direction.
Unlike standard LoRA, the LoRA scaling hyperparameter $\alpha$ is not used as a forward multiplier in our SVF implementation.

\paragraph{TinyLoRA~\cite{tinylora}.}
A single trainable coefficient mixes fixed random rank-$r$ projections per layer.
All MLP down-projection layers share one scalar trainable weight (full tying).
Frozen projection matrices are drawn with seed~$0$.

\paragraph{Norm-bounded LoRA (NB-LoRA).}
When reported, Cayley-based norm-bounded LoRA uses spectral bound $\delta{=}1.0$ on the effective update.
Optional Frobenius penalty sweeps use separate weight grids from the \penaltyname{} runs.

\paragraph{Interpretability caveats.}
Low-capacity baselines (VeRA, TinyLoRA, SVF) occupy a different parameter budget than rank-$896$ LoRA and should be read as an underfitting regime rather than a matched-capacity head-to-head.
PiSSA and SVF differ from standard LoRA in how scaling constants enter the forward pass.
We record the trainable parameter count for every configuration.

\subsection{Model scale variants}
\label{subsec:scale-variants}

Appendix tables for \texttt{Qwen/Qwen2.5-3B-Instruct} also train only the MLP down-projection matrices but increase adapter rank to $r{=}2048$ with $\alpha{=}4096$ ($\alpha/r{=}2$).
All other SFT knobs follow Section~\ref{subsec:default-hp} unless a table caption notes a deviation.

\subsection{Reproducibility}
\label{subsec:repro}

\paragraph{Completed mixed-domain repeats.}
The completed mixed-domain repeat executions use a fixed mixture seed $0$, the same $0.45/0.10/0.45$ domain weights, and a $10^6$ supervised-token cap. Adapter runs use rank $896$, $\alpha=1792$, micro-batch $1$, gradient accumulation $2$, learning rate $10^{-4}$, and weight decay $0.01$. The penalized repeat ramps to $0.1$ over $50$ steps from step $0$; TinyLoRA retains one shared scalar across all targeted layers. Adapter runs finish at step $1557$ with $999{,}761$ logged tokens; their rank and held-out metrics are last logged at steps $1550$ and $1500$. The dense run finishes at step $407$ with $1{,}000{,}373$ logged tokens, rank logging at step $407$, and held-out logging at step $400$. The dense and adapter optimizer/batch settings differ as recorded above.

\section{Additional experimental results and robustness analyses}
\label{sec:additional-results}

This appendix supplements Section~\ref{sec:experiments} with mixed-domain baseline comparisons and repeat executions, rank and penalty sensitivity analyses, and results across model families and single-task settings. We first present the mixed-domain results, then turn to single-task adaptation. Results applying the penalty to PiSSA and LoRA-XS appear in the main text.

\paragraph{Reporting conventions.}
Dense FT denotes dense fine-tuning of only the targeted
\texttt{down\_proj} matrices, rather than full-model fine-tuning.
For LoRA, $\lambda$ denotes the R\'enyi-2 penalty weight:
$\lambda=0$ corresponds to standard, unregularized LoRA, and
$\lambda>0$ enables the penalty.
A dash in the $\lambda$ column indicates that the penalty weight
is not applicable.

For entries of the form $\bar{x}\pm s$, $\bar{x}=n^{-1}\sum_i x_i$ is the mean and $s=\sqrt{\sum_i(x_i-\bar{x})^2/(n-1)}$ the sample standard deviation over the reported $n$ training runs $x_1,\ldots,x_n$ (typically $n=3$, with exceptions stated in each table). Entries without $\pm$ are reported without an uncertainty estimate.

Dense baselines update only the targeted \texttt{down\_proj} matrices. Appendix~\ref{sec:appendix-baselines} gives the training and measurement protocols; captions retain exceptions specific to each table.

\subsection{Mixed-domain baseline comparisons}
\label{subsec:mixed-domain-baselines-numeric}
\label{subsec:additional-mixed-domain-baselines}

\paragraph{Endpoints corresponding to the main figure.}
Table~\ref{tab:mixed-task-baseline-comparison} reports the full numeric endpoints for the mixed-domain continual-learning experiment summarized in Figure~\ref{fig:mixed-tasks-train-forgetting-tradeoff}.

\begin{table*}[!tp]
    \centering
    \begingroup
    \small
    \setlength{\tabcolsep}{3pt}\spaceskip=1.5pt
    \def\nosd{\phantom{ $\pm$ 0.000}}%
    \begin{adjustbox}{max width=\textwidth}
    \begin{tabular}{@{}lrrrrrrrr@{}}
    \toprule
    Method & $n$ & \shortstack[r]{\#\\Params.} & \shortstack[r]{eRank\\($\uparrow$)} & \shortstack[r]{Train PPL\\($\downarrow$)} & \shortstack[r]{StarCoder PPL\\($\downarrow$)} & \shortstack[r]{Wiki PPL\\($\downarrow$)} & \shortstack[r]{GSM8K PPL\\($\downarrow$)} & \shortstack[r]{WMT cs--en PPL\\($\downarrow$)} \\
    \midrule
    \multicolumn{9}{l}{\textit{Standard}} \\
    Dense FT & 1 & --- & 844.479\nosd & 5.643\nosd & 3.640\nosd & 16.563\nosd & 1.817\nosd & 9.223\nosd \\
    Standard LoRA & 3 & 123,863,040 & 360.537 $\pm$ 5.190 & 4.773 $\pm$ 0.093 & 5.815 $\pm$ 0.811 & 19.601 $\pm$ 1.056 & 1.929 $\pm$ 0.044 & 15.018 $\pm$ 0.802 \\
    \midrule
    \multicolumn{9}{l}{\textit{LoRA + \penaltyname{} (Ours)}} \\
    LoRA ($\lambda=0.01$) & 3 & 123,863,040 & 667.960 $\pm$ 1.366 & 4.381 $\pm$ 0.007 & 4.143 $\pm$ 0.006 & 15.290 $\pm$ 0.067 & 1.836 $\pm$ 0.010 & 10.156 $\pm$ 0.034 \\
    LoRA ($\lambda=0.1$) & 3 & 123,863,040 & 887.772 $\pm$ 0.333 & 4.525 $\pm$ 0.016 & 4.098 $\pm$ 0.007 & 14.648 $\pm$ 0.046 & 1.832 $\pm$ 0.008 & 9.517 $\pm$ 0.172 \\
    LoRA ($\lambda=1.0$) & 1 & 123,863,040 & 895.515\nosd & 4.688\nosd & 4.003\nosd & 14.085\nosd & 1.800\nosd & 8.909\nosd \\
    \midrule
    \multicolumn{9}{l}{\textit{LoRA variants}} \\
    PiSSA & 2 & 123,863,040 & 815.725 $\pm$ 8.182 & 4.286 $\pm$ 0.003 & 5.162 $\pm$ 0.102 & 21.565 $\pm$ 0.658 & 1.878 $\pm$ 0.002 & 13.160 $\pm$ 0.020 \\
    PiSSA ($\lambda=0.1$) & 1 & 123,863,040 & 892.218\nosd & 4.335\nosd & 4.769\nosd & 18.634\nosd & 1.874\nosd & 12.109\nosd \\
    LoRA-XS & 3 & 19,267,584 & 566.817 $\pm$ 4.951 & 4.578 $\pm$ 0.005 & 5.517 $\pm$ 0.050 & 21.431 $\pm$ 0.584 & 1.901 $\pm$ 0.010 & 14.691 $\pm$ 0.158 \\
    NB-LoRA & 3 & 123,863,040 & 583.243 $\pm$ 1.475 & 4.396 $\pm$ 0.002 & 4.404 $\pm$ 0.028 & 15.705 $\pm$ 0.073 & 1.809 $\pm$ 0.003 & 10.196 $\pm$ 0.107 \\
    VeRA & 3 & 43,008 & 632.194 $\pm$ 0.015 & 11.468 $\pm$ 0.001 & 3.832 $\pm$ 0.001 & 13.005 $\pm$ 0.006 & 2.373 $\pm$ 0.000 & 9.309 $\pm$ 0.007 \\
    SVF & 3 & 21,504 & 783.579 $\pm$ 0.075 & 9.694 $\pm$ 0.257 & 3.895 $\pm$ 0.002 & 13.186 $\pm$ 0.006 & 2.061 $\pm$ 0.000 & 8.814 $\pm$ 0.006 \\
    TinyLoRA & 3 & 1 & 681.531 $\pm$ 0.233 & 11.549 $\pm$ 0.001 & 3.833 $\pm$ 0.000 & 13.006 $\pm$ 0.008 & 2.388 $\pm$ 0.001 & 9.341 $\pm$ 0.003 \\
    \bottomrule
    \end{tabular}
    \end{adjustbox}\par
    \endgroup
    \caption{Dense FT denotes dense down-projection fine-tuning.
        Mixed-domain results for Qwen2.5-0.5B corresponding to Figure~\ref{fig:mixed-tasks-train-forgetting-tradeoff}, with MLP down-projection adapters ($r=896$, $\alpha=1792$) trained on the GSM8K/WMT19 Czech--English/Aya Hausa mixture.
        Rows with $n>1$ report the mean $\pm$ sample standard deviation over $n$ training executions. Rows with $n=1$ report a single run. PiSSA ($n=2$) excludes one resumed execution with apparent data replay, as discussed in the text.
        All rows use Qwen2.5-0.5B (base, not Instruct); for Dense FT this is taken from the job launch records.
    }
    \label{tab:mixed-task-baseline-comparison}
    \label{tab:arxiv-curated-qwen-2p5-0p5b-baselines}
\end{table*}

\paragraph{PiSSA endpoint-integrity check.}
The original three-run PiSSA aggregate had Train PPL $3.916\pm0.641$. All three runs share the intended base checkpoint, mixture, adapter, and optimization settings. However, seed 3 was preempted and resumed at step 1500; its subsequent 68 supervised-token-count increments exactly repeat an earlier segment, consistent with data replay. Its terminal Train PPL fell to $3.176$, while its held-out probes remained at the pre-resume checkpoint. We therefore exclude this execution from all six terminal-summary metrics, rather than selecting different runs for different metrics. Seeds 0 and 6 give Train PPL $4.286\pm0.003$ ($n=2$). The original measurements and the resume audit remain archived; the token-count evidence indicates apparent replay but does not prove input-token identity.

\paragraph{Comparisons across the primary models.}
Tables~\ref{tab:arxiv-curated-qwen-2p5-0p5b-baselines}, \ref{tab:arxiv-curated-qwen-2p5-3b-it-baselines}, and~\ref{tab:arxiv-curated-llama-3p2-3b-it-baselines} report mixed-domain baseline comparisons for Qwen2.5-0.5B, Qwen2.5-3B-Instruct, and Llama-3.2-3B-Instruct, respectively. The Qwen2.5-0.5B table is the merged source for Figure~\ref{fig:mixed-tasks-train-forgetting-tradeoff}; the other two tables extend the comparison to larger models. Each summary retains its reported run group and uncertainty convention, subject to the PiSSA endpoint exclusion above; superseded individual runs are not pooled into those summaries.

The WMT cs--en columns in the primary baseline and rank-sweep summaries are reconstructed from matched training runs. The tables report effective rank and Train, StarCoder, Wiki, and GSM8K perplexities with their stated uncertainty conventions; WMT margins use sample standard deviations over the same aggregation units.

\begin{table*}[!tp]
    \centering
    \begingroup
    \small
    \setlength{\tabcolsep}{1.4pt}\spaceskip=1.5pt
    \setbox0=\hbox{%
    \begin{tabular}{@{}lrrrrrrr@{}}
    \toprule
    Method & $\lambda$ & \shortstack[r]{eRank\\($\uparrow$)} & \shortstack[r]{Train PPL\\($\downarrow$)} & \shortstack[r]{StarCoder PPL\\($\downarrow$)} & \shortstack[r]{Wiki PPL\\($\downarrow$)} & \shortstack[r]{GSM8K PPL\\($\downarrow$)} & \shortstack[r]{WMT cs--en PPL\\($\downarrow$)} \\
    \midrule
    Standard LoRA & 0 & 466.103 $\pm$ 18.995 & 3.826 $\pm$ 0.043 & 4.078 $\pm$ 0.298 & 12.591 $\pm$ 0.334 & 1.728 $\pm$ 0.006 & 6.034 $\pm$ 0.187 \\
    LoRA + penalty & 0.01 & 591.066 $\pm$ 11.191 & 3.678 $\pm$ 0.008 & 3.335 $\pm$ 0.036 & 10.704 $\pm$ 0.103 & 1.707 $\pm$ 0.002 & 5.381 $\pm$ 0.330 \\
    LoRA + penalty & 0.1 & 1057.340 $\pm$ 60.090 & 3.676 $\pm$ 0.041 & 4.059 $\pm$ 0.988 & 9.979 $\pm$ 0.277 & 1.696 $\pm$ 0.010 & 4.753 $\pm$ 0.381 \\
    \midrule
    Dense FT & --- & 1848.000 $\pm$ 4.599 & 4.376 $\pm$ 0.354 & 2.879 $\pm$ 0.009 & 10.634 $\pm$ 1.261 & 1.678 $\pm$ 0.025 & 4.187 $\pm$ 0.287 \\
    \midrule
    VeRA & --- & 1457.454 $\pm$ 0.360 & 9.352 $\pm$ 0.000 & 2.819 $\pm$ 0.001 & 9.142 $\pm$ 0.001 & 2.546 $\pm$ 0.002 & 5.382 $\pm$ 0.002 \\
    \bottomrule
    \end{tabular}%
    }
    \ifdim\wd0>\textwidth
        \PackageError{arxiv-curated}{Table exceeds text width: qwen-2p5-3b-it-baselines}{Adjust the curated layout; do not scale it down.}
    \fi
    \typeout{CURATED-WIDTH qwen-2p5-3b-it-baselines: \the\wd0; limit \the\textwidth}
    \leavevmode\box0\par
    \endgroup
    \caption{Mixed-domain baseline comparison for Qwen2.5-3B-Instruct; LoRA rank $r=2048$.}
    \label{tab:arxiv-curated-qwen-2p5-3b-it-baselines}
    \par\smallskip
\end{table*}

\begin{table*}[!tp]
    \centering
    \begingroup
    \small
    \setlength{\tabcolsep}{1.4pt}\spaceskip=1.5pt
    \setbox0=\hbox{%
    \begin{tabular}{@{}lrrrrrrr@{}}
    \toprule
    Method & $\lambda$ & \shortstack[r]{eRank\\($\uparrow$)} & \shortstack[r]{Train PPL\\($\downarrow$)} & \shortstack[r]{StarCoder PPL\\($\downarrow$)} & \shortstack[r]{Wiki PPL\\($\downarrow$)} & \shortstack[r]{GSM8K PPL\\($\downarrow$)} & \shortstack[r]{WMT cs--en PPL\\($\downarrow$)} \\
    \midrule
    Standard LoRA & 0 & 985.353 $\pm$ 36.936 & 3.984 $\pm$ 0.027 & 7.031 $\pm$ 0.372 & 18.547 $\pm$ 0.475 & 2.114 $\pm$ 0.024 & 7.127 $\pm$ 0.452 \\
    LoRA + penalty & 0.01 & 1504.231 $\pm$ 6.870 & 3.837 $\pm$ 0.109 & 5.420 $\pm$ 0.054 & 15.924 $\pm$ 0.094 & 2.057 $\pm$ 0.016 & 6.310 $\pm$ 0.477 \\
    LoRA + penalty & 0.1 & 2725.094 $\pm$ 5.379 & 3.679 $\pm$ 0.055 & 4.497 $\pm$ 0.025 & 13.327 $\pm$ 0.124 & 1.981 $\pm$ 0.014 & 4.725 $\pm$ 0.075 \\
    \midrule
    Dense FT & --- & 2833.119 $\pm$ 0.030 & 3.975 $\pm$ 0.000 & 3.983 $\pm$ 0.001 & 13.889 $\pm$ 0.001 & 1.890 $\pm$ 0.000 & 4.097 $\pm$ 0.001 \\
    \midrule
    LoRA-XS & --- & 1840.308 $\pm$ 19.541 & 4.192 $\pm$ 0.057 & 32.341 $\pm$ 18.745 & 34.911 $\pm$ 4.410 & 2.208 $\pm$ 0.021 & 14.077 $\pm$ 5.181 \\
    PiSSA & --- & 2640.110 $\pm$ 10.920 & 3.503 $\pm$ 0.048 & 7.376 $\pm$ 0.080 & 19.143 $\pm$ 0.473 & 2.024 $\pm$ 0.016 & 6.315 $\pm$ 0.059 \\
    SVF & --- & 2669.245 $\pm$ 0.176 & 7.768 $\pm$ 0.000 & 3.939 $\pm$ 0.001 & 11.309 $\pm$ 0.003 & 2.385 $\pm$ 0.000 & 4.542 $\pm$ 0.002 \\
    TinyLoRA & --- & 2295.437 $\pm$ 0.555 & 9.094 $\pm$ 0.008 & 4.000 $\pm$ 0.004 & 11.474 $\pm$ 0.019 & 2.511 $\pm$ 0.001 & 4.598 $\pm$ 0.007 \\
    VeRA & --- & 2103.619 $\pm$ 0.028 & 9.012 $\pm$ 0.000 & 3.993 $\pm$ 0.001 & 11.448 $\pm$ 0.002 & 2.500 $\pm$ 0.000 & 4.596 $\pm$ 0.001 \\
    \bottomrule
    \end{tabular}%
    }
    \ifdim\wd0>\textwidth
        \PackageError{arxiv-curated}{Table exceeds text width: llama-3p2-3b-it-baselines}{Adjust the curated layout; do not scale it down.}
    \fi
    \typeout{CURATED-WIDTH llama-3p2-3b-it-baselines: \the\wd0; limit \the\textwidth}
    \leavevmode\box0\par
    \endgroup
    \caption{Mixed-domain baseline comparison for Llama-3.2-3B-Instruct; LoRA rank $r=3072$.}
    \label{tab:arxiv-curated-llama-3p2-3b-it-baselines}
    \par\smallskip
\end{table*}

\paragraph{Earlier Qwen2.5-3B-Instruct endpoints.}
The following table retains the earlier mixed-domain Qwen2.5-3B-Instruct comparison, including adapter and penalty settings absent from the corresponding aggregated summary. It is a separate set of reported endpoints and should not be interpreted as additional independent seeds for that summary.

\begin{table*}[!tp]
    \centering
    \small
    \begin{adjustbox}{max width=\textwidth}
    \begin{tabular}{@{}lrrrrrr@{}}
    \toprule
    Method                          & eRank ($\uparrow$) & Train PPL ($\downarrow$) & StarCoder PPL ($\downarrow$)  & Wiki PPL ($\downarrow$) & GSM8K PPL ($\downarrow$) & WMT cs--en PPL ($\downarrow$) \\ \midrule
    \multicolumn{7}{l}{\textit{Standard}} \\
Dense FT                & 1842.690 & 3.968 & 2.869 & 9.179 & 1.650 & 3.855 \\
Standard LoRA                  &  467.316 & 3.777 & 3.743 & 12.229 & 1.722 & 6.078 \\
    \midrule
    \multicolumn{7}{l}{\textit{LoRA + \penaltyname{} (Ours)}} \\
LoRA ($\lambda=0.01$)                  &  603.564 & 3.686 & 3.352 & 10.727 & 1.705 & 5.729 \\
LoRA ($\lambda=0.1$)                   & 1108.815 & 3.629 & 3.008 & 9.736 & 1.688 & 4.321 \\
LoRA ($\lambda=1.0$)                   & 1985.115 & 3.678 & 2.885 & 9.331 & 1.683 & 4.070 \\
    \midrule
    \multicolumn{7}{l}{\textit{LoRA variants}} \\
PiSSA                           & 1768.671 & 4.006 & 3.474 & 11.145 & 1.747 & 5.550 \\
LoRA-XS                        & 1239.648 & 3.833 & 6.060 & 18.087 & 1.834 & 9.195 \\
VeRA                           & 1457.040 & 9.352 & 2.818 & 9.143 & 2.547 & 5.384 \\
SVF                            & 1810.497 & 7.398 & 2.790 & 9.001 & 2.140 & 5.042 \\
TinyLoRA                       & 1548.348 & 9.482 & 2.827 & 9.171 & 2.578 & 5.414 \\
    \bottomrule
    \end{tabular}
    \end{adjustbox}

    \caption{
    Mixed-domain baseline comparison for Qwen2.5-3B-Instruct; LoRA rank=2048
    Dense FT denotes dense down-projection fine-tuning.
    }
\label{tab:mixed-task-baseline-comparison-qwen-25-3b-instruct}
\end{table*}

\subsection{Mixed-domain rank and penalty sensitivity}
\label{subsec:additional-mixed-domain-ranks}

These sweeps examine how the learning--retention balance changes with the configured rank and penalty weight. Fuller spectral utilization need not improve every perplexity metric, so effective rank should be interpreted jointly with mixture fit and the held-out probes.

\paragraph{Rank and penalty summaries for the primary models.}
Tables~\ref{tab:arxiv-curated-qwen-2p5-0p5b-rank-sweep}, \ref{tab:arxiv-curated-qwen-2p5-3b-it-rank-sweep}, and~\ref{tab:arxiv-curated-llama-3p2-3b-it-rank-sweep} collect the reported lower-rank and penalty-weight settings for Qwen2.5-0.5B, Qwen2.5-3B-Instruct, and Llama-3.2-3B-Instruct. The maximum-rank comparisons appear in Section~\ref{subsec:mixed-domain-baselines-numeric} rather than being repeated here. Only reported configurations are included.

\begin{table*}[!tp]
    \centering
    \begingroup
    \small
    \setlength{\tabcolsep}{3pt}
    \setbox0=\hbox{%
    \begin{tabular}{@{}rrrrrrrr@{}}
    \toprule
    Rank & $\lambda$ & \shortstack[r]{eRank\\($\uparrow$)} & \shortstack[r]{Train PPL\\($\downarrow$)} & \shortstack[r]{StarCoder PPL\\($\downarrow$)} & \shortstack[r]{Wiki PPL\\($\downarrow$)} & \shortstack[r]{GSM8K PPL\\($\downarrow$)} & \shortstack[r]{WMT cs--en PPL\\($\downarrow$)} \\
    \midrule
    4 & 0 & 3.014 $\pm$ 0.075 & 6.096 $\pm$ 0.009 & 3.926 $\pm$ 0.004 & 13.615 $\pm$ 0.037 & 1.802 $\pm$ 0.001 & 8.370 $\pm$ 0.045 \\
    4 & 0.01 & 3.999 $\pm$ 0.000 & 6.086 $\pm$ 0.013 & 3.932 $\pm$ 0.009 & 13.575 $\pm$ 0.017 & 1.803 $\pm$ 0.001 & 8.425 $\pm$ 0.062 \\
    4 & 0.1 & 4.000 $\pm$ 0.000 & 6.595 $\pm$ 0.012 & 3.942 $\pm$ 0.016 & 13.564 $\pm$ 0.001 & 1.821 $\pm$ 0.001 & 8.403 $\pm$ 0.068 \\
    \midrule
    16 & 0 & 9.689 $\pm$ 0.077 & 5.419 $\pm$ 0.006 & 3.953 $\pm$ 0.003 & 13.780 $\pm$ 0.020 & 1.783 $\pm$ 0.002 & 8.382 $\pm$ 0.008 \\
    16 & 0.01 & 15.990 $\pm$ 0.000 & 5.391 $\pm$ 0.005 & 3.950 $\pm$ 0.006 & 13.602 $\pm$ 0.031 & 1.780 $\pm$ 0.001 & 8.310 $\pm$ 0.022 \\
    16 & 0.1 & 15.999 $\pm$ 0.000 & 5.717 $\pm$ 0.004 & 3.941 $\pm$ 0.002 & 13.578 $\pm$ 0.013 & 1.795 $\pm$ 0.000 & 8.332 $\pm$ 0.039 \\
    \midrule
    128 & 0 & 74.494 $\pm$ 0.391 & 4.605 $\pm$ 0.002 & 4.012 $\pm$ 0.017 & 14.202 $\pm$ 0.047 & 1.767 $\pm$ 0.004 & 8.715 $\pm$ 0.079 \\
    128 & 0.01 & 126.714 $\pm$ 0.074 & 4.611 $\pm$ 0.002 & 3.961 $\pm$ 0.012 & 13.865 $\pm$ 0.032 & 1.763 $\pm$ 0.001 & 8.470 $\pm$ 0.028 \\
    128 & 0.1 & 127.951 $\pm$ 0.001 & 4.749 $\pm$ 0.002 & 3.958 $\pm$ 0.010 & 13.863 $\pm$ 0.005 & 1.767 $\pm$ 0.003 & 8.424 $\pm$ 0.046 \\
    \midrule
    256 & 0 & 145.032 $\pm$ 0.175 & 4.450 $\pm$ 0.001 & 4.067 $\pm$ 0.006 & 14.674 $\pm$ 0.057 & 1.778 $\pm$ 0.004 & 9.406 $\pm$ 0.064 \\
    256 & 0.01 & 250.540 $\pm$ 0.111 & 4.465 $\pm$ 0.003 & 3.992 $\pm$ 0.010 & 14.145 $\pm$ 0.048 & 1.777 $\pm$ 0.003 & 8.742 $\pm$ 0.069 \\
    256 & 0.1 & 255.777 $\pm$ 0.002 & 4.559 $\pm$ 0.001 & 3.986 $\pm$ 0.013 & 13.975 $\pm$ 0.037 & 1.776 $\pm$ 0.001 & 8.626 $\pm$ 0.058 \\
    \midrule
    512 & 0 & 265.043 $\pm$ 0.619 & 4.375 $\pm$ 0.009 & 4.183 $\pm$ 0.008 & 15.529 $\pm$ 0.129 & 1.806 $\pm$ 0.005 & 10.292 $\pm$ 0.505 \\
    512 & 0.01 & 473.743 $\pm$ 0.910 & 4.310 $\pm$ 0.132 & 4.056 $\pm$ 0.006 & 14.486 $\pm$ 0.120 & 1.794 $\pm$ 0.007 & 9.321 $\pm$ 0.060 \\
    512 & 0.1 & 510.839 $\pm$ 0.011 & 4.469 $\pm$ 0.001 & 4.025 $\pm$ 0.021 & 14.281 $\pm$ 0.061 & 1.802 $\pm$ 0.004 & 9.056 $\pm$ 0.013 \\
    \bottomrule
    \end{tabular}%
    }
    \ifdim\wd0>\textwidth
        \PackageError{arxiv-curated}{Table exceeds text width: qwen-2p5-0p5b-rank-sweep}{Adjust the curated layout; do not scale it down.}
    \fi
    \typeout{CURATED-WIDTH qwen-2p5-0p5b-rank-sweep: \the\wd0; limit \the\textwidth}
    \leavevmode\box0\par
    \endgroup
    \caption{Mixed-domain LoRA results for Qwen2.5-0.5B at the reported rank and penalty settings.}
    \label{tab:arxiv-curated-qwen-2p5-0p5b-rank-sweep}
    \par\smallskip
\end{table*}

\begin{table*}[!tp]
    \centering
    \begingroup
    \small
    \setlength{\tabcolsep}{3pt}
    \setbox0=\hbox{%
    \begin{tabular}{@{}rrrrrrrr@{}}
    \toprule
    Rank & $\lambda$ & \shortstack[r]{eRank\\($\uparrow$)} & \shortstack[r]{Train PPL\\($\downarrow$)} & \shortstack[r]{StarCoder PPL\\($\downarrow$)} & \shortstack[r]{Wiki PPL\\($\downarrow$)} & \shortstack[r]{GSM8K PPL\\($\downarrow$)} & \shortstack[r]{WMT cs--en PPL\\($\downarrow$)} \\
    \midrule
    4 & 0 & 2.294 $\pm$ 0.042 & 4.549 $\pm$ 0.001 & 2.880 $\pm$ 0.004 & 9.124 $\pm$ 0.011 & 1.667 $\pm$ 0.001 & 3.894 $\pm$ 0.013 \\
    4 & 0.01 & 4.000 $\pm$ 0.000 & 4.569 $\pm$ 0.002 & 2.873 $\pm$ 0.003 & 9.186 $\pm$ 0.039 & 1.664 $\pm$ 0.003 & 3.893 $\pm$ 0.013 \\
    4 & 0.1 & 4.000 $\pm$ 0.000 & 4.836 $\pm$ 0.009 & 2.867 $\pm$ 0.005 & 9.155 $\pm$ 0.002 & 1.672 $\pm$ 0.001 & 3.884 $\pm$ 0.009 \\
    \midrule
    16 & 0 & 6.847 $\pm$ 0.092 & 4.211 $\pm$ 0.002 & 2.888 $\pm$ 0.007 & 9.148 $\pm$ 0.035 & 1.659 $\pm$ 0.002 & 3.830 $\pm$ 0.005 \\
    16 & 0.01 & 12.674 $\pm$ 0.066 & 4.524 $\pm$ 0.227 & 2.861 $\pm$ 0.005 & 9.102 $\pm$ 0.060 & 1.668 $\pm$ 0.002 & 3.824 $\pm$ 0.006 \\
    16 & 0.1 & 15.991 $\pm$ 0.004 & 4.219 $\pm$ 0.004 & 2.838 $\pm$ 0.006 & 9.019 $\pm$ 0.048 & 1.656 $\pm$ 0.001 & 3.794 $\pm$ 0.017 \\
    \midrule
    256 & 0 & 100.634 $\pm$ 0.592 & 3.612 $\pm$ 0.001 & 2.936 $\pm$ 0.023 & 9.539 $\pm$ 0.063 & 1.640 $\pm$ 0.005 & 3.829 $\pm$ 0.035 \\
    256 & 0.01 & 254.984 $\pm$ 0.025 & 3.632 $\pm$ 0.004 & 2.857 $\pm$ 0.003 & 9.210 $\pm$ 0.044 & 1.641 $\pm$ 0.001 & 3.878 $\pm$ 0.023 \\
    256 & 0.1 & 255.915 $\pm$ 0.002 & 4.038 $\pm$ 0.006 & 2.858 $\pm$ 0.007 & 9.114 $\pm$ 0.036 & 1.652 $\pm$ 0.002 & 3.970 $\pm$ 0.032 \\
    \midrule
    512 & 0 & 187.680 $\pm$ 3.891 & 3.561 $\pm$ 0.007 & 3.008 $\pm$ 0.036 & 9.708 $\pm$ 0.081 & 1.641 $\pm$ 0.003 & 3.991 $\pm$ 0.111 \\
    512 & 0.01 & 506.662 $\pm$ 0.037 & 3.569 $\pm$ 0.005 & 2.851 $\pm$ 0.010 & 9.185 $\pm$ 0.093 & 1.638 $\pm$ 0.003 & 3.856 $\pm$ 0.022 \\
    512 & 0.1 & 511.599 $\pm$ 0.005 & 3.628 $\pm$ 0.003 & 2.856 $\pm$ 0.004 & 9.162 $\pm$ 0.061 & 1.641 $\pm$ 0.003 & 3.849 $\pm$ 0.043 \\
    \bottomrule
    \end{tabular}%
    }
    \ifdim\wd0>\textwidth
        \PackageError{arxiv-curated}{Table exceeds text width: qwen-2p5-3b-it-rank-sweep}{Adjust the curated layout; do not scale it down.}
    \fi
    \typeout{CURATED-WIDTH qwen-2p5-3b-it-rank-sweep: \the\wd0; limit \the\textwidth}
    \leavevmode\box0\par
    \endgroup
    \caption{Mixed-domain LoRA results for Qwen2.5-3B-Instruct at the reported rank and penalty settings.}
    \label{tab:arxiv-curated-qwen-2p5-3b-it-rank-sweep}
    \par\smallskip
\end{table*}

\begin{table*}[!tp]
    \centering
    \begingroup
    \small
    \setlength{\tabcolsep}{3pt}
    \setbox0=\hbox{%
    \begin{tabular}{@{}rrrrrrrr@{}}
    \toprule
    Rank & $\lambda$ & \shortstack[r]{eRank\\($\uparrow$)} & \shortstack[r]{Train PPL\\($\downarrow$)} & \shortstack[r]{StarCoder PPL\\($\downarrow$)} & \shortstack[r]{Wiki PPL\\($\downarrow$)} & \shortstack[r]{GSM8K PPL\\($\downarrow$)} & \shortstack[r]{WMT cs--en PPL\\($\downarrow$)} \\
    \midrule
    4 & 0 & 2.680 $\pm$ 0.028 & 4.514 $\pm$ 0.002 & 4.071 $\pm$ 0.008 & 11.476 $\pm$ 0.043 & 1.899 $\pm$ 0.003 & 3.908 $\pm$ 0.007 \\
    4 & 0.01 & 4.000 $\pm$ 0.000 & 4.502 $\pm$ 0.005 & 4.023 $\pm$ 0.004 & 11.409 $\pm$ 0.010 & 1.902 $\pm$ 0.002 & 3.922 $\pm$ 0.005 \\
    4 & 0.1 & 4.000 $\pm$ 0.000 & 4.650 $\pm$ 0.001 & 4.016 $\pm$ 0.008 & 11.431 $\pm$ 0.008 & 1.912 $\pm$ 0.001 & 3.934 $\pm$ 0.013 \\
    \midrule
    16 & 0 & 8.969 $\pm$ 0.053 & 4.183 $\pm$ 0.001 & 4.119 $\pm$ 0.009 & 11.532 $\pm$ 0.024 & 1.884 $\pm$ 0.001 & 3.919 $\pm$ 0.007 \\
    16 & 0.001 & 14.144 $\pm$ 0.080 & 4.145 $\pm$ 0.001 & 4.021 $\pm$ 0.009 & 11.331 $\pm$ 0.015 & 1.881 $\pm$ 0.001 & 3.899 $\pm$ 0.001 \\
    16 & 0.01 & 15.993 $\pm$ 0.000 & 4.170 $\pm$ 0.002 & 3.982 $\pm$ 0.007 & 11.315 $\pm$ 0.002 & 1.881 $\pm$ 0.002 & 3.911 $\pm$ 0.005 \\
    16 & 0.1 & 16.000 $\pm$ 0.000 & 4.294 $\pm$ 0.005 & 3.971 $\pm$ 0.002 & 11.328 $\pm$ 0.017 & 1.887 $\pm$ 0.001 & 3.924 $\pm$ 0.004 \\
    \midrule
    32 & 0 & 17.755 $\pm$ 0.034 & 3.996 $\pm$ 0.000 & 4.124 $\pm$ 0.005 & 11.550 $\pm$ 0.030 & 1.869 $\pm$ 0.001 & 3.933 $\pm$ 0.004 \\
    32 & 0.001 & 27.620 $\pm$ 0.161 & 3.961 $\pm$ 0.003 & 4.005 $\pm$ 0.006 & 11.351 $\pm$ 0.018 & 1.865 $\pm$ 0.002 & 3.901 $\pm$ 0.004 \\
    32 & 0.01 & 31.971 $\pm$ 0.001 & 3.990 $\pm$ 0.002 & 3.967 $\pm$ 0.003 & 11.312 $\pm$ 0.001 & 1.869 $\pm$ 0.001 & 3.898 $\pm$ 0.008 \\
    32 & 0.1 & 31.998 $\pm$ 0.000 & 4.094 $\pm$ 0.002 & 3.970 $\pm$ 0.004 & 11.318 $\pm$ 0.010 & 1.873 $\pm$ 0.001 & 3.915 $\pm$ 0.008 \\
    \midrule
    64 & 0 & 35.952 $\pm$ 0.100 & 3.823 $\pm$ 0.003 & 4.106 $\pm$ 0.024 & 11.651 $\pm$ 0.031 & 1.855 $\pm$ 0.002 & 3.924 $\pm$ 0.014 \\
    64 & 0.001 & 53.691 $\pm$ 0.206 & 3.783 $\pm$ 0.001 & 4.008 $\pm$ 0.003 & 11.384 $\pm$ 0.034 & 1.857 $\pm$ 0.002 & 3.912 $\pm$ 0.002 \\
    64 & 0.01 & 63.889 $\pm$ 0.002 & 3.815 $\pm$ 0.002 & 3.954 $\pm$ 0.009 & 11.347 $\pm$ 0.014 & 1.862 $\pm$ 0.002 & 3.891 $\pm$ 0.002 \\
    64 & 0.1 & 63.993 $\pm$ 0.000 & 3.898 $\pm$ 0.001 & 3.951 $\pm$ 0.003 & 11.281 $\pm$ 0.009 & 1.863 $\pm$ 0.000 & 3.896 $\pm$ 0.010 \\
    \midrule
    128 & 0 & 72.138 $\pm$ 0.122 & 3.661 $\pm$ 0.001 & 4.102 $\pm$ 0.006 & 11.726 $\pm$ 0.018 & 1.848 $\pm$ 0.002 & 3.937 $\pm$ 0.014 \\
    128 & 0.01 & 127.566 $\pm$ 0.008 & 3.650 $\pm$ 0.002 & 3.962 $\pm$ 0.013 & 11.342 $\pm$ 0.014 & 1.854 $\pm$ 0.001 & 3.901 $\pm$ 0.003 \\
    128 & 0.1 & 127.970 $\pm$ 0.000 & 3.717 $\pm$ 0.002 & 3.958 $\pm$ 0.004 & 11.323 $\pm$ 0.017 & 1.858 $\pm$ 0.003 & 3.906 $\pm$ 0.001 \\
    \midrule
    256 & 0 & 143.539 $\pm$ 0.657 & 3.518 $\pm$ 0.005 & 4.109 $\pm$ 0.024 & 11.828 $\pm$ 0.024 & 1.853 $\pm$ 0.003 & 3.958 $\pm$ 0.019 \\
    256 & 0.01 & 253.483 $\pm$ 0.062 & 3.514 $\pm$ 0.002 & 3.971 $\pm$ 0.006 & 11.425 $\pm$ 0.036 & 1.849 $\pm$ 0.002 & 3.917 $\pm$ 0.008 \\
    256 & 0.1 & 255.855 $\pm$ 0.001 & 3.562 $\pm$ 0.002 & 3.974 $\pm$ 0.011 & 11.422 $\pm$ 0.024 & 1.847 $\pm$ 0.001 & 3.921 $\pm$ 0.005 \\
    \midrule
    512 & 0 & 265.793 $\pm$ 1.573 & 3.470 $\pm$ 0.006 & 4.251 $\pm$ 0.006 & 12.180 $\pm$ 0.157 & 1.855 $\pm$ 0.002 & 4.061 $\pm$ 0.021 \\
    512 & 0.01 & 497.274 $\pm$ 0.676 & 3.444 $\pm$ 0.007 & 4.016 $\pm$ 0.004 & 11.655 $\pm$ 0.018 & 1.861 $\pm$ 0.007 & 3.975 $\pm$ 0.012 \\
    512 & 0.1 & 511.284 $\pm$ 0.009 & 3.458 $\pm$ 0.002 & 3.992 $\pm$ 0.004 & 11.577 $\pm$ 0.013 & 1.857 $\pm$ 0.002 & 3.951 $\pm$ 0.011 \\
    \bottomrule
    \end{tabular}%
    }
    \ifdim\wd0>\textwidth
        \PackageError{arxiv-curated}{Table exceeds text width: llama-3p2-3b-it-rank-sweep}{Adjust the curated layout; do not scale it down.}
    \fi
    \typeout{CURATED-WIDTH llama-3p2-3b-it-rank-sweep: \the\wd0; limit \the\textwidth}
    \leavevmode\box0\par
    \endgroup
    \caption{Mixed-domain LoRA results for Llama-3.2-3B-Instruct at the reported rank and penalty settings.}
    \label{tab:arxiv-curated-llama-3p2-3b-it-rank-sweep}
    \par\smallskip
\end{table*}

\subsubsection{Interpretation at extreme ranks}
\label{subsec:extreme-ranks}

\paragraph{Rank one has no spectral utilization margin.}
For a nonzero rank-one update, there is one nonzero singular value, so both normalized spectra have a single entry equal to $1$. Consequently, $\erank(\Delta W)=\rerank(\Delta W)=1$, independently of its magnitude and direction. The exact capacity term is therefore constant over nonzero rank-one updates and supplies no preference among them. This is an algebraic boundary case; we report no rank-one experiment. Numerical stabilization near the zero update is separate from this statement about the exact objective.

\paragraph{Small ranks limit the available benefit.}
Table~\ref{tab:arxiv-curated-qwen-2p5-0p5b-rank-sweep} illustrates the smallest measured mixed-domain rank, $r{=}4$. With $\lambda{=}0.1$, the reported mean $\erank$ increases from $3.014$ to $4.000$, while mean Wiki PPL changes from $13.615$ to $13.564$ and mean StarCoder PPL from $3.926$ to $3.942$; mean Train PPL increases from $6.096$ to $6.595$. Thus saturating the spectral rank can accompany only small, mixed changes in retention and worse training fit. The penalty redistributes energy among at most $r$ nonzero singular directions; it cannot increase the configured rank budget.

\paragraph{Maximal matrix rank still permits spectral concentration.}
For an adapted matrix of shape $d_{\mathrm{out}}\times d_{\mathrm{in}}$, the rank ceiling is $\min\{r,d_{\mathrm{out}},d_{\mathrm{in}}\}$. The $r{=}896$ mixed-domain comparison reaches the maximal matrix rank for the targeted Qwen2.5-0.5B projections, but standard LoRA realizes a mean effective rank of $360.537$. In Table~\ref{tab:mixed-task-baseline-comparison}, $\lambda{=}0.1$ raises the mean to $887.772$, lowers mean Wiki PPL from $19.601$ to $14.648$ and mean StarCoder PPL from $5.815$ to $4.098$, and improves mean Train PPL from $4.773$ to $4.525$. These are comparisons within the rank-$896$ setting; together with the smaller-rank sweep, they show that the scope for improving spectral utilization depends on the rank budget.

At maximal rank, LoRA can represent any update to each targeted matrix, but its factorized optimization differs from optimizing that matrix directly. The factors contain $r(d_{\mathrm{in}}+d_{\mathrm{out}})$ parameters per matrix, compared with $d_{\mathrm{in}}d_{\mathrm{out}}$ for a dense update; at maximal matrix rank the factor count exceeds the dense count. This regime tests spectral utilization without providing parameter savings for those matrices.

\begin{samepage}
\noindent Increasing $r$ beyond the matrix-rank ceiling does not expand the space of representable updates.\par
\end{samepage}

\paragraph{Maximum effective rank is not a performance optimum.}
At $r{=}256$, increasing $\lambda$ from $0.01$ to $0.1$ raises the reported mean $\erank$ from $250.540$ to $255.777$, but mean Train PPL worsens from $4.465$ to $4.559$, even as mean Wiki PPL improves from $14.145$ to $13.975$ (Table~\ref{tab:arxiv-curated-qwen-2p5-0p5b-rank-sweep}). Effective rank describes spectral utilization, while the task loss and retention probes determine whether that utilization is useful. Penalty strength must therefore be chosen for the desired learning--retention balance.

\subsection{Generalization across model families and scales}
\label{subsec:supplementary-model-comparisons}

Tables~\ref{tab:arxiv-curated-olmo-2-13b-it-rank-sweep}, \ref{tab:arxiv-curated-olmo-2-32b-it-rank-sweep}, and~\ref{tab:arxiv-curated-pythia-2p8b-rank-sweep} extend the mixed-domain rank and penalty comparisons to OLMo-2-13B-Instruct, OLMo-2-32B-Instruct, and Pythia-2.8B. The reported settings show increased effective-rank utilization, while the changes in perplexity depend on the model, evaluation probe, and penalty weight.

\begin{table*}[!tp]
    \centering
    \begingroup
    \small
    \setlength{\tabcolsep}{3pt}
    \setbox0=\hbox{%
    \begin{tabular}{@{}rrrrrrrr@{}}
    \toprule
    Rank & $\lambda$ & \shortstack[r]{eRank\\($\uparrow$)} & \shortstack[r]{Train PPL\\($\downarrow$)} & \shortstack[r]{StarCoder PPL\\($\downarrow$)} & \shortstack[r]{Wiki PPL\\($\downarrow$)} & \shortstack[r]{GSM8K PPL\\($\downarrow$)} & \shortstack[r]{WMT cs--en PPL\\($\downarrow$)} \\
    \midrule
    128 & 0 & 66.698 & 3.157 & 2.406 & 5.694 & 1.861 & 4.057 \\
    128 & 0.0001 & 67.741 & 3.146 & 2.406 & 5.692 & 1.853 & 4.049 \\
    128 & 0.001 & 111.389 & 3.123 & 2.345 & 5.629 & 1.856 & 4.009 \\
    128 & 0.01 & 127.737 & 3.148 & 2.318 & 5.625 & 1.855 & 3.906 \\
    \midrule
    256 & 0 & 132.729 & 3.731 & 2.336 & 5.653 & 1.901 & 4.146 \\
    256 & 0.0001 & 139.286 & 3.720 & 2.334 & 5.659 & 1.902 & 4.099 \\
    256 & 0.001 & 202.409 & 3.723 & 2.318 & 5.637 & 1.897 & 4.059 \\
    256 & 0.01 & 254.285 & 3.719 & 2.299 & 5.614 & 1.910 & 3.951 \\
    256 & 0.1 & 255.934 & 3.738 & 2.313 & 5.651 & 1.926 & 3.893 \\
    \midrule
    512 & 0 & 260.827 & 3.809 & 2.352 & 5.701 & 1.877 & 4.036 \\
    512 & 0.0001 & 280.339 & 3.793 & 2.350 & 5.669 & 1.873 & 4.095 \\
    512 & 0.001 & 370.021 & 3.781 & 2.318 & 5.669 & 1.880 & 4.085 \\
    512 & 0.01 & 506.224 & 3.767 & 2.303 & 5.608 & 1.887 & 3.928 \\
    512 & 0.1 & 511.356 & 3.742 & 2.304 & 5.623 & 1.889 & 3.902 \\
    \bottomrule
    \end{tabular}%
    }
    \ifdim\wd0>\textwidth
        \PackageError{arxiv-curated}{Table exceeds text width: olmo-2-13b-it-rank-sweep}{Adjust the curated layout; do not scale it down.}
    \fi
    \typeout{CURATED-WIDTH olmo-2-13b-it-rank-sweep: \the\wd0; limit \the\textwidth}
    \leavevmode\box0\par
    \endgroup
    \caption{Mixed-domain LoRA rank and penalty sweep for OLMo-2-13B-Instruct.}
    \label{tab:arxiv-curated-olmo-2-13b-it-rank-sweep}
    \par\smallskip
\end{table*}

\begin{table*}[!tp]
    \centering
    \begingroup
    \small
    \setlength{\tabcolsep}{3pt}
    \setbox0=\hbox{%
    \begin{tabular}{@{}rrrrrrrr@{}}
    \toprule
    Rank & $\lambda$ & \shortstack[r]{eRank\\($\uparrow$)} & \shortstack[r]{Train PPL\\($\downarrow$)} & \shortstack[r]{StarCoder PPL\\($\downarrow$)} & \shortstack[r]{Wiki PPL\\($\downarrow$)} & \shortstack[r]{GSM8K PPL\\($\downarrow$)} & \shortstack[r]{WMT cs--en PPL\\($\downarrow$)} \\
    \midrule
    64 & 0 & 25.830 & 2.900 & 2.183 & 4.305 & 1.813 & 3.503 \\
    64 & 0.001 & 59.852 & 2.880 & 2.197 & 4.261 & 1.811 & 3.427 \\
    64 & 0.01 & 63.950 & 2.892 & 2.219 & 4.263 & 1.820 & 3.467 \\
    \midrule
    128 & 0 & 46.339 & 2.879 & 2.192 & 4.318 & 1.822 & 3.513 \\
    128 & 0.0001 & 52.728 & 2.877 & 2.199 & 4.318 & 1.803 & 3.482 \\
    128 & 0.001 & 106.312 & 2.838 & 2.218 & 4.281 & 1.825 & 3.430 \\
    128 & 0.01 & 127.830 & 2.836 & 2.229 & 4.263 & 1.809 & 3.455 \\
    \midrule
    256 & 0 & 88.660 & 2.879 & 2.231 & 4.366 & 1.834 & 3.600 \\
    256 & 0.001 & 156.287 & 2.829 & 2.213 & 4.338 & 1.810 & 3.472 \\
    256 & 0.01 & 254.762 & 2.836 & 2.233 & 4.281 & 1.831 & 3.423 \\
    \bottomrule
    \end{tabular}%
    }
    \ifdim\wd0>\textwidth
        \PackageError{arxiv-curated}{Table exceeds text width: olmo-2-32b-it-rank-sweep}{Adjust the curated layout; do not scale it down.}
    \fi
    \typeout{CURATED-WIDTH olmo-2-32b-it-rank-sweep: \the\wd0; limit \the\textwidth}
    \leavevmode\box0\par
    \endgroup
    \caption{Mixed-domain LoRA rank and penalty sweep for OLMo-2-32B-Instruct.}
    \label{tab:arxiv-curated-olmo-2-32b-it-rank-sweep}
    \par\smallskip
\end{table*}

\begin{table*}[!tp]
    \centering
    \begingroup
    \small
    \setlength{\tabcolsep}{3pt}
    \setbox0=\hbox{%
    \begin{tabular}{@{}rrrrrrr@{}}
    \toprule
    Rank & $\lambda$ & \shortstack[r]{eRank\\($\uparrow$)} & \shortstack[r]{Train PPL\\($\downarrow$)} & \shortstack[r]{StarCoder PPL\\($\downarrow$)} & \shortstack[r]{Wiki PPL\\($\downarrow$)} & \shortstack[r]{WMT cs--en PPL\\($\downarrow$)} \\
    \midrule
    128 & 0 & 45.837 & 6.361 & 2.328 & 12.830 & 6.065 \\
    128 & 0.01 & 125.942 & 5.952 & 2.023 & 11.932 & 5.850 \\
    128 & 0.1 & 127.959 & 6.069 & 2.034 & 11.938 & 5.799 \\
    \midrule
    256 & 0 & 97.686 & 5.933 & 2.376 & 12.882 & 6.427 \\
    256 & 0.01 & 248.776 & 5.823 & 2.111 & 12.315 & 5.997 \\
    256 & 0.1 & 255.853 & 5.799 & 2.052 & 12.092 & 6.022 \\
    \midrule
    512 & 0 & 193.826 & 5.668 & 2.529 & 13.812 & 6.603 \\
    512 & 0.01 & 473.058 & 5.622 & 2.136 & 12.655 & 6.205 \\
    512 & 0.1 & 511.353 & 5.658 & 2.081 & 12.256 & 6.161 \\
    \midrule
    2560 & 0 & 581.591 & 7.821 & 16.568 & 24.901 & 19.201 \\
    2560 & 0.01 & 1180.866 & 6.053 & 3.592 & 17.055 & 9.474 \\
    2560 & 0.1 & 2329.172 & 6.517 & 3.111 & 15.284 & 8.614 \\
    \bottomrule
    \end{tabular}%
    }
    \ifdim\wd0>\textwidth
        \PackageError{arxiv-curated}{Table exceeds text width: pythia-2p8b-rank-sweep}{Adjust the curated layout; do not scale it down.}
    \fi
    \typeout{CURATED-WIDTH pythia-2p8b-rank-sweep: \the\wd0; limit \the\textwidth}
    \leavevmode\box0\par
    \endgroup
    \caption{Mixed-domain LoRA rank and penalty sweep for Pythia-2.8B.}
    \label{tab:arxiv-curated-pythia-2p8b-rank-sweep}
    \par\smallskip
\end{table*}

\subsection{Single-task baseline comparisons}
\label{subsec:single-task-baselines}

We next separate the effects of adaptation to individual tasks from those of mixed-domain fine-tuning. The following comparisons are grouped by dataset and, within each dataset, by model size. Training-task perplexity measures fit to the adaptation data, while StarCoder and Wiki perplexities probe retention of the reference distributions.

\paragraph{Aya Hausa instruction following.}
Tables~\ref{tab:aya-hausa-baseline-comparison} and~\ref{tab:aya-hausa-baseline-comparison-qwen-25-3b} compare standard LoRA, penalty settings, and alternative adapter parameterizations on Hausa instruction following.

\begin{table}[tbp]
\centering
\small
\setlength{\tabcolsep}{3pt}
\begin{adjustbox}{max width=\columnwidth}
\begin{tabular}{@{}lrrrr@{}}
\toprule
Method &  \shortstack{eRank\\($\uparrow$)} & \shortstack{Train\\PPL ($\downarrow$)} & \shortstack{StarCoder\\PPL ($\downarrow$)} & \shortstack{Wiki\\PPL ($\downarrow$)} \\
\midrule
\multicolumn{5}{l}{\textit{Standard}} \\
Dense FT                     & 805.510        & 20.977    & 4.243 & 23.393 \\
Standard LoRA                       & 340.517       & 22.580     & 3.998 & 21.098 \\
\midrule
\multicolumn{5}{l}{\textit{LoRA + \penaltyname{} (Ours)}} \\
LoRA ($\lambda=0.01$)         & 524.953       & 22.324 & 3.881 & 20.040 \\
LoRA ($\lambda=1.0$)          & 667.507       & 22.814 & 3.731 & 18.734 \\
\midrule
\multicolumn{5}{l}{\textit{LoRA variants}} \\
VeRA                                & 594.675       & 145.805   & 3.513 & 16.323 \\
TinyLoRA                              & 681.668       & 146.773   & 3.514 & 16.325 \\
LoRA-XS                             & 495.382       & 23.519    & 5.615 & 30.420 \\
SVF                                 & 735.685       & 124.129   & 3.568 & 16.491 \\
\bottomrule
\end{tabular}
\end{adjustbox}
\caption{Dense FT denotes dense down-projection fine-tuning.
    Aya Hausa instruction-following adaptation for Qwen2.5-0.5B, with MLP down-projection adapters ($r{=}896$, $\alpha{=}1792$). Train PPL measures fit to the adaptation task; StarCoder and Wiki PPL are held-out retention probes. Dense down-projection fine-tuning updates only the targeted matrices.
}
\label{tab:aya-hausa-baseline-comparison}
\end{table}

\begin{table}[tbp]
    \centering
    \small
\setlength{\tabcolsep}{3pt}
    \begin{adjustbox}{max width=\columnwidth}
    \begin{tabular}{@{}lrrrr@{}}
    \toprule
    Method &  \shortstack{eRank\\($\uparrow$)} & \shortstack{Train\\PPL ($\downarrow$)} & \shortstack{StarCoder\\PPL ($\downarrow$)} & \shortstack{Wiki\\PPL ($\downarrow$)} \\
    \midrule
    \multicolumn{5}{l}{\textit{Standard}} \\
Dense FT & 1841.092 & 24.753 & 2.961 & 11.008 \\
Standard LoRA &  522.523 & 16.819 & 3.179 & 14.540 \\
    \midrule
    \multicolumn{5}{l}{\textit{LoRA + \penaltyname{} (Ours)}} \\
LoRA ($\lambda=0.01$) & 1160.793 & 16.614 & 2.974 & 12.713 \\
LoRA ($\lambda=0.1$) & 1963.771 & 18.833 & 2.959 & 12.398 \\
LoRA ($\lambda=1.0$) & 2045.368 & 20.089 & 2.955 & 12.026 \\
    \midrule
    \multicolumn{5}{l}{\textit{LoRA variants}} \\
PiSSA & 1737.381 & 15.737 & 3.600 & 16.839 \\
VeRA & 1347.394 & 77.322 & 2.867 & 11.863 \\
TinyLoRA & 1548.345 & 78.206 & 2.870 & 11.880 \\
SVF & 1669.220 & 60.631 & 2.853 & 11.753 \\
    \bottomrule
    \end{tabular}
    \end{adjustbox}
    \caption{Dense FT denotes dense down-projection fine-tuning.
    Aya Hausa instruction-following adaptation for Qwen2.5-3B-Instruct, with MLP down-projection adapters ($r{=}2048$, $\alpha{=}4096$). Train PPL measures fit to the adaptation task; StarCoder and Wiki PPL are held-out retention probes. Dense down-projection fine-tuning updates only the targeted matrices.
}
\label{tab:aya-hausa-baseline-comparison-qwen-25-3b}
\end{table}

\paragraph{GSM8K mathematical reasoning.}
Tables~\ref{tab:gsm8k-baseline-comparison} and~\ref{tab:gsm8k-baseline-comparison-qwen-25-3b} give the corresponding single-task GSM8K comparisons for Qwen2.5-0.5B and Qwen2.5-3B-Instruct. They should be distinguished from the held-out GSM8K probe reported after mixed-domain training.

\begin{table}[tbp]
\centering
\small
\setlength{\tabcolsep}{3pt}
\begin{adjustbox}{max width=\columnwidth}
\begin{tabular}{@{}lrrrr@{}}
\toprule
Method                           & \shortstack{eRank\\($\uparrow$)}   & \shortstack{Train\\PPL ($\downarrow$)} & \shortstack{StarCoder\\PPL ($\downarrow$)} & \shortstack{Wiki\\PPL ($\downarrow$)} \\ \midrule
\multicolumn{5}{l}{\textit{Standard}} \\
Dense FT                  & 839.449      & 1.773   & 3.760        & 18.726 \\
Standard LoRA                    & 439.390       & 1.826   & 3.794        & 18.846 \\
\midrule
\multicolumn{5}{l}{\textit{LoRA + \penaltyname{} (Ours)}} \\
LoRA ($\lambda=0.01$)      & 766.796      & 1.808   & 3.655        & 17.753 \\
LoRA ($\lambda=0.1$)       & 891.192      & 1.795   & 3.629        & 17.397 \\
LoRA ($\lambda=1.0$)       & 895.619      & 1.776   & 3.606        & 17.100 \\
\midrule
\multicolumn{5}{l}{\textit{LoRA variants}} \\
VeRA                             & 648.593      & 2.345   & 3.514        & 16.309 \\
TinyLoRA                         & 681.665      & 2.367   & 3.515        & 16.326 \\
SVF                              & 782.616      & 1.970   & 3.571        & 16.665 \\
\bottomrule
\end{tabular}
\end{adjustbox}
\caption{Dense FT denotes dense down-projection fine-tuning.
    Single-task GSM8K adaptation for Qwen2.5-0.5B, with MLP down-projection adapters ($r{=}896$, $\alpha{=}1792$). Train PPL measures fit to the adaptation task; StarCoder and Wiki PPL are held-out retention probes. Dense down-projection fine-tuning updates only the targeted matrices. Unpenalized PiSSA and LoRA-XS results appear in Table~\ref{tab:gsm8k-pissaloraxs-rank-penalty}.
}
\label{tab:gsm8k-baseline-comparison}
\end{table}

\begin{table}[tbp]
    \centering
    \small
\setlength{\tabcolsep}{3pt}
    \begin{adjustbox}{max width=\columnwidth}
    \begin{tabular}{@{}lrrrr@{}}
    \toprule
    Method                           & \shortstack{eRank\\($\uparrow$)}   & \shortstack{Train\\PPL ($\downarrow$)} & \shortstack{StarCoder\\PPL ($\downarrow$)} & \shortstack{Wiki\\PPL ($\downarrow$)} \\ \midrule
    \multicolumn{5}{l}{\textit{Standard}} \\
Standard LoRA                    & 452.096      & 1.880     & 15.795       & 27.333 \\
    \midrule
    \multicolumn{5}{l}{\textit{LoRA + \penaltyname{} (Ours)}} \\
LoRA ($\lambda=0.01$)      & 754.421      & 1.722     & 5.370        & 17.178 \\
LoRA ($\lambda=0.1$)       & 1999.507     & 1.646     & 2.910        & 12.009 \\
LoRA ($\lambda=1.0$)       & 2030.334     & 1.636     & 2.907        & 11.988 \\
    \midrule
    \multicolumn{5}{l}{\textit{LoRA variants}} \\
PiSSA                            & 1727.320     & 1.752     & 5.488        & 18.881 \\
    \bottomrule
    \end{tabular}
    \end{adjustbox}
    \caption{
    Single-task GSM8K adaptation for Qwen2.5-3B-Instruct, with MLP down-projection adapters ($r{=}2048$, $\alpha{=}4096$). Train PPL measures fit to the adaptation task; StarCoder and Wiki PPL are held-out retention probes.
}
\label{tab:gsm8k-baseline-comparison-qwen-25-3b}
\end{table}

\paragraph{WMT19 Czech--English translation.}
Tables~\ref{tab:wmt19-cs-en-baseline-comparison} and~\ref{tab:wmt19-cs-en-baseline-comparison-qwen-25-3b} compare adaptation methods for Qwen2.5-0.5B and Qwen2.5-3B-Instruct on Czech--English translation. Any row-specific schedule differences are specified in the table caption.

\begin{table}[tbp]
\centering
\small
\setlength{\tabcolsep}{3pt}
\begin{adjustbox}{max width=\columnwidth}
\begin{tabular}{@{}lrrrr@{}}
\toprule
Method                      & \shortstack{eRank\\($\uparrow$)}   & \shortstack{Train\\PPL ($\downarrow$)}  & \shortstack{StarCoder\\PPL ($\downarrow$)}  & \shortstack{Wiki\\PPL ($\downarrow$)} \\ \midrule
\multicolumn{5}{l}{\textit{Standard}} \\
Dense FT             & 850.116 & 3.807     & 4.699         & 28.886 \\
Standard LoRA               & 385.658 & 4.365     & 9.003         & 49.387 \\
\midrule
\multicolumn{5}{l}{\textit{LoRA + \penaltyname{} (Ours)}} \\
LoRA ($\lambda=0.01$)  & 627.248 & 3.970 & 4.170 & 25.886 \\
LoRA ($\lambda=0.1$)   & 884.367 & 5.138 & 5.809 & 37.391 \\
LoRA ($\lambda=1.0$)  & 894.984 & 4.520      & 5.587         & 36.460 \\
\midrule
\multicolumn{5}{l}{\textit{LoRA variants}} \\
PiSSA      & 839.492 & 3.887 & 5.301 & 33.887 \\
TinyLoRA   & 681.665 & 8.551 & 3.514 & 16.327 \\
VeRA       & 670.908 & 8.353 & 3.513 & 16.328 \\
SVF        & 827.278 & 7.312 & 3.603 & 17.216 \\
LoRA-XS    & 614.227 & 4.381 & 13.762 & 53.907 \\
\bottomrule
\end{tabular}
\end{adjustbox}
\caption{Dense FT denotes dense down-projection fine-tuning.
    WMT19 Czech--English translation adaptation for Qwen2.5-0.5B, with MLP down-projection adapters ($r{=}896$, $\alpha{=}1792$). Train PPL measures fit to the adaptation task; StarCoder and Wiki PPL are held-out retention probes. Dense down-projection fine-tuning updates only the targeted matrices.
}
\label{tab:wmt19-cs-en-baseline-comparison}
\end{table}

\begin{table}[tbp]
\centering
\small
\setlength{\tabcolsep}{3pt}
\begin{adjustbox}{max width=\columnwidth}
\begin{tabular}{@{}lrrrr@{}}
\toprule
Method                      & \shortstack{eRank\\($\uparrow$)}   & \shortstack{Train\\PPL ($\downarrow$)}  & \shortstack{StarCoder\\PPL ($\downarrow$)}  & \shortstack{Wiki\\PPL ($\downarrow$)} \\ \midrule
\multicolumn{5}{l}{\textit{Standard}} \\
Dense FT             & 1863.290 & 2.921     & 2.939         & 12.425 \\
Standard LoRA               & 494.069 & 3.667     & 5.191         & 22.857 \\
\midrule
\multicolumn{5}{l}{\textit{LoRA + \penaltyname{} (Ours)}} \\
LoRA ($\lambda=0.01$)  & 1140.120 & 3.250     & 3.473         & 18.344 \\
LoRA ($\lambda=0.1$)   & 1950.694 & 3.018     & 4.709         & 24.844 \\
LoRA ($\lambda=1.0$)  & 2046.575 & 2.954     & 2.973         & 12.535 \\
\midrule
\multicolumn{5}{l}{\textit{LoRA variants}} \\
PiSSA                       & 1772.069 & 3.570     & 3.126         & 14.704 \\
TinyLoRA                  & 1548.343 & 5.560     & 2.869         & 11.870 \\
VeRA                       & 1486.087 & 5.038     & 2.867         & 11.857 \\
SVF                        & 1828.385 & 4.031     & 2.872         & 11.783 \\
LoRA-XS                    & 1309.694 & 3.625     & 20.329        & 40.168 \\
\bottomrule
\end{tabular}
\end{adjustbox}
        \caption{Dense FT denotes dense down-projection fine-tuning.
    WMT19 Czech--English translation adaptation for Qwen2.5-3B-Instruct, with MLP down-projection adapters with rank=2048.
}
\label{tab:wmt19-cs-en-baseline-comparison-qwen-25-3b}
\end{table}

\subsection{Single-task rank and penalty sensitivity}
\label{subsec:single-task-rank-sweeps}

The single-task sweeps vary the adapter rank and penalty weight for Qwen2.5-0.5B and Qwen2.5-7B-Instruct. These model sizes differ from the 0.5B and 3B baseline comparisons above. We retain this distinction explicitly rather than treating the tables as one complete model-by-rank grid.

\paragraph{Aya Hausa.}
Tables~\ref{tab:aya-hausa-lora-rank-sweep-qwen-25-05b} and~\ref{tab:aya-hausa-lora-rank-sweep-qwen-25-7b} report training-task fit and reference-distribution perplexities at the available ranks and penalty weights. Interpret changes within each model and rank.

\begin{table*}[t]
    \centering
    \small
    \begin{adjustbox}{max width=\textwidth}
    \begin{tabular}{@{}lrrrrr@{}}
    \toprule
    rank & $\lambda$ & eRank ($\uparrow$) & Train PPL ($\downarrow$) & Wiki PPL ($\downarrow$) & StarCoder PPL ($\downarrow$) \\ \midrule
    \multicolumn{6}{l}{\textit{LoRA rank 512}} \\
    512 & 0 & 229.271 & 23.578 & 19.643 & 3.905 \\
    512 & 0.01 & 383.111 & 23.449 & 18.746 & 3.734 \\
    512 & 0.1 & 393.541 & 23.651 & 18.277 & 3.691 \\
    512 & 1.0 & 511.799 & 28.556 & 18.009 & 3.720 \\
    \midrule
    \multicolumn{6}{l}{\textit{LoRA rank 256}} \\
    256 & 0 & 124.345 & 25.593 & 18.226 & 3.732 \\
    256 & 0.01 & 131.174 & 25.488 & 18.292 & 3.713 \\
    256 & 0.1 & 210.333 & 25.718 & 17.686 & 3.637 \\
    256 & 1.0 & 240.983 & 26.139 & 17.471 & 3.628 \\
    \midrule
    \multicolumn{6}{l}{\textit{LoRA rank 128}} \\
    128 & 0 & 62.911 & 28.346 & 17.962 & 3.694 \\
    128 & 0.01 & 70.402 & 28.223 & 17.712 & 3.660 \\
    128 & 0.1 & 102.260 & 28.552 & 17.436 & 3.608 \\
    128 & 1.0 & 125.775 & 29.007 & 17.359 & 3.611 \\
    \midrule
    \multicolumn{6}{l}{\textit{LoRA rank 64}} \\
    64 & 0 & 30.952 & 32.222 & 17.705 & 3.656 \\
    64 & 0.01 & 35.640 & 31.852 & 17.625 & 3.647 \\
    64 & 0.1 & 51.719 & 32.145 & 17.369 & 3.603 \\
    64 & 1.0 & 62.994 & 32.672 & 17.286 & 3.602 \\
    \midrule
    \multicolumn{6}{l}{\textit{LoRA rank 32}} \\
    32 & 0 & 15.831 & 36.577 & 17.637 & 3.651 \\
    32 & 0.01 & 17.621 & 36.593 & 17.428 & 3.636 \\
    32 & 0.1 & 25.542 & 36.689 & 17.245 & 3.599 \\
    32 & 1.0 & 31.582 & 37.286 & 17.138 & 3.594 \\
    \midrule
    \multicolumn{6}{l}{\textit{LoRA rank 16}} \\
    16 & 0 & 7.960 & 41.687 & 17.432 & 3.635 \\
    16 & 0.01 & 8.927 & 41.823 & 17.373 & 3.624 \\
    16 & 0.1 & 12.788 & 41.862 & 17.125 & 3.596 \\
    16 & 1.0 & 15.812 & 42.358 & 17.103 & 3.588 \\
    \midrule
    \multicolumn{6}{l}{\textit{LoRA rank 8}} \\
    8 & 0 & 4.253 & 48.162 & 17.362 & 3.642 \\
    8 & 0.01 & 4.849 & 47.708 & 17.374 & 3.618 \\
    8 & 0.1 & 6.515 & 47.886 & 17.267 & 3.613 \\
    8 & 1.0 & 7.933 & 48.202 & 17.003 & 3.582 \\
    \midrule
    \multicolumn{6}{l}{\textit{LoRA rank 4}} \\
    4 & 0 & 2.620 & 54.124 & 17.398 & 3.619 \\
    4 & 0.01 & 2.831 & 55.016 & 17.192 & 3.609 \\
    4 & 0.1 & 3.478 & 54.783 & 17.169 & 3.609 \\
    4 & 1.0 & 3.950 & 55.126 & 17.070 & 3.586 \\
    \bottomrule
    \end{tabular}
    \end{adjustbox}
    \caption{LoRA rank sweep for Qwen2.5-0.5B on Aya Hausa instruction following.}
\label{tab:aya-hausa-lora-rank-sweep-qwen-25-05b}
\end{table*}

\paragraph{GSM8K.}
Tables~\ref{tab:gsm8k-lora-rank-sweep-qwen-25-05b} and~\ref{tab:gsm8k-lora-rank-sweep-qwen-25-7b} show how rank and penalty weight affect mathematical adaptation and retention. Missing standard-LoRA or penalty settings are left unreported rather than inferred from neighboring ranks.

\begin{table*}[t]
\renewenvironment{table}[1][]{}{}%
\begin{minipage}[t]{\columnwidth}
  \vspace{0pt}%
  \begin{table}[t]
    \centering
    \small
    \begin{adjustbox}{max width=\columnwidth}
    \begin{tabular}{@{}lrrrrr@{}}
    \toprule
    rank & $\lambda$ & eRank ($\uparrow$) & Train PPL ($\downarrow$) & Wiki PPL ($\downarrow$) & StarCoder PPL ($\downarrow$) \\ \midrule
    \multicolumn{6}{l}{\textit{LoRA rank 4}} \\
    4 & 0 & 1.906 & 22.892 & 10.239 & 2.527 \\
    4 & 0.01 & 2.023 & 22.899 & 10.255 & 2.528 \\
    4 & 0.1 & 3.285 & 22.852 & 10.236 & 2.518 \\
    4 & 1.0 & 3.998 & 22.839 & 10.258 & 2.517 \\
    \midrule
    \multicolumn{6}{l}{\textit{LoRA rank 32}} \\
    32 & 0 & 11.403 & 19.147 & 10.419 & 2.561 \\
    32 & 0.01 & 13.352 & 19.149 & 10.324 & 2.542 \\
    32 & 0.1 & 25.044 & 19.127 & 10.222 & 2.519 \\
    32 & 1.0 & 31.776 & 19.057 & 10.225 & 2.512 \\
    \midrule
    \multicolumn{6}{l}{\textit{LoRA rank 64}} \\
    64 & 0 & 23.164 & 17.946 & 10.443 & 2.574 \\
    64 & 0.1 & 47.636 & 17.919 & 10.251 & 2.521 \\
    64 & 1.0 & 63.314 & 17.928 & 10.230 & 2.512 \\
    \midrule
    \multicolumn{6}{l}{\textit{LoRA rank 256}} \\
    256 & 0 & 93.822 & 15.886 & 10.703 & 2.607 \\
    256 & 0.01 & 101.891 & 15.881 & 10.424 & 2.559 \\
    256 & 0.1 & 152.213 & 15.883 & 10.288 & 2.519 \\
    256 & 1.0 & 252.009 & 16.039 & 10.253 & 2.514 \\
    \bottomrule
    \end{tabular}
    \end{adjustbox}
    \caption{LoRA rank and penalty-weight sweep on Aya Hausa for Qwen2.5-7B-Instruct.
}
\label{tab:aya-hausa-lora-rank-sweep-qwen-25-7b}
\end{table}

  \vspace{1em}
  \begin{table}[t]
    \centering
    \small
    \begin{adjustbox}{max width=\columnwidth}
    \begin{tabular}{@{}lrrrrr@{}}
    \toprule
    rank & $\lambda$ & eRank ($\uparrow$) & Train PPL ($\downarrow$) & StarCoder PPL ($\downarrow$) & Wiki PPL ($\downarrow$) \\ \midrule
    \multicolumn{6}{l}{\textit{LoRA rank 4}} \\
    4 & 0 & 2.380 & 1.359 & 2.519 & 10.243 \\
    4 & 0.01 & 4.000 & 1.359 & 2.515 & 10.227 \\
    4 & 0.1 & 4.000 & 1.358 & 2.510 & 10.230 \\
    4 & 1.0 & 4.000 & 1.359 & 2.509 & 10.222 \\
    \midrule
    \multicolumn{6}{l}{\textit{LoRA rank 8}} \\
    8 & 0.01 & 7.995 & 1.355 & 2.509 & 10.231 \\
    8 & 0.1 & 8.000 & 1.354 & 2.507 & 10.223 \\
    8 & 1.0 & 8.000 & 1.354 & 2.509 & 10.230 \\
    \midrule
    \multicolumn{6}{l}{\textit{LoRA rank 32}} \\
    32 & 0 & 14.939 & 1.356 & 2.527 & 10.318 \\
    32 & 0.01 & 31.695 & 1.355 & 2.510 & 10.261 \\
    32 & 0.1 & 31.999 & 1.354 & 2.504 & 10.258 \\
    32 & 1.0 & 32.000 & 1.354 & 2.507 & 10.243 \\
    \midrule
    \multicolumn{6}{l}{\textit{LoRA rank 64}} \\
    64 & 0 & 30.912 & 1.369 & 2.526 & 10.328 \\
    64 & 0.01 & 62.968 & 1.361 & 2.513 & 10.287 \\
    64 & 0.1 & 63.996 & 1.362 & 2.506 & 10.263 \\
    64 & 1.0 & 64.000 & 1.361 & 2.508 & 10.278 \\
    \midrule
    \multicolumn{6}{l}{\textit{LoRA rank 256}} \\
    256 & 0 & 132.050 & 1.419 & 2.537 & 10.391 \\
    256 & 0.01 & 249.371 & 1.405 & 2.518 & 10.279 \\
    256 & 0.1 & 255.885 & 1.404 & 2.510 & 10.283 \\
    256 & 1.0 & 255.990 & 1.406 & 2.510 & 10.297 \\
    \bottomrule
    \end{tabular}
    \end{adjustbox}
    \caption{LoRA rank and penalty-weight sweep on GSM8K for Qwen2.5-7B-Instruct.
}
\label{tab:gsm8k-lora-rank-sweep-qwen-25-7b}
\end{table}

\end{minipage}\hfill
\begin{minipage}[t]{\columnwidth}
  \vspace{0pt}
  \begin{table}[t]
    \centering
    \small
    \begin{adjustbox}{max width=\columnwidth}
    \begin{tabular}{@{}lrrrrr@{}}
    \toprule
    rank & $\lambda$ & eRank ($\uparrow$) & Train PPL ($\downarrow$) & StarCoder PPL ($\downarrow$) & Wiki PPL ($\downarrow$) \\ \midrule
    \multicolumn{6}{l}{\textit{LoRA rank 4}} \\
    4 & 0 & 3.051 & 1.757 & 3.565 & 16.799 \\
    4 & 0.01 & 4.000 & 1.756 & 3.544 & 16.670 \\
    4 & 0.1 & 4.000 & 1.756 & 3.544 & 16.703 \\
    4 & 1.0 & 4.000 & 1.758 & 3.546 & 16.710 \\
    \midrule
    \multicolumn{6}{l}{\textit{LoRA rank 8}} \\
    8 & 0 & 5.652 & 1.749 & 3.568 & 16.849 \\
    8 & 0.01 & 8.000 & 1.749 & 3.545 & 16.682 \\
    8 & 0.1 & 8.000 & 1.747 & 3.544 & 16.733 \\
    8 & 1.0 & 8.000 & 1.749 & 3.547 & 16.699 \\
    \midrule
    \multicolumn{6}{l}{\textit{LoRA rank 16}} \\
    16 & 0 & 10.806 & 1.741 & 3.577 & 16.914 \\
    16 & 0.01 & 15.998 & 1.740 & 3.543 & 16.697 \\
    16 & 0.1 & 16.000 & 1.740 & 3.545 & 16.726 \\
    16 & 1.0 & 16.000 & 1.740 & 3.545 & 16.760 \\
    \midrule
    \multicolumn{6}{l}{\textit{LoRA rank 32}} \\
    32 & 0 & 21.559 & 1.735 & 3.587 & 16.998 \\
    32 & 0.01 & 31.991 & 1.736 & 3.546 & 16.755 \\
    32 & 0.1 & 31.999 & 1.734 & 3.544 & 16.772 \\
    32 & 1.0 & 32.000 & 1.733 & 3.551 & 16.772 \\
    \midrule
    \multicolumn{6}{l}{\textit{LoRA rank 64}} \\
    64 & 0 & 44.033 & 1.731 & 3.597 & 17.116 \\
    64 & 0.01 & 63.954 & 1.733 & 3.546 & 16.791 \\
    64 & 0.1 & 63.995 & 1.732 & 3.546 & 16.794 \\
    64 & 1.0 & 63.999 & 1.730 & 3.552 & 16.836 \\
    \midrule
    \multicolumn{6}{l}{\textit{LoRA rank 128}} \\
    128 & 0 & 89.142 & 1.734 & 3.614 & 17.236 \\
    128 & 0.01 & 127.770 & 1.736 & 3.552 & 16.853 \\
    128 & 0.1 & 127.974 & 1.735 & 3.553 & 16.867 \\
    \midrule
    \multicolumn{6}{l}{\textit{LoRA rank 256}} \\
    256 & 0 & 172.453 & 1.745 & 3.647 & 17.462 \\
    256 & 0.01 & 254.767 & 1.745 & 3.569 & 16.914 \\
    256 & 0.1 & 255.858 & 1.743 & 3.572 & 16.929 \\
    256 & 1.0 & 255.984 & 1.742 & 3.574 & 17.025 \\
    \midrule
    \multicolumn{6}{l}{\textit{LoRA rank 512}} \\
    512 & 0 & 307.809 & 1.780 & 3.706 & 17.998 \\
    512 & 0.01 & 494.446 & 1.770 & 3.601 & 17.110 \\
    512 & 0.1 & 511.207 & 1.765 & 3.590 & 17.077 \\
    512 & 1.0 & 511.908 & 1.763 & 3.607 & 17.255 \\
    \bottomrule
    \end{tabular}
    \end{adjustbox}
    \caption{LoRA rank and penalty-weight sweep on GSM8K for Qwen2.5-0.5B.
}
\label{tab:gsm8k-lora-rank-sweep-qwen-25-05b}
\end{table}

\end{minipage}
\end{table*}

\section{Comparing LoRA ranks}

\subsection{Intruder-dimension analysis by LoRA rank (mixed task)}

Tables~\ref{tab:mixed-intruder-complement-r4} and~\ref{tab:mixed-intruder-complement-r64} report thresholded overlap metrics for the mixed continual-learning setting at LoRA ranks $4$ and $64$, with rank-$896$ reference rows in Table~\ref{tab:mixed-intruder-complement}.
The strong-match mass, \(1-\mathrm{WeightedUnaligned}\), is approximately $0.52$ at rank $4$, $0.18$ at rank $64$, and $0.04$ at rank $896$ for standard LoRA.
Thus lower-rank adapters show more very strong thresholded matches to pretrained singular directions under this diagnostic.

\begin{table*}[t]
\centering
\small
\begin{adjustbox}{max width=\textwidth}
\begin{tabular}{@{}lrrr@{}}
\toprule
Method & Weighted unaligned ($\downarrow$) & Unaligned count ($\downarrow$) & Unaligned rate (\%) ($\downarrow$) \\
\midrule
\multicolumn{4}{l}{\textit{LoRA rank 4}} \\
Standard LoRA & 0.484 & 3703 & 76.130 \\
LoRA ($\lambda=0.01$) & 0.463 & 3840 & 78.950 \\
LoRA ($\lambda=0.1$) & 0.484 & 3837 & 78.890 \\
LoRA ($\lambda=1.0$) & 0.451 & 3651 & 75.060 \\
\bottomrule
\end{tabular}
\end{adjustbox}
\caption{Unaligned spectral mass in the mixed continual-learning setting for LoRA rank $r=4$ (GSM8K, WMT19 cs--en, and Aya); \emph{lower} is better.
    Metrics follow Table~\ref{tab:mixed-intruder-complement} with the same $\epsilon=0.9$ and ambient $d=4864$.
    At rank $4$, unaligned rate is much lower ($\approx 75$--$79\%$) than at rank $896$ ($\gtrsim 96\%$) because strong-match counts exceed $1000$ rather than $\approx 140$.}
\label{tab:mixed-intruder-complement-r4}
\end{table*}

\begin{table*}[t]
\centering
\small
\begin{adjustbox}{max width=\textwidth}
\begin{tabular}{@{}lrrr@{}}
\toprule
Method & Weighted unaligned ($\downarrow$) & Unaligned count ($\downarrow$) & Unaligned rate (\%) ($\downarrow$) \\
\midrule
\multicolumn{4}{l}{\textit{LoRA rank 64}} \\
Standard LoRA & 0.823 & 4512.300 & 92.770 \\
LoRA ($\lambda=0.01$) & 0.825 & 4513.700 & 92.800 \\
LoRA ($\lambda=0.1$) & 0.834 & 4525.300 & 93.040 \\
LoRA ($\lambda=1.0$) & 0.823 & 4511.700 & 92.760 \\
\bottomrule
\end{tabular}
\end{adjustbox}
\caption{Unaligned spectral mass in the mixed continual-learning setting for LoRA rank $r=64$; \emph{lower} is better.
    Metrics follow Table~\ref{tab:mixed-intruder-complement} with the same $\epsilon=0.9$ and ambient $d=4864$.
    Rank $64$ sits between rank $4$ and rank $896$ on unaligned rate ($\approx 93\%$ vs.\ $\gtrsim 96\%$ at rank $896$).}
\label{tab:mixed-intruder-complement-r64}
\end{table*}

\subsection{Soft permutation analysis by LoRA rank (mixed task)}

Tables~\ref{tab:mixed-soft-permutation-r4} and~\ref{tab:mixed-soft-permutation-r64} report soft permutation metrics at LoRA ranks $4$ and $64$ in the mixed continual-learning setting.
At rank $4$, Standard LoRA has \(H_{\mathrm{soft}}\approx 0.77\); at rank $64$, \(H_{\mathrm{soft}}\approx 0.87\); at rank $896$, \(H_{\mathrm{soft}}\approx 0.91\).

\begin{table}[t]
\centering
\small
\begin{adjustbox}{max width=\columnwidth}
\begin{tabular}{@{}lr@{}}
\toprule
Method & $H_{\mathrm{soft}}$ ($\downarrow$) \\
\midrule
\multicolumn{2}{l}{\textit{LoRA rank 4}} \\
Standard LoRA & 0.775 \\
LoRA ($\lambda=0.01$) & 0.783 \\
LoRA ($\lambda=0.1$) & 0.788 \\
LoRA ($\lambda=1.0$) & 0.768 \\
\bottomrule
\end{tabular}
\end{adjustbox}
\caption{
    Soft permutation-metric analysis in the mixed continual-learning setting (GSM8K, WMT19 cs--en, and Aya) for LoRA rank $4$.
    Assignments are sharper than at ranks $64$ and $896$ ($H_{\mathrm{soft}}\approx 0.77$--$0.79$; Table~\ref{tab:mixed-soft-permutation-r64}; rank-896: $H_{\mathrm{soft}}\approx0.91$).
    \penaltyname{} weights have little effect on the entropy summary at this rank.
}
\label{tab:mixed-soft-permutation-r4}
\end{table}

\begin{table}[t]
\centering
\small
\begin{adjustbox}{max width=\columnwidth}
\begin{tabular}{@{}lr@{}}
\toprule
Method & $H_{\mathrm{soft}}$ ($\downarrow$) \\
\midrule
\multicolumn{2}{l}{\textit{LoRA rank 64}} \\
Standard LoRA & 0.872 \\
LoRA ($\lambda=0.01$) & 0.874 \\
LoRA ($\lambda=0.1$) & 0.875 \\
LoRA ($\lambda=1.0$) & 0.873 \\
\bottomrule
\end{tabular}
\end{adjustbox}
\caption{
    Soft permutation-metric analysis in the mixed continual-learning setting (GSM8K, WMT19 cs--en, and Aya) for LoRA rank $64$.
    Assignments remain diffuse ($H_{\mathrm{soft}}\approx 0.87$), slightly sharper than at rank $896$ for Standard LoRA ($H_{\mathrm{soft}}\approx 0.91$).
    \penaltyname{} weights have little effect on the entropy summary at this rank.
}
\label{tab:mixed-soft-permutation-r64}
\end{table}

\subsection{Subspace alignment analysis by LoRA rank (mixed task)}

Tables~\ref{tab:mixed-subspace-alignment-r4}--\ref{tab:mixed-subspace-alignment-r256} report subspace alignment metrics at LoRA ranks $4$, $64$, and $256$ in the mixed continual-learning setting.
Top-$k$ alignment changes with rank, but so does the uniform baseline: dense FT versus standard LoRA has alignment approximately $0.05$, $0.07$, and $0.09$ at ranks $4$, $64$, and $256$, respectively.
At rank $4$, the R\'enyi-2-penalized and standard LoRA update subspaces have alignment approximately $0.24$, compared with a uniform baseline of approximately $0.0008$.

\begin{table}[t]
\centering
\small
\begin{adjustbox}{max width=\columnwidth}
\begin{tabular}{@{}lr@{}}
\toprule
Comparison & $\operatorname{Align}_k$ \\
\midrule
Dense FT vs.\ Standard LoRA & 0.045 \\
\midrule
\multicolumn{2}{l}{\textit{\penaltyname{} vs.\ Dense FT}} \\
$\lambda=0.01$ vs.\ Dense FT & 0.053 \\
$\lambda=0.1$ vs.\ Dense FT & 0.056 \\
$\lambda=1.0$ vs.\ Dense FT & 0.054 \\
\midrule
\multicolumn{2}{l}{\textit{\penaltyname{} vs.\ Standard LoRA}} \\
$\lambda=0.01$ vs.\ Standard LoRA & 0.244 \\
$\lambda=0.1$ vs.\ Standard LoRA & 0.238 \\
$\lambda=1.0$ vs.\ Standard LoRA & 0.240 \\
\bottomrule
\end{tabular}
\end{adjustbox}
\caption{Dense FT denotes dense down-projection fine-tuning.
    $\lambda$ denotes the weight of the \penaltyname{} regularizer.
    Right-singular subspace alignment of updates $\Delta W=W_{\mathrm{ft}}-W_0$ in the mixed-domain setting (GSM8K, WMT19 cs--en, and Aya), with $k=r=4$.
    We report $\operatorname{Align}_k=\|V_{s,k}^\top V_{t,k}\|_F^2/k$, where $V_{s,k}$ and $V_{t,k}$ contain the leading $k$ right singular vectors of the two updates.
    The dense baseline update is also restricted to the leading $k$ directions.
    The independent random-orientation reference is $k/d\approx 0.000822$, with $d=4864$; this is an expected overlap, not a significance threshold.
}
\label{tab:mixed-subspace-alignment-r4}
\end{table}

\begin{table*}[t]
\renewenvironment{table}[1][]{}{}%
\begin{minipage}[t]{\columnwidth}
  \vspace{0pt}%
  \begin{table}[t]
\centering
\small
\begin{adjustbox}{max width=\columnwidth}
\begin{tabular}{@{}lr@{}}
\toprule
Comparison & $\operatorname{Align}_k$ \\
\midrule
Dense FT vs.\ Standard LoRA & 0.066 \\
\midrule
\multicolumn{2}{l}{\textit{\penaltyname{} vs.\ Dense FT}} \\
$\lambda=0.01$ vs.\ Dense FT & 0.068 \\
$\lambda=0.1$ vs.\ Dense FT & 0.075 \\
$\lambda=1.0$ vs.\ Dense FT & 0.071 \\
\midrule
\multicolumn{2}{l}{\textit{\penaltyname{} vs.\ Standard LoRA}} \\
$\lambda=0.01$ vs.\ Standard LoRA & 0.086 \\
$\lambda=0.1$ vs.\ Standard LoRA & 0.084 \\
$\lambda=1.0$ vs.\ Standard LoRA & 0.083 \\
\bottomrule
\end{tabular}
\end{adjustbox}
\caption{Dense FT denotes dense down-projection fine-tuning.
    $\lambda$ denotes the weight of the \penaltyname{} regularizer.
    Right-singular subspace alignment of updates $\Delta W=W_{\mathrm{ft}}-W_0$ in the mixed-domain setting (GSM8K, WMT19 cs--en, and Aya), with $k=r=64$.
    We report $\operatorname{Align}_k=\|V_{s,k}^\top V_{t,k}\|_F^2/k$, where $V_{s,k}$ and $V_{t,k}$ contain the leading $k$ right singular vectors of the two updates.
    The dense baseline update is also restricted to the leading $k$ directions.
    The independent random-orientation reference is $k/d\approx 0.013158$, with $d=4864$; this is an expected overlap, not a significance threshold.
}
\label{tab:mixed-subspace-alignment-r64}
\end{table}

\end{minipage}\hfill
\begin{minipage}[t]{\columnwidth}
  \vspace{0pt}
  \begin{table}[t]
\centering
\small
\begin{adjustbox}{max width=\columnwidth}
\begin{tabular}{@{}lr@{}}
\toprule
Comparison & $\operatorname{Align}_k$ \\
\midrule
Dense FT vs.\ Standard LoRA & 0.094 \\
\midrule
\multicolumn{2}{l}{\textit{\penaltyname{} vs.\ Dense FT}} \\
$\lambda=0.01$ vs.\ Dense FT & 0.095 \\
$\lambda=0.1$ vs.\ Dense FT & 0.100 \\
$\lambda=1.0$ vs.\ Dense FT & 0.102 \\
\midrule
\multicolumn{2}{l}{\textit{\penaltyname{} vs.\ Standard LoRA}} \\
$\lambda=0.01$ vs.\ Standard LoRA & 0.095 \\
$\lambda=0.1$ vs.\ Standard LoRA & 0.093 \\
$\lambda=1.0$ vs.\ Standard LoRA & 0.096 \\
\bottomrule
\end{tabular}
\end{adjustbox}
\caption{Dense FT denotes dense down-projection fine-tuning.
    $\lambda$ denotes the weight of the \penaltyname{} regularizer.
    Right-singular subspace alignment of updates $\Delta W=W_{\mathrm{ft}}-W_0$ in the mixed-domain setting (GSM8K, WMT19 cs--en, and Aya), with $k=r=256$.
    We report $\operatorname{Align}_k=\|V_{s,k}^\top V_{t,k}\|_F^2/k$, where $V_{s,k}$ and $V_{t,k}$ contain the leading $k$ right singular vectors of the two updates.
    The dense baseline update is also restricted to the leading $k$ directions.
    The independent random-orientation reference is $k/d\approx 0.052632$, with $d=4864$; this is an expected overlap, not a significance threshold.
}
\label{tab:mixed-subspace-alignment-r256}
\end{table}

\end{minipage}
\end{table*}

\FloatBarrier
\section{Implementation Details for Subspace Alignment Metrics}
\label{app:subspace-alignment-implementation}

This appendix describes implementation details for the subspace alignment metrics used in our analysis. The main goal is to compute alignment between a source singular-vector basis and a target top-\(k\) subspace efficiently and reproducibly.

\subsection{Basic Setup}

Let \(V_s,V_t\in\mathbb R^{d\times d}\) contain the right singular vectors of the source and target update matrices, respectively. For Tables~\ref{tab:mixed-subspace-alignment-r4}--\ref{tab:mixed-subspace-alignment-r256}, each update is defined as \(\Delta W=W_{\mathrm{ft}}-W_0\). For example, the source may be a R\'enyi-2-penalized LoRA solution and the target may be a dense down-projection fine-tuning solution or a standard LoRA solution.

We are interested in comparing subspaces spanned by selected sets of \(k\) columns of \(V_s\) to the target top-\(k\) subspace
\begin{equation}
    \mathcal S_t^{(k)}
    :=
    \operatorname{span}
    \{v_{t,1},\dots,v_{t,k}\}.
\end{equation}
Let
\begin{equation}
    Q_t := [v_{t,1},\ldots,v_{t,k}]\in\mathbb R^{d\times k}
\end{equation}
be an orthonormal basis for this target subspace.

For a subset \(R\subseteq\{1,\dots,d\}\) of size \(k\), let
\begin{equation}
    Q_{s,R}:=[v_{s,i}]_{i\in R}\in\mathbb R^{d\times k}
\end{equation}
be an orthonormal basis for the corresponding source subspace. The subspace alignment score is
\begin{equation}
    \operatorname{Align}(R,t)
    :=
    \frac{1}{k}
    \left\|
        Q_{s,R}^\top Q_t
    \right\|_F^2.
\end{equation}
Equivalently, if \(\theta_1,\dots,\theta_k\) are the principal angles between the source subspace and the target subspace, then
\begin{equation}
    \operatorname{Align}(R,t)
    =
    \frac{1}{k}
    \sum_{\ell=1}^k \cos^2\theta_\ell.
\end{equation}

Tables~\ref{tab:mixed-subspace-alignment-r4}--\ref{tab:mixed-subspace-alignment-r256} use the leading \(k\) singular directions of each update.

\subsection{Precomputing Per-Direction Target Overlap Scores}

A useful implementation trick is that one need not recompute a full \(k\times k\) subspace overlap for every subset \(R\). Instead, define the per-direction target-overlap scores
\begin{equation}
    q_i
    :=
    \left\|
        v_{s,i}^\top Q_t
    \right\|_2^2
    =
    \sum_{j=1}^k
    \left\langle v_{s,i}, v_{t,j}\right\rangle^2,
    \qquad i=1,\dots,d.
\end{equation}
In matrix form,
\begin{equation}
    q
    =
    \operatorname{rowsum}
    \left(
        (V_s^\top Q_t)\odot (V_s^\top Q_t)
    \right),
\end{equation}
where \(\operatorname{rowsum}\) denotes the vector of row sums.

Then, for any subset \(R\) of size \(k\),
\begin{equation}
    \operatorname{Align}(R,t)
    =
    \frac{1}{k}
    \sum_{i\in R} q_i.
\end{equation}
Thus, after a single matrix multiplication \(V_s^\top Q_t\), the subset alignments reduce to summing selected entries of the vector \(q\).

\subsection{Analytic Uniform Baseline}

A uniformly sampled subset provides a natural null baseline. If \(R\) is sampled uniformly without replacement from all \(k\)-element subsets of \(\{1,\dots,d\}\), then
\begin{equation}
    \operatorname{Align}_{\mathrm{unif}}(R,t)
    =
    \frac{1}{k}\sum_{i\in R}q_i.
\end{equation}
The expectation of this quantity has a closed form. Since \(V_s\) is an orthonormal basis and \(Q_t\) has orthonormal columns,
\begin{equation}
    \sum_{i=1}^d q_i
    =
    \left\|V_s^\top Q_t\right\|_F^2
    =
    \left\|Q_t\right\|_F^2
    =
    k.
\end{equation}
Therefore,
\begin{equation}
    \mathbb E_R
    \left[
        \operatorname{Align}_{\mathrm{unif}}(R,t)
    \right]
    =
    \frac{1}{d}\sum_{i=1}^d q_i
    =
    \frac{k}{d}.
\end{equation}
We therefore use
\begin{equation}
    \operatorname{Align}_{\mathrm{unif}}
    :=
    \frac{k}{d}
\end{equation}
as the analytic uniform baseline, avoiding unnecessary Monte Carlo estimation.

\end{document}